\documentclass[10pt]{article} 
\usepackage[accepted]{tmlr}

\usepackage{amsmath,amsfonts,bm}

\def\eqref#1{equation~\ref{#1}}

\def\1{\bm{1}}

\DeclareMathAlphabet{\mathsfit}{\encodingdefault}{\sfdefault}{m}{sl}
\SetMathAlphabet{\mathsfit}{bold}{\encodingdefault}{\sfdefault}{bx}{n}

\usepackage{hyperref}
\usepackage{url}

\usepackage[acronym]{glossaries}
\usepackage{svg}
\usepackage{subcaption}
\usepackage{tabularx}
\usepackage{booktabs}
\usepackage{multirow}
\usepackage{graphicx}
\usepackage{soul}

\title{Flipped Classroom:\\Effective Teaching for Time Series Forecasting}

\author{\name Philipp Teutsch \email philipp.teutsch@tu-ilmenau.de\\
      \addr Technische Universität Ilmenau
      \AND
      \name Patrick M{\"a}der \email patrick.maeder@tu-ilmenau.de\\
      \addr Technische Universität Ilmenau\\
      Friedrich-Schiller-Universität Jena}

\def\month{10}  
\def\year{2022} 
\def\openreview{\url{https://openreview.net/forum?id=w3x20YEcQK}} 

\definecolor{GoodGreen}{rgb}{0,0.6,0}
\definecolor{BadRed}{rgb}{0.8,0.2,0}

\newcommand{\good}[1]{\textcolor{GoodGreen}{#1}}
\newcommand{\bad}[1]{\textcolor{BadRed}{#1}}

\newacronym{amt}{AMT}{Automatic Machine Translation}
\newacronym{arnn}{AntisymmetricRNN}{Antisymmetric Recurrent Neural Network}
\newacronym{bptt}{BPTT}{Backpropagation Through Time}
\newacronym{cl}{CL}{Curriculum Learning}
\newacronym{cornn}{coRNN}{Coupled Oscillatory Recurrent Neural Network}
\newacronym{fr}{FR}{free running}
\newacronym{gan}{GAN}{Generative Adversarial Network}
\newacronym{gru}{GRU}{Gated Recurrent Unit}
\newacronym{irnn}{IRNN}{incremental Recurrent Neural Network}
\newacronym{lle}{LLE}{largest Lyapunov exponent}
\newacronym{lstm}{LSTM}{Long Short Term Memory}
\newacronym{lrnn}{LRNN}{Lipschitz RNN}
\newacronym{lt}{LT}{Lyapunov Time}
\newacronym{ml}{ML}{machine learning}
\newacronym{mse}{RMSE}{Root Mean Squared Error}
\newacronym{nlp}{NLP}{Natural Language Processing}
\newacronym{nrmse}{NRMSE}{Normalized Root Mean Squared Error}
\newacronym{ode}{ODE}{ordinary differential equation}
\newacronym{plrnn}{PLRNN}{piecewise linear recurrent neural network}
\newacronym{rl}{RL}{Reinforcement Learning}
\newacronym{rlrop}{RLROP}{Reduce Learning Rate on Plateau}
\newacronym{rmse}{RMSE}{Root Mean Squared Error}
\newacronym{rnn}{RNN}{Recurrent Neural Network}
\newacronym{stf}{STF}{sparse teacher forcing}
\newacronym{tcn}{TCN}{Time Convolutional Network}
\newacronym{tf}{TF}{teacher forcing}
\newacronym{tts}{TTS}{Text-To-Speech}
\newacronym{urnn}{uRNN}{unitary evolution RNN}
\newacronym{lsode}{LSODE}{Livermore solver for ordinary differential equations}
\newacronym{dde}{DDE}{delayed differential equation}

\begin{document}

\maketitle

\begin{abstract}
Sequence-to-sequence models based on LSTM and GRU are a most popular choice for forecasting time series data reaching state-of-the-art performance. Training such models can be delicate though. The two most common training strategies within this context are teacher forcing (TF) and free running (FR). TF can be used to help the model to converge faster but may provoke an exposure bias issue due to a discrepancy between training and inference phase. FR helps to avoid this but does not necessarily lead to better results, since it tends to make the training slow and unstable instead. Scheduled sampling was the first approach tackling these issues by picking the best from both worlds and combining it into a curriculum learning (CL) strategy. Although scheduled sampling seems to be a convincing alternative to FR and TF, we found that, even if parametrized carefully, scheduled sampling may lead to premature termination of the training when applied for time series forecasting. To mitigate the problems of the above approaches we formalize CL strategies along the training as well as the training iteration scale. We propose several new curricula, and systematically evaluate their performance in two experimental sets. For our experiments, we utilize six datasets generated from prominent chaotic systems. We found that the newly proposed increasing training scale curricula with a probabilistic iteration scale curriculum consistently outperforms previous training strategies yielding an NRMSE improvement of up to 81\% over FR or TF training. For some datasets we additionally observe a reduced number of training iterations. We observed that all models trained with the new curricula yield higher prediction stability allowing for longer prediction horizons.
\end{abstract}

\section{Introduction}\label{sec:introduction}
Advanced \glspl*{rnn} such as \gls*{lstm} \citep{hochreiter1997long} and \gls*{gru} \citep{cho2014properties} achieved significant results in predicting sequential data \citep{chung2014empirical,nowak2017lstm,yin2017comparative}. Such sequential data  can for example be textual data as processed for \gls*{nlp} tasks where \gls*{rnn} models were the method of choice  for a long time, before feed-forward architectures like transformers showed superior results in processing natural language data \citep{devlin2018bert,yang2019xlnet,radford2019language,brown2020language}. Shifting the view to the field of modeling dynamical or even chaotic systems, encoder-decoder \glspl*{rnn} are still the method of choice for forecasting such continuous time series data  \citep{wang2019neural,thavarajah2021fast,vlachas2018data,sehovac2019forecasting,sehovac2020deep,shen2020sequence,walther2022automatic,pandey2022direct}.

Nevertheless encoder-decoder \glspl*{rnn} do have their difficulties, especially when it comes to effectively training them. Firstly, there is the exposure bias issue that can appear when \gls*{tf} is used for training the model. \gls*{tf} is the strategy that is typically applied when training \glspl*{rnn} for time series sequence-to-sequence tasks \citep{williams1989learning}, regardless of the type of data. To understand the problem, we first give a quick side note about the motivation behind \gls*{tf}. Its main advantage is that it can significantly reduce the number of steps a model needs to converge during training and improve its stability \citep{miao2020flow}. However, \gls*{tf} may result in worse model generalization due to a discrepancy between training and testing data distribution. It is less resilient against self-induced perturbations caused by prediction errors in the inference phase \citep{sangiorgio2020robustness}. However, simply training the model without \gls*{tf} in \gls*{fr} mode instead does not necessarily provide convincing results, as can be seen in Section~\ref{subsec:results}. Rather a more sophisticated strategy is needed to guide the model through the training procedure.

Several authors propose methods to mitigate the exposure bias or improve training stability and results in general inventing certain strategies \citep{bengio2015scheduled,nicolai2020noise,lamb2016professor,liu2020teacher,dou2019attention,hofmann2021synaptic}. Even though most of these methods address the training for \gls*{nlp} tasks, such as text translation, text completion or image captioning, we focus on time series forecasting. Our datasets consist of sequences sampled from approximations of different chaotic systems. With these systems, the next state can always be derived from the past state(s) deterministically, but they also tend to be easily irritated by small perturbations. Thus the trained model needs to be especially resilient against small perturbations when auto-regressively predicting future values, otherwise those small errors will quickly accumulate to larger errors \citep{sangiorgio2020robustness}. These aspects make the kind of data we use especially challenging. Since \gls*{tf} avoids the error accumulation only during training, we argue the models are even more vulnerable to issues like the exposure bias and will thus more likely fail to forecast larger horizons.

Around the interest of predicting particularly (chaotic) dynamical systems with \glspl*{rnn}, another field has formed that very intensively studies several specialized \gls*{rnn}-based approaches proposed to stabilize the training process and prevent exploding gradients. Most of propositions include architectural tweaks or even new \gls*{rnn} architectures considering the specifics of dynamical systems and their theory \citep{lusch2018deep,vlachas2018data,schmidt2019identifying,champion2019data,chang2019antisymmetricrnn,rusch2020coupled,erichson2020lipschitz,rusch2021long,li2021markov}.

\citet{monfared2021train}, for example, performed a theoretical analysis relating \gls*{rnn} dynamics to loss gradients and argue that this analysis is especially insightful for chaotic systems. With this in mind they suggest a kind of \gls*{stf}, inspired by the work of \citet{williams1989learning} that uses information about the degree of chaos of the treated dynamical system. As a result, they form a training strategy that is applicable without any architectural adaptations and without further hyper-parameters. Their results using a vanilla \gls*{rnn}, a \gls*{plrnn} and an \gls*{lstm} for the Lorenz \citep{lorenz1963deterministic} and the Rössler \citep{rossler1976equation} systems show clear superiority of applying  chaos-dependent \gls*{stf}.

Reservoir computing \glspl*{rnn} were successfully applied to chaotic system forecasting and analysis tasks as well.  For example, \citet{pathak2017using} propose a reservoir computing approach that fits the attractor of chaotic  systems and predicts their Lyapunov exponents.

In this paper, we focus on CL training strategies like scheduled sampling that require no architectural changes of the model and thus can be applied easily for different and existing sequence-to-sequence (seq2seq) models. All presented strategies will be evaluated across different benchmark datasets. Our main contributions are the following: First, assembling a set of training strategies for encoder-decoder \glspl*{rnn} that can be applied for existing seq2seq models without adapting their architecture. Second, presenting a collection of strategies' hyper-parameter configurations that optimize the performance of the trained model. Third, proposing a ``flipped classroom'' like strategy that outperforms all existing comparable approaches on several datasets sampled from different chaotic systems. Fourth, proposing a method that yields  substantially better prediction stability and therefore allows for forecasting longer horizons.

The course of the paper continues with Section~\ref{sec:background}, where we provide the background of  sequence-to-sequence \glspl*{rnn} and the conventional ways to train them. We also give a short introduction to chaotic  behavior of our data. In Section~\ref{sec:related_work} we examine existing approaches dealing with the difficulty of training seq2seq \glspl*{rnn} in the context of different applications. Also, we give an overview of work treating \glspl*{rnn} for chaotic system prediction in particular. Section~\ref{sec:strategies} describes how we designed our training strategies and how they are applied. Further information about the experimental setup and our results we present in Section~\ref{sec:experiments}. In Section~\ref{sec:discussion} we discuss the results, the strengths and limitations of the different strategies. Finally, in Section~\ref{sec:conclusion}, we  conclude our findings.

\section{Background}\label{sec:background}
\begin{figure}[htb]
	\centering
	\begin{subfigure}{0.485\linewidth}
		\includegraphics[width=1.0\linewidth]{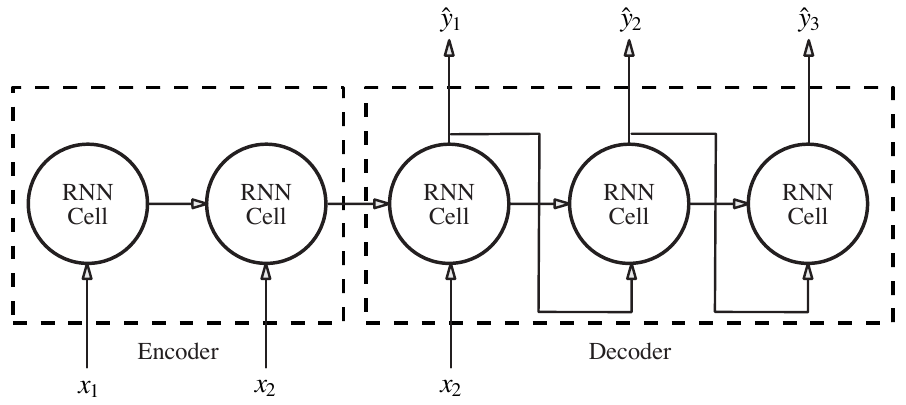}
		\caption{}
		\label{fig:encoder_decoder_architecture}
	\end{subfigure}
	\hspace*{\fill}
	\begin{subfigure}{0.485\linewidth}
		\includegraphics[width=1.0\linewidth]{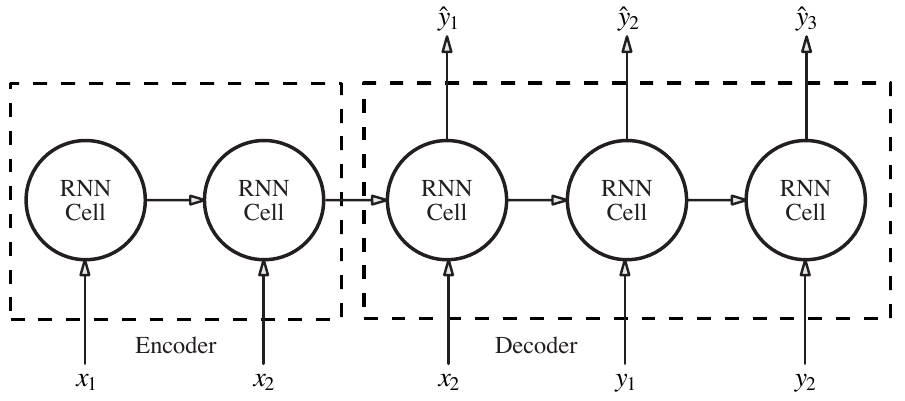}
		\caption{}
		\label{fig:tf_encoder_decoder_architecture}
	\end{subfigure}

	\caption{RNN encoder-decoder architecture in inference phase (a) and in training phase using  teacher forcing (b)}
\end{figure}

Within this paper, we study \gls*{cl} training strategies for sequence-to-sequence \glspl*{rnn}. We are using an encoder-decoder architecture \citep{chung2014empirical} as model to forecast time series data. The rough structure of the encoder-decoder architecture is shown in Figure~\ref{fig:encoder_decoder_architecture}. It consists of two separate \glspl*{rnn}, an encoder and a decoder. The encoder is trained to build up a hidden state representing the recent history of the processed time series. It takes an input sequence $(x_1,x_2,\dots,x_n)$ of $n$ input values where each value $x_j \in \mathbb{R}^d$ is a $d$ dimensional vector. For each processed step, the encoder updates its hidden state to provide context information for the following steps. The last encoder hidden state (after processing $x_n$) is then used as initial hidden state of the decoder. Triggered by a sequence's preceding value, the decoder's task is predicting its next future value while taking into account the sequence's history via the accumulated hidden state. For a trained network in the inference phase, that means that the decoder's preceding prediction is auto-regressively fed back as input into the decoder (aka auto-regressive prediction) (cp. Fig.~\ref{fig:encoder_decoder_architecture}). All $m$ outputs $y_j \in \mathbb{R}^d$ together form the output sequence $(y_1, y_2, \dots, y_m)$.

\subsection{Training Sequence Prediction Tasks}
While training, the decoder's inputs may either be previously predicted outputs (free running) or known previous values stemming from the given dataset (teacher forcing). Training in \textbf{free running} mode might be an intuitive approach, but slows down the training. This is due to accumulated error throughout multiple prediction steps without the support of teacher forced inputs -- especially in the early training epochs \citep{nicolai2020noise}. In contrast, \textbf{teacher forcing} aims to avoid this problem by utilizing the corresponding previous ground truth value as decoder input rather than the previously predicted value. This way, the model learns from the beginning of the training on to adapt its weights to perfect input values and converges noticeable faster \citep{miao2020flow} (cp. Fig.~\ref{fig:tf_encoder_decoder_architecture}). However,\gls*{tf} also bears a significant drawback, since during training the model is never exposed to the noisy predicted values it will later face in the inference phase. Therefore does often not generalize very well. A model trained with \gls*{tf} solely learned to predict on basis of perfect values and is thus vulnerable for small perturbations on its input, this effect is called \textbf{exposure bias} \citep{ranzato2015sequence}.

\subsection{Chaotic Systems}\label{subsec:chaotic_systems}
The data we are using for our experiments later in Section~\ref{sec:experiments} is generated from chaotic systems. Whether or not a dynamical  system is chaotic, can be confirmed by considering its Lyapunov exponents $\lambda_k$ for $k \in [1, d]$. Given an initial perturbation $\varepsilon_0$ the exponential rate with which the perturbation will increase (or decrease) in the direction of dimension $i$ is the Lyapunov exponent $\lambda_k~=~\lim\limits_{t \rightarrow \infty} \frac{1}{t}  \ln(\frac{||\varepsilon_t||}{||\varepsilon_0||})$ \citep{dingwell2006lyapunov}. That is, the Lyapunov exponents denote how sensitive the system is to the initial conditions (initial state). A deterministic dynamical system with at least one positive Lyapunov exponent while being aperiodic in its asymptotic limit is called \textbf{chaotic}  \citep{dingwell2006lyapunov}. Analogously, dynamical systems with at least two positive, one negative, and one zero  Lyapunov exponent are called \textbf{hyper-chaotic} systems. Dingwell points out that the \gls*{lle} can be used as a measure  to compare the chaotic behavior of different systems.

\section{Related Work}
\label{sec:related_work}
\citet{schmidt2019generalization} defines exposure bias as describing ``a lack of generalization with respect to an -- usually implicit and potentially task and domain dependent -- measure other than maximum-likelihood'' meaning that when the exclusive objective is to maximize the likelihood between the output and the target sequence, one can use \gls*{tf} during training \citep{goodfellow2016deep}. However, Goodfellow et al. argue that the kind of input the model sees while testing will typically diverge from the training data and the trained model may lack the ability of correcting its own mistakes. Thus, in practice, the \gls*{tf} can be a proper training strategy but may hinder the model to learn compensating its inaccuracies. He et\,al. study exposure bias for natural language generation tasks \citep{he2019quantifying}. They use sequence- and word-level quantification metrics to observe the influence of diverging prefix distributions on the distribution of the generated sequences. Two distributions are generated. One with and one without induced exposure bias. Those two distributions are then compared on basis of the corresponding corpus-bleu scores \citep{papineni2002bleu}. The study concludes that for language generation, the effect of the exposure bias is less serious than widely believed.

Several studies propose approaches to overcome the exposure bias induced by \gls*{tf}. The earliest of these studies proposes \textbf{scheduled sampling} \citep{bengio2015scheduled}. Scheduled sampling tries to take the advantages of training with \gls*{tf} while also acclimating the trained model to its own generated data error. It does that by using ground truth values as input for a subset of the training steps and predicted values for the remaining. $\epsilon_{i}$ denotes the \gls*{tf} probability at step $i$. Accordingly, the probability of using the predicted value is $1 - \epsilon_i$. During training $\epsilon_{i}$ decreases from  $\epsilon_{s}$ to $\epsilon_{e}$. This procedure changes the input data while training, making it a curriculum learning approach, as which scheduled sampling was proposed and works without major architectural adaptions. Originally proposed for image captioning tasks, scheduled sampling was also applied, e.g., for sound event detection \citep{drossos2019language}. Nicolai and Silfverberg consider and study the \gls*{tf} probability $\epsilon$ as a hyper-parameter. Rather than using a decay function that determines the decrease of $\epsilon_i$ over the course of training epochs, they use a fix \gls*{tf} probability throughout the training \citep{nicolai2020noise}. They observed a moderate improvement compared to strict \gls*{tf} training. Scheduled sampling is not restricted to \gls*{rnn}-based sequence-to-sequence models though, it has also been studied for transformer architectures \citep{mihaylova2019scheduled}. Mihaylova and Martins tested their modified transformer on two translation tasks but could only observe improved test results for one of them.

Apart from scheduled sampling \citep{bengio2015scheduled}, a number of approaches have been proposed typically aiming  to mitigate the exposure bias problem by adapting model architectures beyond an encoder-decoder design. \textbf{Professor forcing} \citep{lamb2016professor} is one of these more interfering approaches that aims to guide the teacher forced model in  training by embedding it into a \gls*{gan} framework \citep{goodfellow2014generative}. This framework consists of two  encoder-decoder \glspl*{rnn} that respectively form the generator and the discriminator. The generating \glspl*{rnn} have  shared weights that are trained with the same target sequence using their respective inputs while at the same time they try to  fool the discriminator by keeping their hidden states and outputs as similar as possible. The authors conclude that their  method, compared to \gls*{tf}, provides better generalization for single and multi-step prediction. In the field of  \gls*{tts}, the concept of professor forcing has also been applied in the \gls*{gan}-based training algorithm proposed by  \citet{guo2019new}. They adapted professor forcing and found that replacing the \gls*{tf} generator with one that uses scheduled sampling improved the results of their \gls*{tts} model in terms of intelligibility. As another approach for  \gls*{tts},  \citep{liu2020teacher} proposed \textbf{teacher-student training} using a training scheme to keep the hidden states  of the model in \gls*{fr} mode close to those of a model that was trained with \gls*{tf}. It applies a compound  objective function to align the states of the teacher and the student model. The authors observe improved naturalness and  robustness of the synthesized speech compared to their baseline. 

Dou~et\,al. \citep{dou2019attention} proposed  \textbf{attention forcing} as yet another training strategy for sequence-to-sequence models relying on an attention  mechanism that forces a reference attention alignment while training the model without \gls*{tf}. They studied \gls*{tts}  tasks and observed a significant gain in quality of the generated speech. The authors conclude that attention forcing is especially robust in cases where the order of predicted output is irrelevant.

The discussed approaches for mitigating exposure bias were proposed in the context of \gls*{nlp} and mainly target speech or text generation. For time series forecasting, \citet{sangiorgio2020robustness} suggest to neglect \gls*{tf} completely and solely train the model in \gls*{fr} mode, thereby, sacrificing the faster convergence of \gls*{tf} and potentially not reaching convergence at all.

In the context of \glspl*{rnn} for forecasting and analyzing dynamical systems, the majority of existing work deals with exploding and vanishing gradients as well as capturing long-term dependencies while preserving the expressiveness of the network. Various studies rely on methods from dynamical systems theory applied to \gls*{rnn} or propose new network architectures.

\citet{lusch2018deep} and \citet{champion2019data} use a modified autoencoder to learn appropriate eigenfunctions that the Koopman operator needs to linearize the nonlinear dynamics of the system. In another study, \citet{vlachas2018data} extend an \gls*{lstm} model with a mean stochastic model to keep its state in the statistical steady state and prevent it from escaping the system's attractor. \citet{schmidt2019identifying} propose a more generalized version of a \gls*{plrnn} \citep{koppe2019identifying} by utilizing a subset of regularized memory units that hold information much longer and can thus keep track of long-term dependencies while the remaining parts of the architecture are designated to approximate the fast-scale dynamics of the underlying dynamical system. The \gls*{arnn} introduced by \citet{chang2019antisymmetricrnn} represents an \gls*{rnn} designed to inherit the stability properties of the underlying \gls*{ode}, ensuring trainability of the network together with its capability of keeping track of long-term dependencies. A similar approach has been proposed as \glspl*{cornn} \citep{rusch2020coupled} that are based on a secondary order \glspl*{ode} modeling a coupled network of controlled forced and damped nonlinear oscillators. The authors prove precise bounds of the \gls*{rnn}'s state gradients and thus the ability of the \gls*{cornn} being a possible solution for exploding or vanishing gradients. \citet{erichson2020lipschitz} propose the Lipschitz \gls*{rnn} having additional hidden-to-hidden matrices enabling the \gls*{rnn} to remain Lipschitz continuous. This stabilizes the network and alleviates the exploding and vanishing gradient problem. In \citet{li2020fourier, li2021markov}, the authors propose the Fourier respectively the Markov neural operator that are built from multiple concatenated Fourier layers that directly work on the Fourier modes of the dynamical system. This way, they retain major portion of the dynamics and forecast the future behavior of the system. Both, the \gls*{irnn} \citep{kag2019rnns} and the time adaptive \gls*{rnn} \citep{kag2021time} use additional recurrent iterations on each input to enable the model of coping different input time scales, where the latter provides a time-varying function that adapts the model's behavior to the time scale of the provided input.

All of this shows the increasing interest in the application of \gls*{ml} models for forecasting and analyzing (chaotic) dynamical systems. To meet this trend, \citet{gilpin2021chaos} recently published a fully featured collection of benchmark datasets being related to chaotic systems including their mathematical properties.

A more general guide of training \glspl*{rnn} for chaotic systems is given by \citet{monfared2021train}. They discuss under which conditions the chaotic behavior of the input destabilizes the \gls*{rnn} and thus leads to exploding gradients during training. As a solution they propose \gls*{stf}, where every $\tau$-th time step a true input value is provided (teacher forced) as input instead of the previous prediction.

\begin{figure*}[htb]
	\centering
	\includegraphics[width=1.0\linewidth]{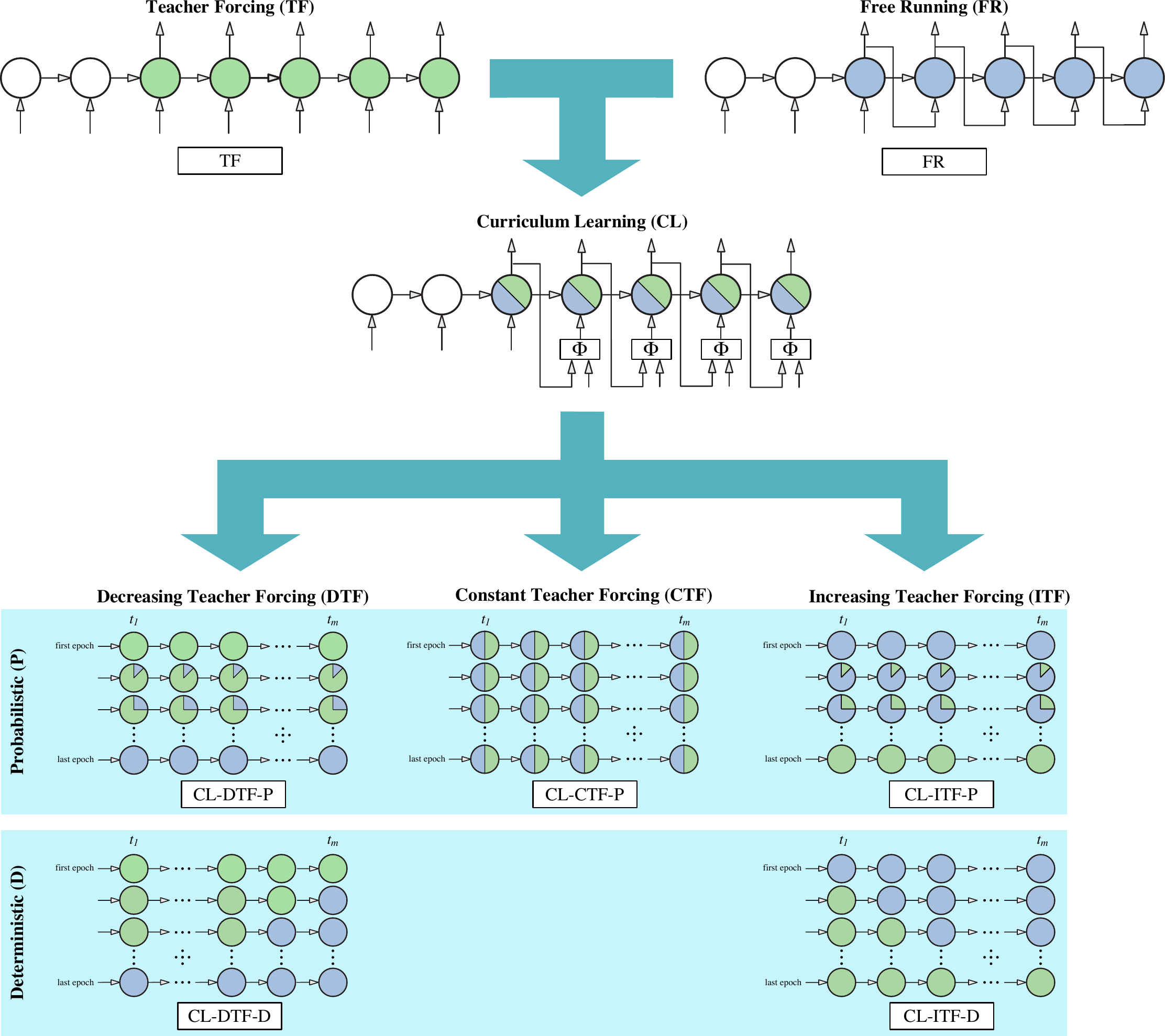}
	
	\caption{Overview of the proposed and evaluated training strategies where teacher forcing (TF) and free running (FR) refer to the two extreme cases that are combined to different \gls*{cl} strategies.}
	\label{fig:all_data_flows}
\end{figure*}

\section{Teaching Strategies}\label{sec:strategies}
Within this section, we systematically discuss existing teaching strategies for sequence-to-sequence prediction models and propose new strategies. All of these will then be evaluated in an experimental study with different time series reported in the following section.

\begin{figure}[htb]
	\centering
	\begin{subfigure}{0.4\linewidth}
		\includegraphics[trim={0.5cm, 0, 0, 0}, width=1.0\linewidth]{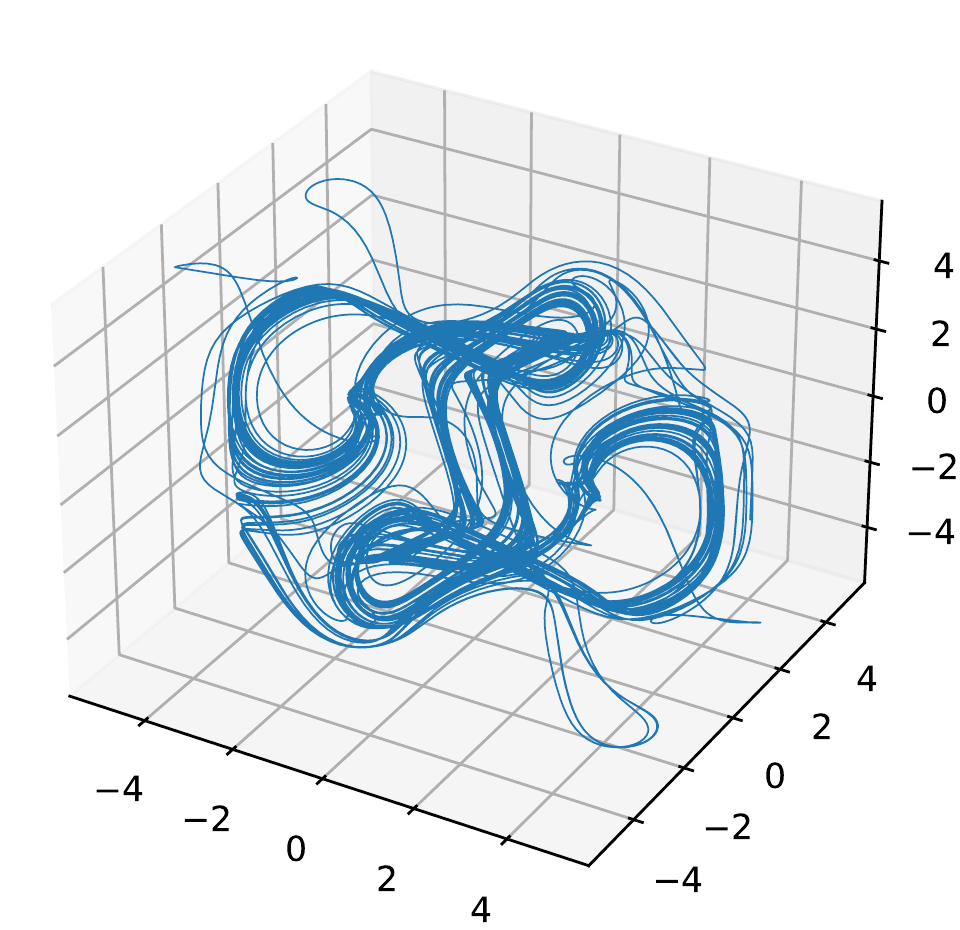}
		\caption{}
		\label{fig:thomas_chaotic_attractor}
	\end{subfigure}
	\begin{subfigure}{0.4\linewidth}
		\includegraphics[trim={0, 0, 0.5cm, 0}, width=1.0\linewidth]{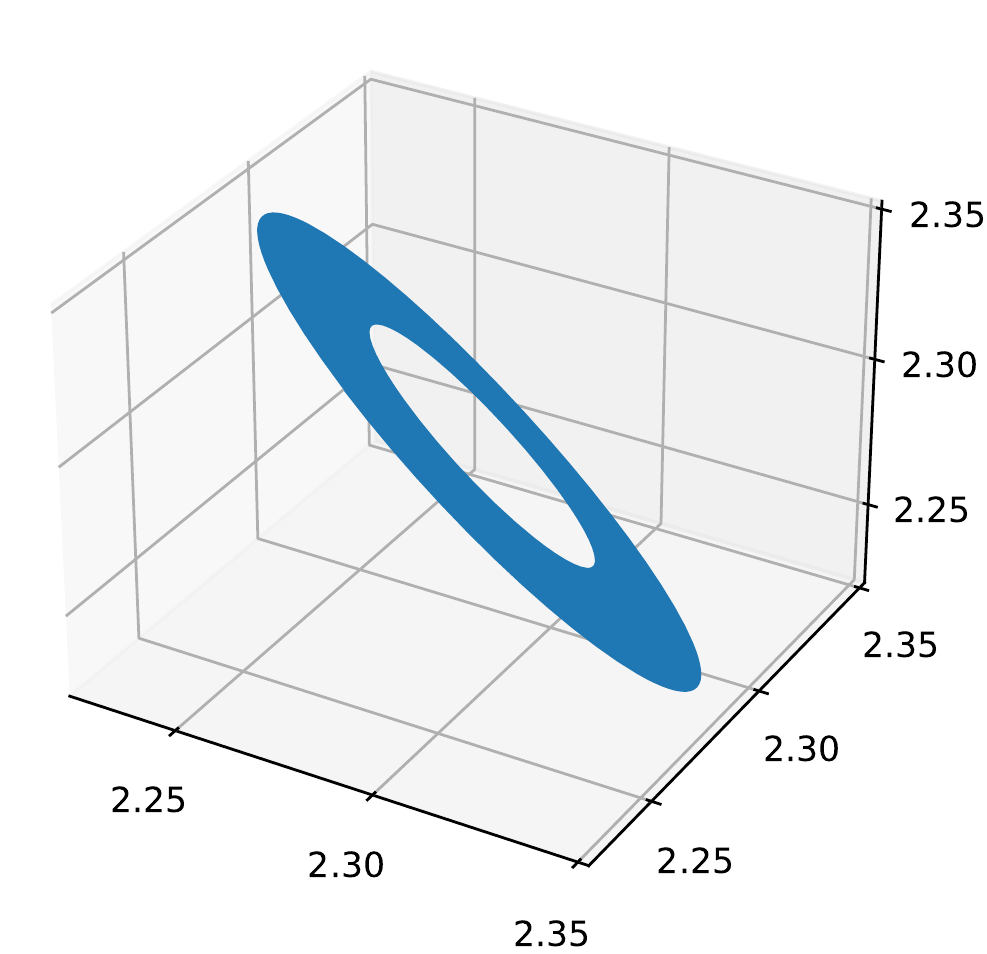}
		\caption{}
		\label{fig:thomas_periodic_attractor}
	\end{subfigure}

	\caption{$30\,000$ time steps sampled with a time delta of $dt=0.1$ of Thomas' cyclically symmetric attractor in (a) a chaotic parametrization and (b) a periodic parametrization}
\end{figure}
	
\begin{figure}[htb]
			\centering
			\includegraphics[trim=0 0 23 20,clip,width=0.55\linewidth]{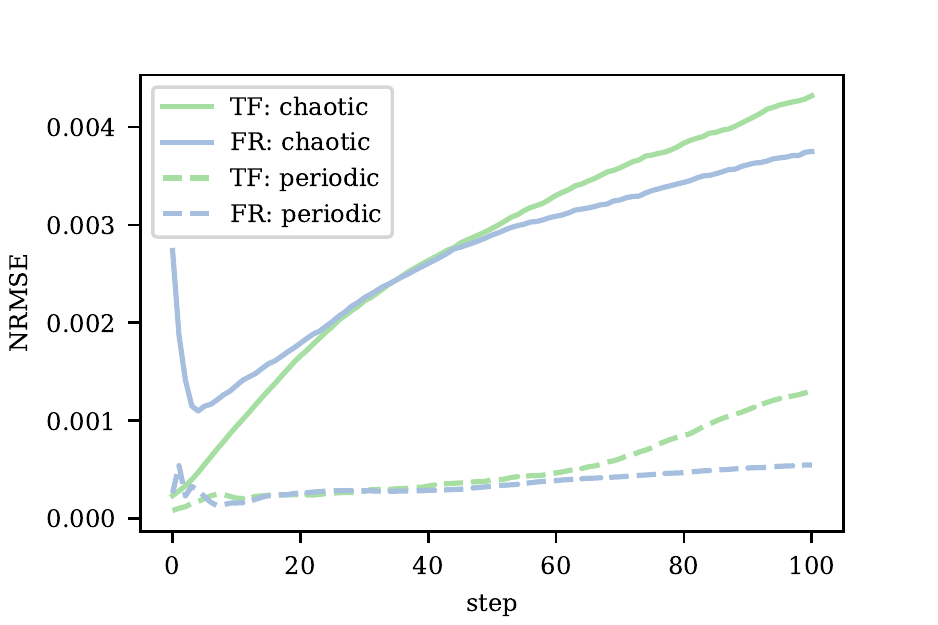}

		\caption{Test NRMSE over 100 predicted time steps of the chaotically and periodically parametrized Thomas attractor (cp. Fig.~\ref{fig:thomas_periodic_attractor}), predicted by GRU models trained with teacher forcing (TF) and free running (FR).}
\label{fig:pretest_plot}
\end{figure}

\subsection{Free Running (FR) vs. Teacher Forcing (TF)}
A widely used training strategy for \gls*{rnn} sequence-to-sequence models is to use \gls*{tf} throughout the whole training. Thereby, data is processed as shown in Fig.~\ref{fig:all_data_flows} (top left), i.e., the model is never exposed to its own predictions during training. A single forward step of the decoder during \gls*{tf} training is denoted as
\begin{equation}\label{eq:teacher_forcing}
	\hat{y}^t = f(y^{t-1}, \theta, c^{t-1}),
\end{equation}
where $y^t$ is the ground truth value for time step $t$, $\hat{y}^t$ is the predicted value for time step $t$, $\theta$ denotes the trainable parameters of the model, and $c^t$ is the decoder's hidden state at time step $t$.

The other extreme form of training is \gls*{fr}, i.e., only the model's first output value is predicted on basis of ground truth input and all subsequent output values of the sequence are predicted on basis of previous predictions throughout the training (cp. Fig.~\ref{fig:all_data_flows} (top right)). A single forward step of the decoder during \gls*{fr} training is denoted as
\begin{equation}\label{eq:free_running}
	\hat{y}^t = f(\hat{y}^{t-1}, \theta, c^{t-1}).
\end{equation}

A claimed major benefit of \gls*{tf} training is faster model convergence and thus reduced training time \citep{miao2020flow}, while a major benefit of \gls*{fr} training is avoided exposure bias arising from solely training with ground truth data, yielding a model that performs less robust on unseen validation data \citep{ranzato2015sequence}. To illustrate these benefits and drawbacks, we utilize the Thomas attractor with two parametrizations, the first resulting in a periodic (cp. Fig.~\ref{fig:thomas_periodic_attractor}) and the second resulting in a chaotic attractor (cp. Fig.~\ref{fig:thomas_chaotic_attractor}). By sampling from the attractors, we build two corresponding datasets of $10\,000$ samples each. For both datasets, we train a single layer encoder-decoder \gls*{gru} following the \gls*{fr} and the \gls*{tf} strategy. Figure~\ref{fig:pretest_plot} shows the \gls*{nrmse} per trained model over $100$ predicted time steps. All models have been initialized with $150$ ground truth values to build up the hidden state before predicting these $100$ time steps. We observe that the chaotic variant is harder to predict for the trained models (cp. blue and green line in the figure) and that those trained with \gls*{tf} tend to predict with a smaller error at the first steps, which then grows relatively fast. In contrast, the prediction error of the \gls*{fr}-trained models starts on a higher level but stays more stable over the prediction horizon, arguing that time series forecasting represents an especially challenging task for sequence-to-sequence models when it exhibits chaotic behavior. The more precise forecasting capabilities of a TF-trained network at the early prediction steps vs. the overall more stable long-term prediction performance of a \gls*{fr}-trained network observed in the Thomas example (cp. Fig~\ref{fig:thomas_chaotic_attractor},~\ref{fig:thomas_periodic_attractor}), motivates the idea of combining both strategies into a \gls*{cl} approach.

\begin{figure*}[htb]
	\centering
	\includegraphics[width=0.55\linewidth]{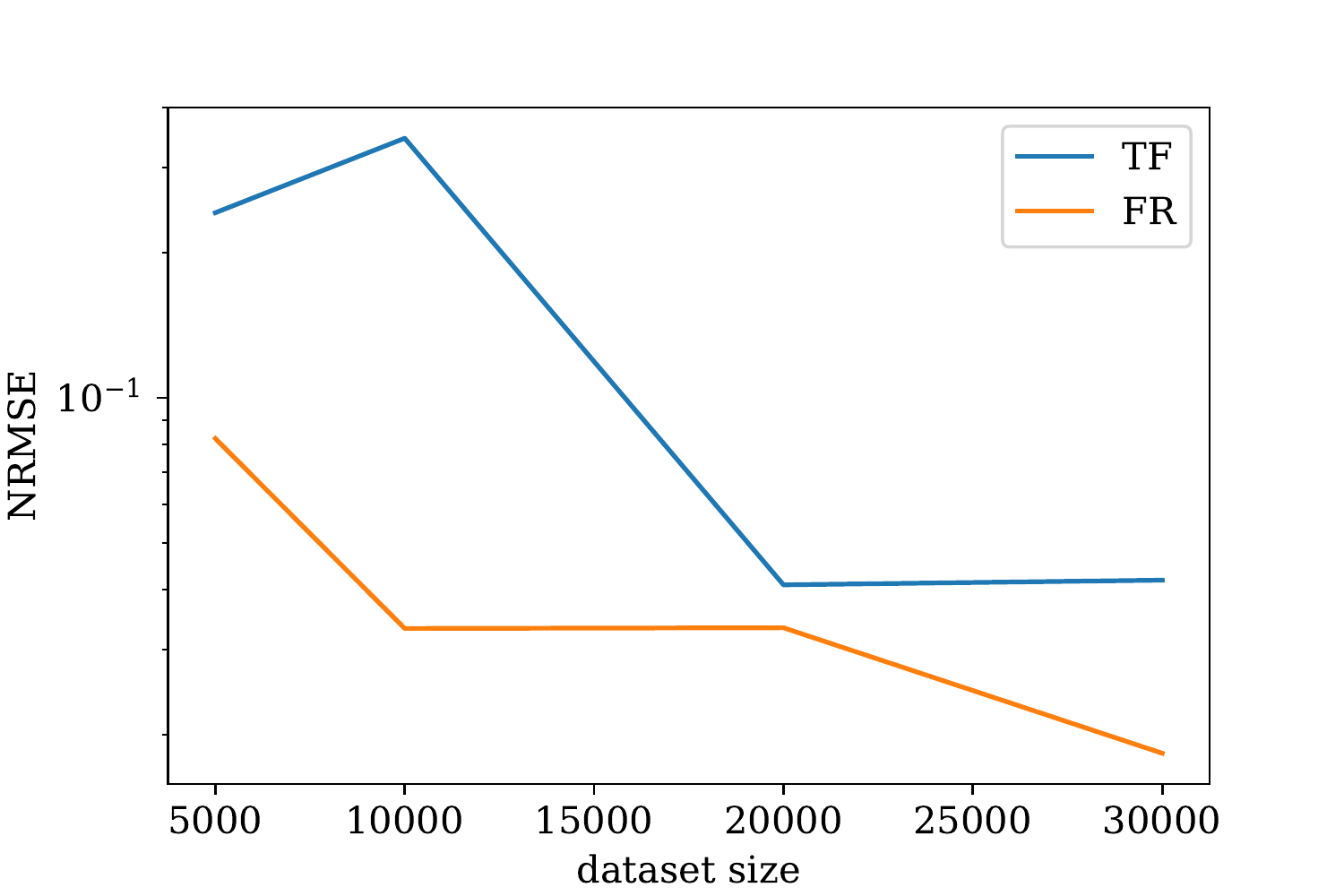}
	\caption{The NRMSE for different dataset sizes when using TF and FR during training for forecasting the Thomas attractor}
	\label{fig:data_sizes_thomas}
\end{figure*}

\citet{schmidt2019generalization} describes the exposure bias in natural language generation as a lack of generalization. This argumentation motivates an analysis of training with \gls*{fr} and \gls*{tf} strategies when applied to forecasting dynamical systems with different amounts of available training data.  Figure~\ref{fig:data_sizes_thomas} shows the NRMSE when forecasting the Thomas attractor using different dataset sizes and reveals that increasing the dataset size yields generally improved model performance for \gls*{tf} as well as \gls*{fr}, while their relative difference is maintained.

\subsection{Curriculum Learning (CL)}
\label{subsec:cl}
Within the context of our work, we denote the curriculum learning concept as combining \gls*{tf} and \gls*{fr} training, i.e., starting from the second decoder step, the curriculum prescribes per decoder step whether to use the ground truth value or the predicted value of the previous time step as input. We formalize a single training step of a \gls*{cl} approach as follows:

\begin{equation}
	\hat{y}^t= \begin{cases}
		f(y^{t-1}, \theta, c^{t-1}), & \text{if } \Phi = 1 \\
		f(\hat{y}^{t-1}, \theta, c^{t-1}), & \text{otherwise}
	\end{cases}
\end{equation}

where the \gls*{tf} decision $\Phi$ governs whether the decoder input is teacher forced or not.  Figure~\ref{fig:all_data_flows} illustrates the data flow of a sequence-to-sequence model training with \gls*{cl} in between the conventional strategies. In our naming scheme CL-DTF-P resembles the scheduled sampling approach proposed by \citet{bengio2015scheduled}. Below, we discuss the different types of curricula on training and iteration scale resulting in different ways for determining $\Phi$.

\subsection{Curriculum on Training Scale}\label{subsec:curriculum_types}
The \gls*{tf} ratio $\epsilon_{i}$ per training iteration $i$ is determined by a curriculum function $C: \mathbb{N}
\rightarrow [0,1]$.
We distinguish three fundamental types of curriculum on training scale. First, constant curricula, where a constant amount of \gls*{tf} is maintained throughout the training denoted as
\begin{equation}\label{eq:constant_epsilon}
\epsilon_{i}=\epsilon.
\end{equation}

\begin{figure}[htb]
	\centering
	\begin{subfigure}{0.4\linewidth}
		\includegraphics[trim=0 0 20 28,clip,width=1.0\linewidth]{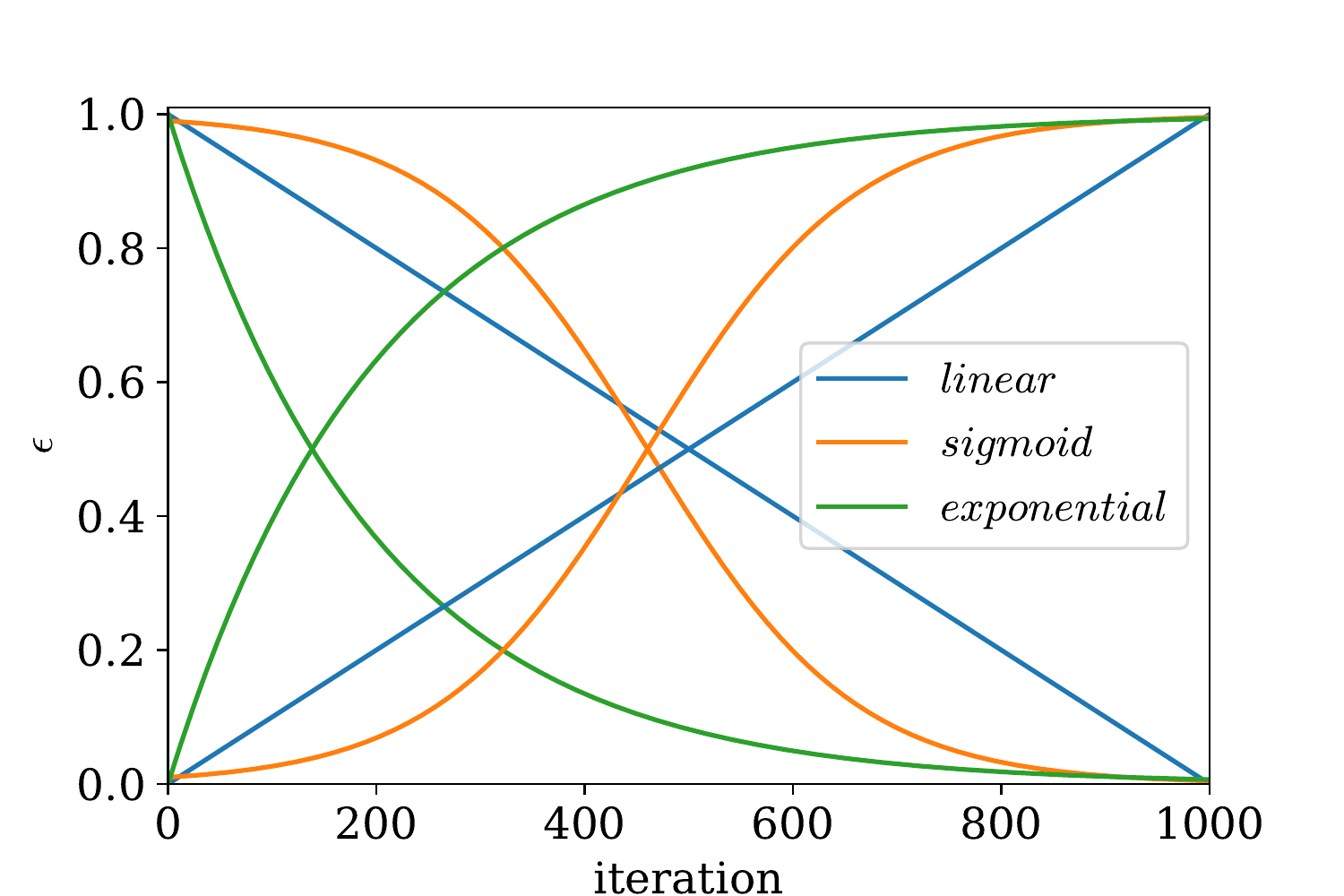}
	\caption{}
	\label{fig:decay_schedules}
	\end{subfigure}
	\begin{subfigure}{0.4\linewidth}
		\includegraphics[trim=0 0 16 30,clip,width=1.0\linewidth]{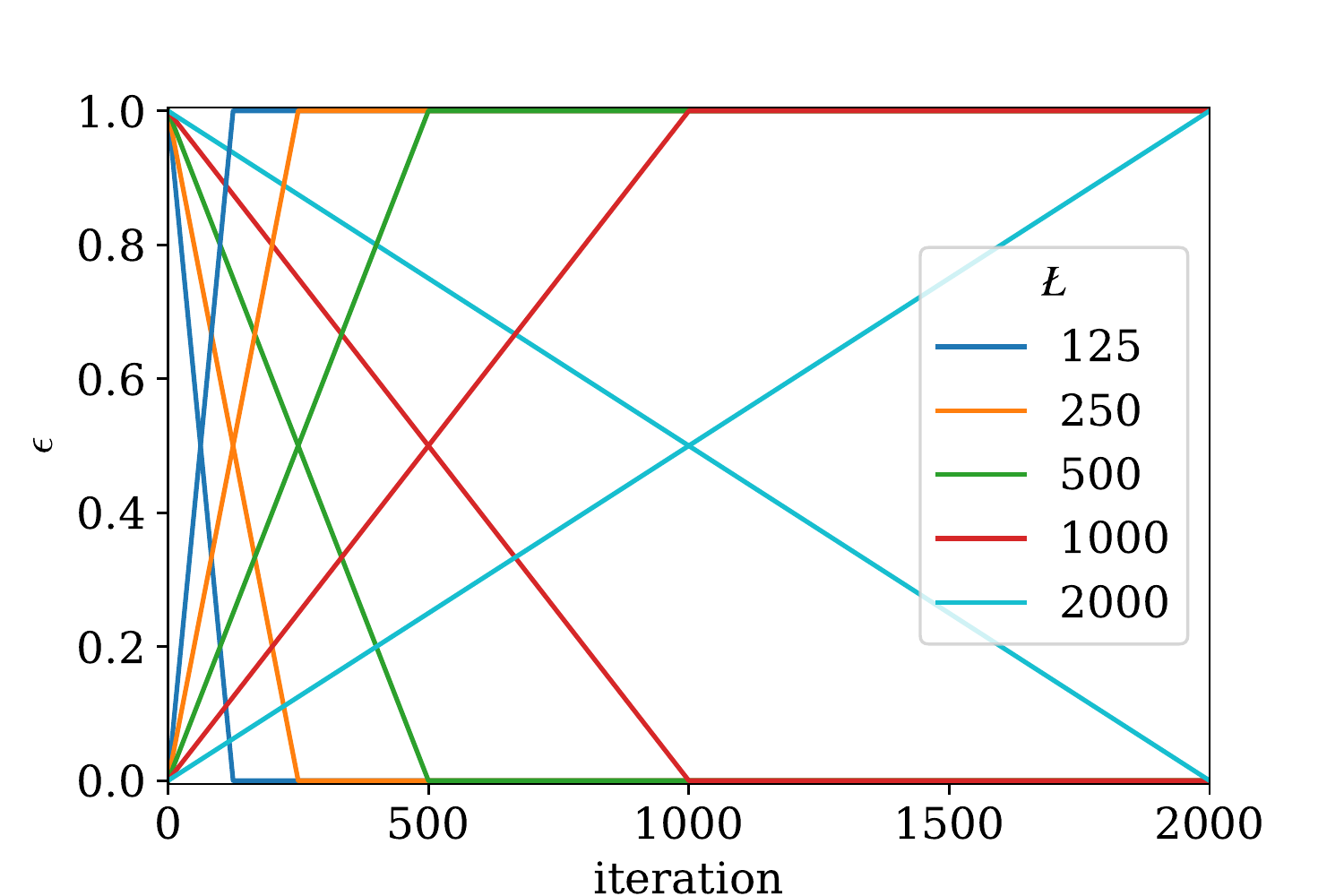}
	\caption{}
	\label{fig:more_decay_schedules}
	\end{subfigure}

	\caption{Examples of different decreasing curricula and their corresponding increasing versions (a) and multiple linear curricula with different pace $\textrm{\L}$ (b).}
\end{figure}

Second, decreasing curricula, where the training starts with a high amount of \gls*{tf} that continuously declines throughout the training. Third, increasing curricula, where the training starts at a low amount of \gls*{tf} that continuously increases throughout the training. Both follow a transition function $C: \mathbb{N} \rightarrow
[\epsilon_{start},\epsilon_{end}]$ denoted as
\begin{equation}\label{eq:constant_curriculum}
\epsilon_i=C(i),
\end{equation}
where $\epsilon_{start}\le\epsilon_{i}\le\epsilon_{i+1}\le\epsilon_{end}$ for increasing curricula, $\epsilon_{start}\ge\epsilon_{i}\ge\epsilon_{i+1}\ge\epsilon_{end}$ for decreasing curricula and $\epsilon_{start}\neq\epsilon_{end}$ for both. The following equations exemplary specify decreasing curricula (cp. Eqs.~\ref{eq:linear_curriculum}--\ref{eq:exponential_curriculum}) following differing transition functions inspired by those used to study the scheduled sampling approach \citep{bengio2015scheduled}
\begin{eqnarray}\label{eq:linear_curriculum}
C_{lin}(i) &=& \max(\epsilon_{end}, \epsilon_{end} + (\epsilon_{start} - \epsilon_{end}) \cdot (1 - \frac{i}{\textrm{\L}})), \nonumber\\
&& \text{with}\ \epsilon_{end} < \frac{\textrm{\L} - 1}{\textrm{\L}},\ \ 1< \textrm{\L},\ \ i \in \mathbb{N},\\
\label{eq:inversesigmoid_curriculum}
C_{invSig}(i) &=& \epsilon_{end} + (\epsilon_{start} - \epsilon_{end}) \cdot \frac{k}{k + e^\frac{i}{k}}, \nonumber\\
&& \text{with}\ \epsilon_{end} < \epsilon_{start},\ \ 1 \leq k,\ \ i \in \mathbb{N},\\
\label{eq:exponential_curriculum}
C_{exp}(i) &=& \epsilon_{end} + (\epsilon_{start} - \epsilon_{end}) \cdot k^i, \nonumber\\
&& \text{with}\ \epsilon_{end} < \epsilon_{start},\ \ 0 < k < 1,\ \ i \in \mathbb{N},
\end{eqnarray}

where the curriculum length parameter $\textrm{\L}$ determines the pace as number of iterations which the curriculum $C_{lin}$ needs to transition from $\epsilon_{start}$ to $\epsilon_{end}$. The curricula $C_{invSig}$ and $C_{exp}$ have no such parameter since the functions never completely reach $\epsilon_{end}$ in theory. In practice though, we adapt the curriculum-specific parameter $k$ to stretch or compress these curricula along the iteration axis to achieve the same effect. Figure~\ref{fig:decay_schedules} exemplary visualizes three decreasing and three increasing curricula, following differing transition functions $C$ and being parametrized with $\epsilon_{start}=1$ and $\epsilon_{end}=0$ and $\epsilon_{start}=0$ and $\epsilon_{end}=1$ respectively. Furthermore, each is parametrized to have a curriculum length of $\textrm{\L}=1\,000$. Figure~\ref{fig:more_decay_schedules} shows examples of decreasing and increasing $C_{lin}$ with different $\textrm{\L}$.

\subsection{Curriculum on Iteration Scale}
$\epsilon_{i}$ prescribes a ratio of \gls*{tf} vs. \gls*{fr} steps for a given training iteration $i$. Based on $\epsilon$ that solely prescribes the amount of \gls*{tf} for an iteration, we can now develop micro curricula for distributing the \gls*{tf} and \gls*{fr} steps, eventually providing a \gls*{tf} decision $\Phi$ per training step. We propose two ways to distribute \gls*{tf} and \gls*{fr} steps within one training iteration: (1) probabilistic -- where $\epsilon$ is interpreted as the probability of being a \gls*{tf} step, and (2) deterministic -- where $\epsilon$ as a rate that determines the number of \gls*{tf} steps trained before moving to \gls*{fr} for the rest of the training sequence. For a probabilistic \gls*{cl}, we denote the \gls*{tf} decision $\Phi_\epsilon$, which is a discrete random variable that is drawn from a Bernoulli distribution: 
\begin{equation}
\Phi_\epsilon \sim \text{Bernoulli}(\epsilon).
\end{equation}

For a deterministic \gls*{cl}, $\Phi$ depends not only on $\epsilon$ but also on the current position $j$ within the predicted sequence of length $m$. Therefore, in this case we denote the \gls*{tf} decision $\Phi_{\epsilon,j}$ as:
\begin{equation}
\Phi_{\epsilon,j} = \begin{cases}
	1, & \text{if } \epsilon \geq \frac{j}{m} \\
	0, & \text{otherwise}.
\end{cases}
\end{equation}

\section{Evaluation}\label{sec:experiments}
To compare the training strategies described in Section~\ref{sec:strategies}, we evaluate each with varying parametrization on six different datasets. Our experiments aim to answer the following six research questions:
\begin{itemize}
	\item[\textbf{RQ1}] \textbf{Baseline teaching strategies.} How well and consistent do the current baseline strategies \gls*{fr} and 
	\gls*{tf} train a model for forecasting dynamical systems?
	\item[\textbf{RQ2}] \textbf{Curriculum learning strategies.} How do the different curriculum learning strategies perform in 
	comparison to the baseline strategies?
	\item[\textbf{RQ3}] \textbf{Training length.} How is training length influenced by the different teaching strategies?
	\item[\textbf{RQ4}] \textbf{Prediction stability.} How stable is a model's prediction performance over longer prediction 
	horizons when trained with the different strategies?
	\item[\textbf{RQ5}] \textbf{Curriculum parametrization.} How much does the curriculum's parametrization influence model 
	performance?
	\item[\textbf{RQ6}] \textbf{Iteration scale curriculum.} How do iteration scale curricula differ in resulting model 
	performance?
\end{itemize}

\subsection{Evaluated Curricula}
In total, we define eight strategies to evaluate in this study (cp. Fig.~\ref{fig:all_data_flows}). For comparison, we train the two baseline methods \gls*{tf} and \gls*{fr} that ``teach'' throughout the entire training or do not ``teach'' at all respectively. All other methods prescribe a teaching curriculum \gls*{cl} throughout the training and we distinguish these strategies along two dimensions: (1) the overall increasing (ITF), constant (CTF), or decreasing (DTF) trend in \gls*{tf} throughout the training curriculum and (2) the probabilistic (P) or deterministic (D) \gls*{tf} distribution within training steps.

\begin{table}[htb]
\caption{Curriculum strategy parameters used during the \textit{baseline}, \textit{exploratory}, and the \textit{essential} experiments}

\label{tab:hyperparameters}
\begin{tabularx}{\linewidth}{c p{4cm} | r X}
\toprule
                                                                                                      &                                                                                                   & \multicolumn{2}{c}{\textbf{Curriculum}} \\
                                                                                                      &  \textbf{Strategy}                                                                       & \textbf{parameter}                                           & \multicolumn{1}{l}{\textbf{evaluated values}} \\
\midrule
\multirow{6}{*}{\rotatebox[origin=c]{90}{\textit{baseline}}}      &                                                                                                   & C                                                                     &  --        \\
                                                                                                      &  FR                                                                                           & $\epsilon$                                                        & $0.0$  \\
                                                                                                      &                                                                                                   & $\textrm{\L}$                                                                 & -- \\
\cmidrule{3-4}
                                                                                                      &                                                                                                  & C                                                                      & -- \\
                                                                                                      & TF                                                                                             & $\epsilon$                                                        & $1.0$ \\
                                                                                                      &                                                                                                  & $\textrm{\L}$                                                                 & -- \\
\cmidrule{2-4}
                                                                                                      &                                                                                                  & C                                                                      & -- \\
\multirow{9}{*}{\rotatebox[origin=c]{90}{\textit{exploratory}}} & CL-CTF-P                                                                                &  $\epsilon$                                                        & $ \{0.25, 0.5, 0.75\}$ \\
                                                                                                      &                                                                                                  & $\textrm{\L}$                                                                  & -- \\
\cmidrule{3-4}
                                                                                                      & \multirow{3}{*}{CL-DTF-P, CL-DTF-D}                & C                                                                      & $\{\text{linear}, \text{inverse~sigmoid},$ $\text{exponential}\}$\\
                                                                                                      &                                                                                                  & $\epsilon_{start}\rightarrow\epsilon_{end}$     & $\{0.25, 0.5, 0.75, 1.0\} \rightarrow\ 0.0$ \\
                                                                                                      &                                                                                                  & $\textrm{\L}$                                                                  & $1000$ \\
\cmidrule{3-4}
                                                                                                      & \multirow{3}{*}{CL-ITF-P, CL-ITF-D}                    & C                                                                     & $\{\text{linear}, \text{inverse~sigmoid},$ $\text{exponential}\}$\\
                                                                                                      &                                                                                                   & $\epsilon_{start}\rightarrow\epsilon_{end}$    & $\{0.0, 0.25, 0.5, 0.75\} \rightarrow\ 1.0$ \\
                                                                                                      &                                                                                                   & $\textrm{\L}$                                                                & $1000$ \\
\cmidrule{2-4}
\multirow{4}{*}{\rotatebox[origin=c]{90}{\textit{essential}}}  & \multirow{4}{*}{\shortstack{CL-DTF-P, CL-DTF-D,\\CL-ITF-P, CL-ITF-D}} & C                                                                     & linear\\
                                                                                                      &                                                                                                  & $\epsilon_{start}\rightarrow\epsilon_{end}$   & $ \{0.0 \rightarrow 1.0, 1.0 \rightarrow 0.0\} $ \\
                                                                                                      &                                                                                                  & $\textrm{\L}$                                                                & $\{62, 125, 250, 500, 1\,000, 2\,000,$ $4\,000, 8\,000, 16\,000, 32\,000\}$ \\

\bottomrule
\end{tabularx}
\end{table}

\begin{table*}[htb]
		\caption{Details of the chaotic systems that were approximated to generate the data used for our experiments}

	\begin{tabularx}{\linewidth}{l p{6cm} l r X } 
	\toprule
	\textbf{System}         & \textbf{ODE/DDE} & \textbf{Parameters}                   & \bm{$d$} &  \textbf{LLE}               \\ 
	\toprule
	Mackey-Glass        & $\frac{dx}{dt} = \beta \frac{x_\tau}{1+x_\tau^n} - \gamma x$ 
	& 
	$\tau = 17$, $n = 10$, $\gamma = 0.1$  &   $1$          &  $0.006$  \\ 
	& \text{with }  $\gamma,\beta,n > 0$ &  $\beta = 0.2$,  $dt = 1.0$    & &   \\ 
	\midrule
	Thomas     & $\frac{dx}{dt} = sin(y) - b x$  & $b=0.1$,  $dt = 0.1$    & $3$    & $0.055$  \\ 
	& $\frac{dy}{dt} = sin(z) - b y$    &   &   &   \\
	& $\frac{dz}{dt} = sin(x) - b z$    &   &   & \\ 
	\midrule
	Rössler                            & $\frac{dx}{dt} = - (y + z)$ & $a = 0.2$, $b = 0.2$      & $3$  &  $0.069$  \\
	& $\frac{dy}{dt} = x + a y$&    $c = 5.7$, $dt = 0.12$        &                &    \\ 
	& $\frac{dz}{dt} = b + z (x - c)$ &                   &                &    \\ 
	\midrule
	Hyper Rössler              & $\frac{dx}{dt} = -y-z$ & $a=0.25$, $b=3$  & $4$ & $0.14$  \\ 
	& $\frac{dy}{dt} = x + ay + w$ &  $c=0.5$,  $d=0.05$     &    &      \\
	& $\frac{dz}{dt} = b + xz$ &    $dt=0.1$   &    &      \\
	& $\frac{dw}{dt} = -cz + dw$ &    &    &      \\
	\midrule
	Lorenz                                   & $\frac{dx}{dt} = -\sigma x + \sigma y$ & $\sigma = 10$, $\beta =  \frac{8}{3}$     &   $3$ &  $0.905$  \\ 
	&  $\frac{dy}{dt} = -x z + \rho x- y$ &   $\rho = 28$, $dt = 0.01$    &                   &     \\ 
	&  $\frac{dz}{dt} = x y - \beta z$ &         &                & \\ 
	\midrule
	Lorenz'96                              &   $\frac{dx_k}{dt}  = -x_{k-2} x_{k-1} + x_{k-1} x_{k+1} - x_k + F$ &  $F=8$, $dt = 0.05$ & $40$   & $1.67$  \\ 
	& for~$k = 1\dots d$~and~$x_{-1} = x_{d}$ &   & &  \\
	\bottomrule
\end{tabularx}
\label{tab:systems_details}
\end{table*}

\subsection{Parametrization of Training Curricula}
Table~\ref{tab:hyperparameters} shows all training-strategy-specific parameters and their values for the evaluated strategies. We subdivide our experiments into three sets: \textit{baseline}, \textit{exploratory}, and  \textit{essential} experiments. The baseline strategies \gls*{fr} and \gls*{tf} do not have any additional parameters. The CL-CTF-P strategy has the $\epsilon$ parameter configuring the strategy's \gls*{tf} ratio. The increasing and decreasing strategies CL-DTF-x and CL-ITF-x are configured by $\epsilon_{start}$ and $\epsilon_{end}$ referring to the initial and eventual amount of \gls*{tf} and the function $C$ transitioning between both. Additionally, $\textrm{\L}$ determines the number of training epochs in between $\epsilon_{start}$ and $\epsilon_{end}$. For the \textit{exploratory} experiments, we utilize a fix $\textrm{\L}=1\,000$, while for the \textit{essential} experiments, we evaluate all strategies solely using a linear transition $C_{linear}$ in the curriculum (cp. Eq.~\ref{eq:linear_curriculum}) with either $\epsilon_{start} = 0$ and $\epsilon_{end} = 1$ (increasing) or $\epsilon_{start} = 1$ and $\epsilon_{end} = 0$ (decreasing).

\subsection{Performance Metrics}
We use the \gls*{nrmse} and the $R^2$ metrics as well as two derived of those to evaluate model performance. \gls*{nrmse} is a normalized version of the \gls*{rmse} where smaller values indicate better prediction performance. For a single value of a sequence, \gls*{nrmse} is calculated as:
\begin{equation}
	\text{NRMSE}(y, \hat{y}) = \frac{\sqrt{\frac{1}{d} \cdot \sum_{j=1}^{d} (y_j - \hat{y}_j)^2}}{\sigma},
\end{equation}
where $y$ is a ground truth vector, $\hat{y}$ is the corresponding prediction, $\sigma$is the standard deviation across the whole dataset, and $d$ is the size of the vectors $y$ and $\hat{y}$. For model evaluation, we calculate the mean \gls*{nrmse} over all $m$ forecasted steps of a sequence. Additionally, we compute and report the \gls*{nrmse} only for the last $\big\lceil\frac{m}{10}\big\rceil$ forecasted steps of a sequence to specifically evaluate model performance at long prediction horizons.

The $R^2$ score lies in the range $(-\infty, 1]$ with higher values referring to better prediction performance.  A score of $0$ 
means that the prediction is as good as predicting the ground truth sequence's mean vector $\bar{y}$. The $R^2$ score is 
computed as:
\begin{equation}
	R^2 = 1 - \frac{\sum_{j=1}^{d} (y_j - \hat{y}_j)^2}{\sum_{j=1}^{d} (y_j - \bar{y}_j)^2}.
\end{equation}
We use the $R^2$ score to compute another metric $\text{LT} R^2>0.9$ measuring the number of \gls*{lt}s that a model can predict without the $R^2$ score dropping below a certain a threshold of $0.9$. Sangiorgio and Dercole \citep{sangiorgio2020robustness} proposed this metric while applying a less strict threshold of $0.7$.

\begin{table*}[htb]
	\caption{Results of the \textit{exploratory} tests with the best hyper-parameter configuration per strategy and system. The arrow besides each metric's column title indicates whether smaller ($\downarrow$) or larger ($\uparrow$) values are favored. The best result values per dataset are printed in bold and the best baseline NRMSEs are underlined. Together with each dataset, we put the corresponding LLE in parenthesis.}

	\label{tab:all_results_best}
	   \resizebox{\linewidth}{!}{
		\begin{tabular}{lc|cc|r|rrrr}
			\toprule
			&                               & \multicolumn{2}{c}{\textbf{Best performing curriculum}} & \textbf{Trained}                                  
			&                                                                 \multicolumn{4}{c}{\textbf{NRMSE over 1LT}}\\ 
			& \textbf{Strategy} & $\bm{C}$ &  $\bm{\epsilon}$                                 & \textbf{epochs}                                  & \textbf{absolut} $\bm{\downarrow}$ & \textbf{rel. impr.} $ \bm{\uparrow}$ & \textbf{@BL epoch} $ \bm{\downarrow} $   
			& \textbf{last 10\%} $\bm{\downarrow}$   \\
			\midrule
			\multirow{7}{*}{\rotatebox[origin=c]{90}{Thomas ($0.055$)}}
			& FR &               constant                 & $ 0.00 $                           &  $ 427 $   & $ \underline{0.03416} $ &  --                                       &   --                        &        $ 0.047222 $          \\ 
			& TF &               constant                 & $ 1.00 $                           &  $ 163 $      & $ 0.34545 $               &  --                                             &   --                        &        $ 0.607954 $           \\ \cmidrule{3-9}
			& CL-CTF-P &  constant                 & $ 0.25 $                           &   $ 450 $   & $ 0.05535 $              & $\bad{-62.03\%}$                &  $ 0.05675 $                       &        $ 0.082443 $      \\ 
			& CL-DTF-P &  inverse sigmoid      & $ 0.75 \searrow 0.00 $  &     $ 598 $ & $ 0.01858 $              &  $ \good{45.61\%}$              &  $ 0.02120 $                       &        $ 0.034325 $   \\ 
			& CL-DTF-D &   exponential           & $ 0.25 \searrow 0.00 $  &     $ 557 $  & $ 0.03229 $              & $ \good{5.47\%}$                &  $ 0.03792 $                       &        $ 0.039749 $   \\ 
			& CL-ITF-P &     exponential           & $ 0.00 \nearrow 1.00 $  & $ 620 $      &    $ 0.01403 $           & $ \good{58.93\%}$              &  $ \bm{0.02026} $              &        $ 0.026014 $      \\ 
			& CL-ITF-D &     exponential           & $ 0.25 \nearrow 1.00 $  &  $ 944 $     &    $\bm{0.01126}$    & $ \good{\bm{67.04\%}} $    &  $ 0.02179 $                        &        $ \bm{0.018571} $  \\ 
			\cmidrule{2-9}
			\multirow{7}{*}{\rotatebox[origin=c]{90}{Rössler ($0.069$)}}
			& FR &        constant                      & $ 0.00 $                             &  $ 3\,863  $   & $ \underline{0.00098} $ &  --                                  &  --                       &        $ 0.000930 $               \\
			& TF &        constant                      & $ 1.00 $                             &   $ 500 $        & $ 0.00743 $           &   --                                            &  --                       &        $ 0.016119 $                 \\ \cmidrule{3-9}
			& CL-CTF-P &       constant      &  $ 0.25 $                             &  $ 2\,081 $     & $ 0.00084 $          & $ \good{14.29\%}$              &  $ 0.00084 $                       &        $ 0.001333 $                \\
			& CL-DTF-P &          linear            &    $ 1.00 \searrow 0.00 $ &  $ 2\,751 $      & $ 0.00083 $          & $ \good{15.31\%}$               &  $ 0.00083 $                       &        $ 0.000931 $                 \\
			& CL-DTF-D &  inverse sigmoid    &    $ 0.25 \searrow 0.00 $ &  $ 4\,113 $    & $ 0.00064 $          & $ \good{34.69\%}$              &  $ 0.00066 $                       &        $ 0.000578 $                \\
			& CL-ITF-P &  inverse sigmoid      &    $ 0.00 \nearrow 1.00 $ &  $ 7\,194 $    & $ 0.00025 $          & $\good{74.49\%}$              &  $ 0.00034 $                       &        $ \bm{0.000358} $ \\ 
			& CL-ITF-D &          linear             &    $ 0.75 \nearrow 1.00 $ &  $ 5\,132 $       & $ \bm{0.00024} $ & $ \good{ \bm{75.51\%} }$   &  $ \bm{0.00031} $              &        $ 0.000390 $               \\
			\cmidrule{2-9}
			\multirow{7}{*}{\rotatebox[origin=c]{90}{Lorenz ($0.905$)}}
			& FR &        constant                      & $ 0.00$ &   $ 918 $ & $ 0.01209 $                                               &  --                                              & --                      &      $ 0.013166 $            \\ 
			& TF &        constant                      & $ 1.00$   &   $ 467 $ & $ \underline{0.00152} $                          &  --                                              & --                   &       $ 0.002244 $           \\ \cmidrule{3-9}
			& CL-CTF-P &        constant         &      $ 0.75$ &  $ 297 $ & $ 0.00167 $                                           & $\bad{-9.87\%}$                & $ 0.00167 $                       &         $ 0.002599 $           \\
			& CL-DTF-P &  inverse sigmoid    &    $ 0.75 \searrow 0.00 $ &   $ 522 $ & $ 0.00168 $                 & $\bad{-10.53\%}$               & $ 0.00162 $                      &        $ 0.002425 $           \\ 
			& CL-DTF-D &  inverse sigmoid    &    $ 1.00 \searrow 0.00 $ &   $ 204 $ & $ 0.00187 $                 & $\bad{-23.03\%}$               & $ 0.00187 $                      &        $ 0.002823 $           \\
			& CL-ITF-P &          linear              &    $ 0.00 \nearrow 1.00 $  &   $ 750 $ & $ 0.00149 $                 & $ \good{1.97\%}$               & $ 0.00217 $                      &        $ 0.002235 $           \\
			& CL-ITF-D &  inverse sigmoid      &    $ 0.75 \nearrow 1.00 $ &   $ 803 $ & $ \bm{0.00124} $       & $ \good{\bm{18.42\%}}  $    & $ \bm{0.00132} $            &        $ \bm{0.002084} $ \\
			\cmidrule{2-9}
			\multirow{7}{*}{\rotatebox[origin=c]{90}{Lorenz'96 ($1.67$)}}
			& FR &        constant                      &  $0.00$   &  $ 8\,125 $ &  $ 0.07273 $                                       &  --                                                & --                      &         $ 0.126511 $          \\
			& TF &        constant                      &   $1.00$      &  $ 4\,175 $ &  $ \underline{0.03805} $                &  --                                               & --                      &         $  0.075583  $        \\ \cmidrule{3-9}
			& CL-CTF-P &        constant         &    $ 0.50$ &  $ 2\,615 $ &  $ 0.07995 $                                      & $\bad{-110.12\%}$                  & $ 0.07995 $                      &         $ 0.140700   $         \\
			& CL-DTF-P &          linear            &    $ 0.75 \searrow 0.00 $ &   $ 939 $ & $  0.04654 $               & $\bad{-22.31\%}$                    & $ 0.04654 $                      &         $ 0.087228 $         \\
			& CL-DTF-D &          linear            &    $ 0.75 \searrow 0.00 $ &  $ 1\,875 $ &  $ 0.04381$              & $\bad{-15.14\%}$                    & $ 0.04381 $                      &         $ 0.081025 $          \\
			& CL-ITF-P &  inverse sigmoid      &    $ 0.25 \nearrow 1.00 $ &  $ 4\,787 $ & $ \bm{0.01854} $   & $ \good{\bm{51.27\%}} $        & $ \bm{0.02016} $            &         $ \bm{0.036651} $ \\
			& CL-ITF-D &  inverse sigmoid      &    $ 0.00 \nearrow 1.00 $ &  $ 3\,263 $ &  $ 0.02093 $          & $ \good{44.99\%}$                & $ 0.02196 $                       &         $ 0.040356 $          \\
			\bottomrule
		\end{tabular}
		       }
\end{table*}

\subsection{Evaluated Datasets}
\label{subsec:eval_datasets}
We use six different time series datasets that we built by approximating six commonly studied chaotic systems (cp. Tab.~\ref{tab:systems_details}), i.e., Mackey-Glass \citep{mackey1977oscillation}, Rössler \citep{rossler1976equation}, Thomas' cyclically symmetric attractor \citep{thomas1999deterministic}, Hyper Rössler \citep{rossler1979equation}, Lorenz \citep{lorenz1963deterministic} and Lorenz'96 \citep{lorenz1996predictability}. Table~\ref{tab:systems_details} shows the differential equations per system and how we parametrized them. These systems differ from each other in the number of dimensions $d$ and the degree of chaos as indicated by the largest lyapunov exponent in the LLE column of Tab.~\ref{tab:systems_details}. The LLEs are approximated values that were published independently in the past \citep{brown1991computing,sprott2007labyrinth,sano1985measurement,sandri1996numerical,hartl2003lyapunov,brajard2020combining}. We generate datasets by choosing an initial state vector of size $d$ and approximate $10\,000$ samples using the respective differential equations. We use SciPy package's implementation of the \gls*{lsode} \citep{radhakrishnan1993description} except for Mackey-Glass which we approximate through the Python module JiTCDDE implementing the \gls*{dde} integration method, as proposed by \citep{shampine2001solving}. Thereby, $dt$ defines the time difference between two sampled states per dataset and is shown in Table~\ref{tab:systems_details}. Where available, we chose $dt$ similar to previous studies aiming for comparability of results. We split each dataset into $80\%$ training samples and $10\%$ validation and testing samples respectively. All data is normalized following a $z$-transform.

\begin{figure}[htb]
	\centering
	\includegraphics[width=0.6\linewidth]{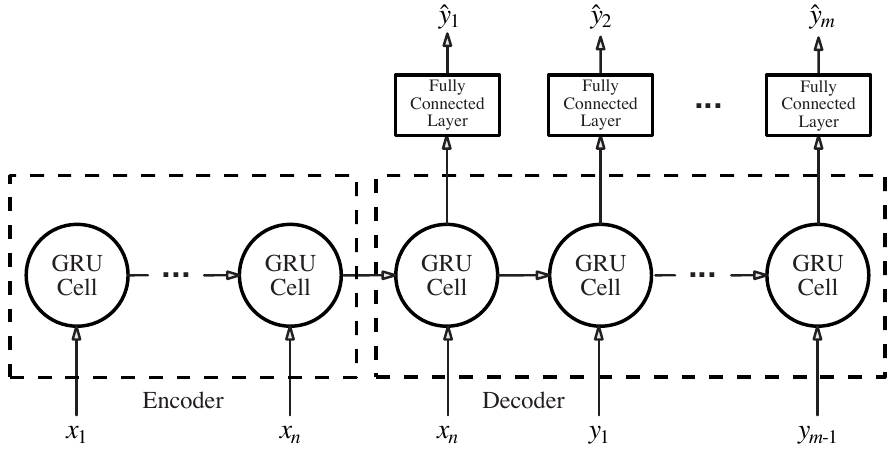}

	\caption{Structure of a simple encoder-decoder GRU used for training with teacher forcing}
	\label{fig:encoder_decoder_gru}
\end{figure}

\subsection{Training Procedure}
All evaluated models follow an encoder-decoder \gls*{gru} architecture with an additional fully connected layer after the decoder (cp. Fig.~\ref{fig:encoder_decoder_gru}). We performed a full grid search for the hyper-parameters learning rate, batch size, learning rate reduction factor, loss plateau, input length $n$ and hidden state size to determine suitable configurations for the experiments. Based on this optimization, we used the Adam \citep{kingma2015adam} optimizer with a batch size of $128$ and apply \gls*{rlrop} with an initial learning rate of $1e^{-3}$ and a reduction factor of $0.6$, i.e., 40\% learning rate reduction, given a loss plateau of $10$ epochs for all datasets except Lorenz'96, where we use a reduction factor of $0.9$ and a $20$ epoch plateau respectively. Furthermore, we found an input length of $n=150$ steps and a \textit{hidden state size} of $256$ to be most suitable. We use early stopping with a \textit{patience} of $100$ epochs and a \textit{minimum improvement threshold} of $1\%$ to ensure the convergence of the model while preventing from overfitting. We train all models with a dataset-specific prediction length $m$ defined as:

\begin{equation}
	m = \bigg\lceil\frac{\text{LT}}{dt}\bigg\rceil = \bigg\lceil\frac{1}{dt \cdot \text{LLE}}\bigg\rceil.
\end{equation}

The reason being that we aim to train for the same forecasting horizon that we mainly evaluate a trained model with. We adapt this horizon to the dataset's \gls*{lt}, thereby aiming for performance measures that are comparable across datasets.

\begin{figure}[htb]
	\centering
	\begin{subfigure}{0.49\linewidth}
		\includegraphics[clip,width=\linewidth]{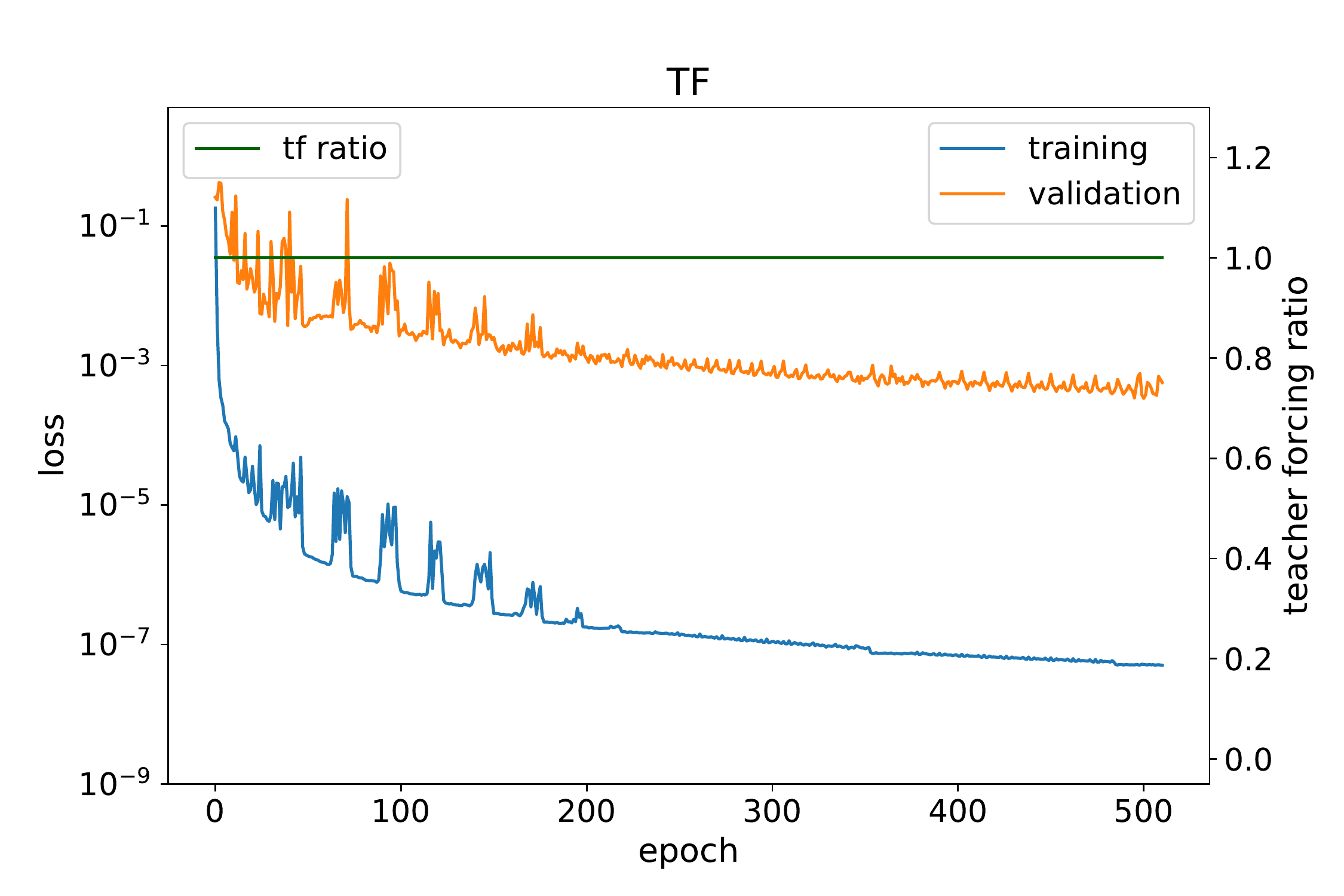}
		\caption{}
	\end{subfigure}
	\begin{subfigure}{0.49\linewidth}
		\includegraphics[clip,width=\linewidth]{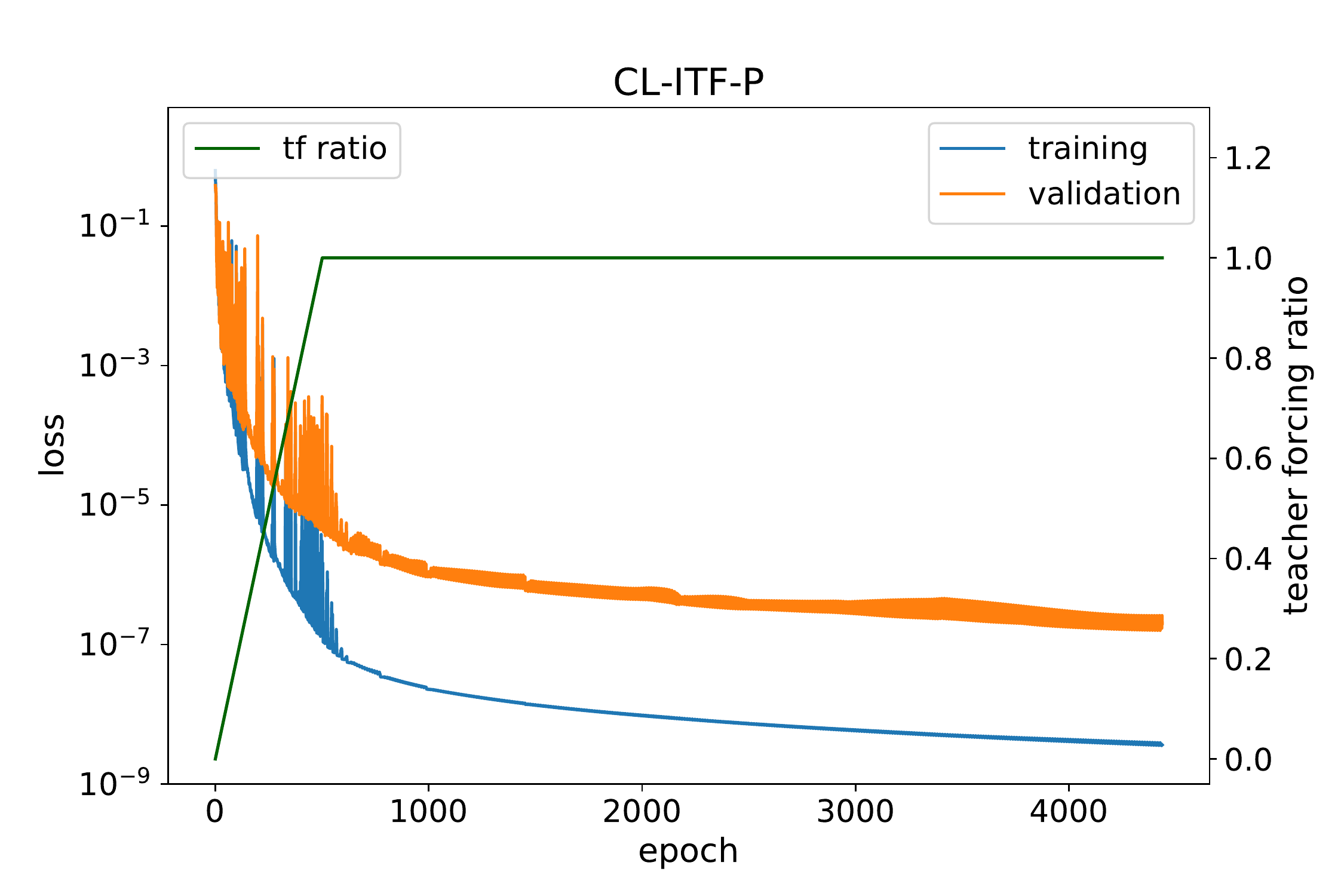}
		\caption{}
	\end{subfigure}

	\caption{Loss and teacher forcing ratio curves for Rössler}
	\label{fig:roessler_loss}
\end{figure}

We provide plots of the training and validation loss curves of the final parametrization per strategy and dataset in Appendix~\ref{apx:loss_curves}. Based on these loss curves, we observe for ITF in contrast to DTF strategies that the training loss tends to move away from the validation loss faster. This is explainable by the fact that with increasing training time, ITF strategies deliver an increasing amount of \gls*{tf} inputs, counteracting the accumulation of error along the forecasted sequence and therefore further reducing training loss. For DTF strategies we observe an opposing behavior. Regarding training iterations, we observe that ITF strategies typically train for a larger number of epochs. Following the \gls*{tf} ratio curve ($\epsilon$) we see that even when the increasing curriculum already arrived at the final $\epsilon_e = 1.0$, which means that it is using solely \gls*{tf} training from there on, it can improve for many more epochs than by simply using \gls*{tf} from the beginning. Since the termination of the training is determined by the early stopping criterion this shows that ITF may facilitate a longer and (regarding the validation loss) typically more successful training process compared to DTF and baseline strategies. We show the corresponding curves exemplary for the Rössler training in Figure~\ref{fig:roessler_loss}.

\begin{table*}[htb]
	\caption{Best curriculum length per strategy and system for all six datasets. The arrow besides a metric's column title indicates whether smaller ($\downarrow$) or larger ($\uparrow$) values are favored. The best result values per dataset are printed in bold and the best baseline NRMSEs are underlined. Together with each dataset, we put the corresponding LLE in parenthesis.}
	\label{tab:all_results_best_curriculum_length}
	  \resizebox{\linewidth}{!}{
		\begin{tabular}{lc | cc | r | rrrrr}
			\toprule
			&                               &   \multicolumn{2}{c}{\textbf{Best performing curriculum}}                    & \textbf{Trained}  &                                                               \multicolumn{4}{c}{\textbf{NRMSE over 1LT}}          & \textbf{\#LT with}                                             \\
			& \textbf{Strategy} &  $\bm{\epsilon}$                              & $\mathbf{\L}$                                & \textbf{epochs}          &  \textbf{absolut}$\bm{\downarrow}$ & \textbf{rel. impr.}$\bm{\uparrow}$ & \textbf{@BL epoch}$\bm{\downarrow}$ & \textbf{last 10\%}$\bm{\downarrow}$ &   $\bm{R^2 > 0.9}\bm{\uparrow}$ \\
			\midrule
			\multirow{7}{*}{\rotatebox[origin=c]{90}{Mackey Glass ($0.006$)}}
			&           FR       &  $ 0.00 $                          &                             -- &  $  4\,713 $   & $  \underline{0.00391}  $         &  --                                     & $ 0.00391 $ & $ 0.004101 $          & $ 4.50 $        \\
			&           TF       &  $ 1.00 $                           &                            -- &  $  44  $        &  $ 0.09535  $                            &  --                                     & $ 0.09535 $ & $ 0.171945 $           & $ 1.64 $         \\ 
			\cmidrule{3-10}
			& CL-CTF-P    &  $ 0.25  $                          &                            -- &  $  2\,918 $   & $  0.00632 $                             & $ \bad{-61.64\%} $        & $ 0.00632 $ & $ 0.006544 $        & $ 4.51 $         \\
			& CL-DTF-P    &  $ 1.00 \searrow 0.00 $    &             $ 2\,000 $  &  $  3\,733  $ & $  0.00215  $                             & $ \good{45.01\%} $         & $ 0.00215 $ & $ 0.003010 $       & $ 4.95 $         \\
			& CL-DTF-D    &  $   1.00 \searrow 0.00 $  &             $ 1\,000  $ &  $   431  $     &  $ 0.00585  $                            & $\bad{-49.62\%} $         & $ 0.00585 $ & $ 0.011022 $            & $ 3.91 $          \\
			& CL-ITF-P      &  $ 0.00 \nearrow 1.00  $   &               $  500  $ &  $  1\,566 $   &  $ \bm{0.00104}  $                    & $\good{\bm{73.40\%}} $        & $ \bm{0.00104} $ & $ \bm{0.001793} $  & $ \bm{5.18} $ \\
			& CL-ITF-D      &  $  0.00 \nearrow 1.00  $ &               $  500  $ &   $ 1\,808  $   &  $ 0.00211 $                              & $\good{46.03\%} $        & $ 0.00211 $ & $ 0.003032 $            & $ 4.99 $          \\
			\cmidrule{2-10}
			\multirow{7}{*}{\rotatebox[origin=c]{90}{Thomas ($0.055$)}}
			&           FR       &  $ 0.00 $                                 &                 -- &   $  427  $ &  $ \underline{0.03416}  $               & --                                         & $ 0.03416 $           & $ 0.047222 $            & $ 2.04 $           \\
			& TF              &   $ 1.00 $                      &  --              &  $ 163 $      & $ 0.34545 $               &  --                                             &   $ 0.34545  $        &        $ 0.607954 $  &   $ 1.73 $        \\ \cmidrule{3-10}
			& CL-CTF-P    &  $  0.25 $                                &                -- &   $  450 $  &  $ 0.05535  $                                    & $\bad{-62.03\%} $           & $ 0.05675 $           & $ 0.082443 $           & $ 1.53 $            \\
			& CL-DTF-P    &  $   1.00 \searrow 0.00  $  &    $ 1\,000 $  &    $ 356  $ & $  0.05084  $                                    & $\bad{-48.83\%} $            & $ 0.05084 $          & $ 0.105585 $            & $ 2.13 $             \\
			& CL-DTF-D    &  $   1.00 \searrow 0.00 $  &     $ 1\,000  $ &  $   326  $ & $  0.10712 $                                      & $\bad{-213.58\%} $            & $ 0.10712 $           & $ 0.206923 $           & $ 1.53 $            \\
			& CL-ITF-P      &  $  0.00 \nearrow 1.00 $  &     $ 500  $ & $    677  $ &  $ \bm{0.00930} $                                & $\good{\bm{72.78\%}} $   & $ \bm{0.01645} $ & $ \bm{0.016729} $   & $ \bm{3.99} $  \\
			& CL-ITF-D     &  $   0.00 \nearrow 1.00 $  &    $ 500 $  &  $   649 $  &  $ 0.01819  $                                         & $\good{ 46.75\%} $            & $ 0.03934 $         & $ 0.030589 $           & $ 2.05 $           \\
			\cmidrule{2-10}
			\multirow{7}{*}{\rotatebox[origin=c]{90}{Rössler ($0.069$)}}
			&           FR       &  $ 0.00 $                                   &                 -- &   $ 3\,863  $ & $  \underline{0.00098} $        &  --                                       & $ 0.00098 $           & $ 0.000930 $        & $ 9.46 $             \\
			&           TF       &  $ 1.00 $                                    &                 -- &  $   500  $ & $  0.00743 $                               &  --                                        & $ 0.00743 $           & $ 0.016119 $             & $ 4.75 $             \\  \cmidrule{3-10}
			& CL-CTF-P    &  $ 0.25  $                              &                -- &  $  2\,081  $ &  $ 0.00084 $                                 & $\good{14.29\%} $          & $ 0.00084 $          & $ 0.001333 $             & $ 7.51 $               \\
			& CL-DTF-P    &  $ 1.00 \searrow 0.00 $  &              $  1\,000  $ &  $  2\,751  $ & $  0.00083  $                       & $\good{15.31\%} $          & $ 0.00083 $           & $ 0.000931 $             & $ 8.46 $              \\
			& CL-DTF-D    &  $ 1.00 \searrow 0.00 $ &             $    125  $ &  $  4\,879  $ & $  0.00100 $                            & $\bad{-2.04\%} $            & $ 0.00116 $           & $ 0.000947 $            & $ 9.28 $               \\
			& CL-ITF-P      &  $ 0.00\nearrow 1.00$      &               $  500 $  & $   4\,523  $ &  $ \bm{0.00019}  $             & $\good{\bm{80.61\%}} $  & $ \bm{0.00022} $ & $ \bm{0.000303} $   & $ \bm{10.23} $ \\
			& CL-ITF-D      &  $ 0.00\nearrow 1.00  $    &             $   4\,000  $ &   $ 7\,267 $  &  $ 0.00027 $                    & $\good{72.24\%} $           & $ 0.00051 $          & $ 0.000368 $        & $ 9.41 $             \\ 
			\cmidrule{2-10}
			\multirow{7}{*}{\rotatebox[origin=c]{90}{Hyper Rössler ($0.14$)}}
			& FR                  &    $ 1.00 $                          &                              -- &  $ 6\,461 $     & $ 0.00599 $                        &   --                                         &   $ 0.00599 $ &      $ 0.007011$                & $ 6.57 $            \\
			& TF                  &    $ 0.00 $                          &                              -- &  $ 2\,788 $     & $ \underline{0.00435} $   &   --                                         &  $ 0.00762 $ &      $ 0.011194$                 & $ 5.24 $            \\ \cmidrule{3-10} 
			& CL-CTF-P     &    $ 0.25 $                          &                              -- &  $ 2\,909 $     & $ 0.01450 $                       &  $ \bad{-233.33\%} $          &  $ 0.01450 $ &      $ 0.015944$                & $ 5.21 $            \\
			& CL-DTF-P     &    $ 1.00 \searrow 0.00 $ &              $ 2\,000$    &  $ 3\,773 $        & $ 0.00560 $                     &  $\bad{28.74\%} $               &  $ 0.00560 $ &      $ 0.007052$                & $ 6.32 $            \\ 
			& CL-DTF-D     &    $ 1.00 \searrow 0.00 $ &             $ 16\,000 $  &  $ 1\,793 $        & $ 0.00490 $                     &  $\bad{12.64\%} $             &  $ 0.00490 $ &      $ 0.007471$                 & $ 6.30 $            \\ 
			& CL-ITF-P       &    $ 0.00 \nearrow 1.00 $ &               $ 125 $      &  $ 2\,802 $        & $ 0.00366 $                     &  $\good{15.86\%} $            &  $ 0.00366 $ &      $ 0.005802$                & $ 6.50 $            \\
			& CL-ITF-D       &    $ 0.00 \nearrow 1.00 $ &               $ 250 $     &  $ 3\,317 $        & $ \bm{0.00326} $             &  $\good{\bm{25.06\%}} $   &  $ \bm{0.00326} $ &       $ \bm{0.004639} $    & $ \bm{6.72} $    \\
			\cmidrule{2-10}
			\multirow{7}{*}{\rotatebox[origin=c]{90}{Lorenz ($0.905$)}}
			&           FR     &  $ 0.00 $                                      &                 -- &   $  918  $ &  $ 0.01209  $                                & --                                               & $ 0.01319 $           &  $ 0.013166  $            & $ 3.31 $        \\
			&           TF     &  $ 1.00 $                                      &                 -- &  $   467 $  &  $ \underline{0.00152} $             & --                                               & $ 0.00152 $           &  $ 0.002244 $            & $ 6.72 $      \\  \cmidrule{3-10}
			& CL-CTF-P  &  $   0.75 $                                    &                -- &   $  297 $  & $  0.00167 $                                  &  $\bad{-9.87\%} $                 & $ 0.00167 $          &  $ 0.002599 $            & $ 6.43 $         \\
			& CL-DTF-P  &  $    1.00 \searrow 0.00  $     &            $    4\,000 $  &   $  450 $  &  $ 0.00124  $                      &  $\good{18.42\%} $                & $ \bm{0.00124} $ &  $ 0.001925  $            & $ 6.64 $        \\
			& CL-DTF-D  &  $     1.00 \searrow 0.00   $    &           $    16\,000 $  &   $  587 $  &  $ 0.00111 $                       &  $\good{26.97\%} $               & $ 0.00127 $           &  $ 0.001650  $           & $ 6.53 $        \\
			& CL-ITF-P    &  $   0.00 \nearrow 1.00 $       &             $     250 $ &  $  1\,137 $  &  $ \bm{0.00060} $              &  $\good{\bm{60.53\%}} $      & $ \bm{0.00124} $ &  $ \bm{0.000883}  $ & $ \bm{7.19} $ \\
			& CL-ITF-D    &  $   0.00 \nearrow 1.00  $      &           $      250 $ &   $  578  $ &  $ 0.00135 $                           &  $\good{11.18\%} $                 & $ 0.00189 $          &  $ 0.001725 $             & $ 4.33 $           \\
			\cmidrule{2-10}
			\multirow{7}{*}{\rotatebox[origin=c]{90}{Lorenz'96 ($1.67$)}}
			&           FR      &  $ 0.00 $                                     &      -- &   $ 8\,125 $  & $  0.07273 $                                         &  --                                            & $ 0.08362 $           & $ 0.126511 $                        & $ 2.34 $         \\
			&           TF      &  $ 1.00 $                                     &       -- &  $  4\,175  $ & $  \underline{0.03805}  $                   &  --                                            & $ 0.03805 $           & $ 0.075583 $                   & $ 3.01 $      \\   \cmidrule{3-10}
			& CL-CTF-P   &  $ 0.50 $                                    &        -- &  $  2\,615 $  &  $ 0.07995 $                                        &  $\bad{-110.12\%} $              & $ 0.07995 $           & $ 0.140700 $                    & $ 2.25 $       \\
			& CL-DTF-P   &  $  1.00 \searrow 0.00  $      &        $        1\,000 $  &   $  983 $  &  $ 0.05278 $                        & $\bad{-38.71\%}  $              & $ 0.05278 $           & $ 0.098130 $                   & $ 2.67 $       \\
			& CL-DTF-D   &  $  1.00 \searrow 0.00  $      &       $         1\,000 $  &  $  4\,083 $  &  $ 0.07119  $                    & $\bad{-87.10\%} $               & $ 0.07119 $            & $ 0.126636 $                   & $ 2.34 $       \\ 
			& CL-ITF-P     &  $  0.00 \nearrow 1.00  $      &         $        250 $  &   $ 3\,886 $  & $  0.01680 $                       & $\good{55.85\%} $             & $ 0.01680 $           & $ 0.032439 $                   & $ 4.01 $  \\
			& CL-ITF-D     &  $   0.00 \nearrow 1.00 $        &         $        250 $  &  $  3\,379 $  &  $ \bm{0.01628} $            & $\good{\bm{57.21\%}} $    & $ \bm{0.01628} $ & $ \bm{0.031464} $          & $ \bm{4.18} $  \\
			\bottomrule
		\end{tabular}
	   }
\end{table*}

\subsection{Results}\label{subsec:results}
\footnote{A reproduction package for the experiments is available on \href{https://github.com/phit3/flipped_classroom}{Github}: https://github.com/phit3/flipped\_classroom. The datasets used in this paper are published on \href{https://doi.org/10.7910/DVN/YEIZDT}{Dataverse}: https://doi.org/10.7910/DVN/YEIZDT.}Table~\ref{tab:all_results_best} shows results for the \textit{exploratory} experiments. Per evaluated strategy and dataset, the table reports resulting model performance in terms of \gls*{nrmse}. We only report that curriculum configuration in terms of transition function $C$ and $\epsilon$ schedule that yields the highest \gls*{nrmse} per strategy and dataset. Each model was used to predict a dataset-spetefic horizon of $1$ \gls*{lt}. The best result per dataset and performance metric is highlighted in bold. First, we study the baseline strategies \gls*{fr} and \gls*{tf} and observe that for two datasets, i.e., Thomas and Rössler, the \gls*{fr} strategy outperforms the \gls*{tf} baseline, while \gls*{tf} outperforms \gls*{fr} for the other two. We select the one that performs best per dataset to measure relative improvement or deterioration gained by training with the respective curriculum learning strategy (cp. column ``NMRSE rel. impr.''). We observe that across all datasets and performance metrics, the CL-ITF-P and CL-ITF-D strategies yield the best and second best-performing model with a relative improvement of $1.97$ -- $80.61\%$ over the best performing baseline strategy. The other curriculum learning strategies perform less consistent across the datasets. The CL-DTF-x strategies yield an improved NRMSE for half of the datasets, while the constant CL-CTF-P only yields an improvement for the Thomas attractor. We separately report the \gls*{nrmse} of the last $10\%$ predicted values of $1$ \gls*{lt} test horizon per dataset to assess how robust a prediction is over time (cp. column ``NRMSE last 10\%''). We observe that the CL-ITF-P and CL-ITF-D strategies also reach the best performance in terms of this metric, meaning that they yield the most robust models.  We further observe a diverse set of curriculum configurations yielding the best performing model per strategy and dataset. That means that all available transition functions, i.e., linear, inverse sigmoid, and exponential, have been discovered as best choice for at least one of the trained models. Further, we observe all evaluated $\epsilon$ as best choice for the CL-CTF-P strategy and one of the datasets respectively. Similarly, the best-performing initial $\epsilon$ for the increasing and decreasing transitions per dataset spans all evaluated values except for $0.5$. The table also reports the number of training iterations until reaching the early stopping criterion (cp. column ``training epochs''). We observe that the two baseline strategies utilize strongly differing numbers of iterations across all datasets. For the Thomas and the Rössler attractor, the \gls*{tf} strategy does not allow for proper model convergence, being characterized by a low number of iterations and a high NRMSE compared to the other strategies. Among the curriculum teaching strategies across all datasets, the strategies with increasing \gls*{tf} ratio CL-ITF-x utilize the most training iterations. These CL-ITF-x strategies also utilize more training iterations than the better-performing baseline strategy across all datasets. To better understand whether the longer training is the sole reason for the higher performance of the CL-ITF-x trained models, we additionally report the performance in terms of NRMSE of all curriculum-trained models after the same number of training iterations as the better performing baseline model, i.e., after $ 427 $ epochs for Thomas, after $ 3\,863 $ epochs for Rössler, after $ 467 $ epochs for Lorenz, and after $ 4\,175 $ epochs for Lorenz'96 (cp. column ``Performance @BL epochs''). We observe across all datasets that the best-performing teaching strategy still remains CL-ITF-P or CL-ITF-D. In conclusion, the \textit{exploratory} experiments demonstrated  that a well-parametrized CL-ITF-x strategy yields a $18.42$ -- $75.51\%$ performance increase across the evaluated datasets.

\begin{figure*}[htb]
	\centering
	\begin{subfigure}{0.4\linewidth}
			\includegraphics[trim=2 2 40 33,clip,width=\linewidth]{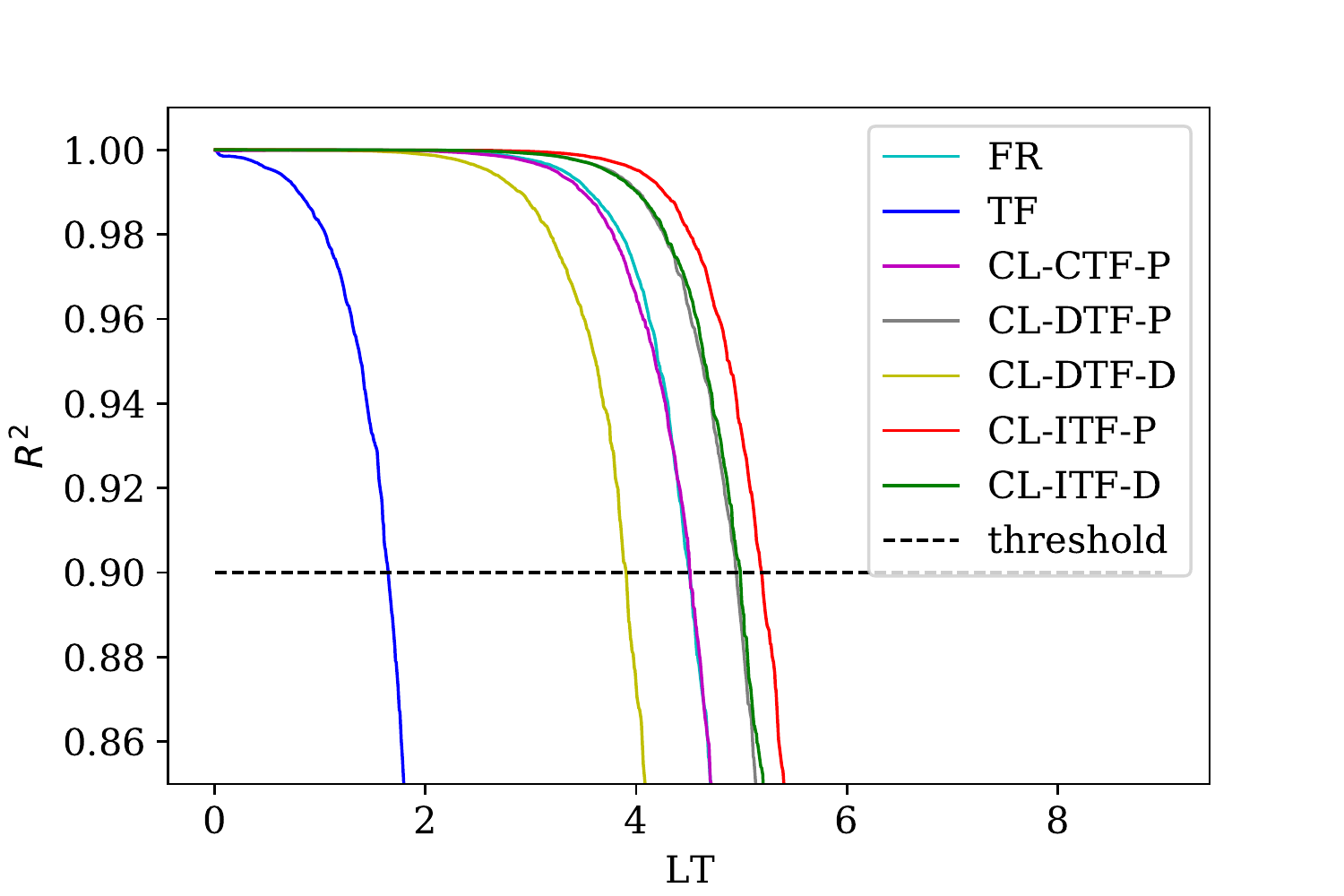}
		\caption{Mackey-Glass}
	\end{subfigure}\hspace{24pt}
	\begin{subfigure}{0.4\linewidth}
		\includegraphics[trim=2 2 40 33,clip,width=\linewidth]{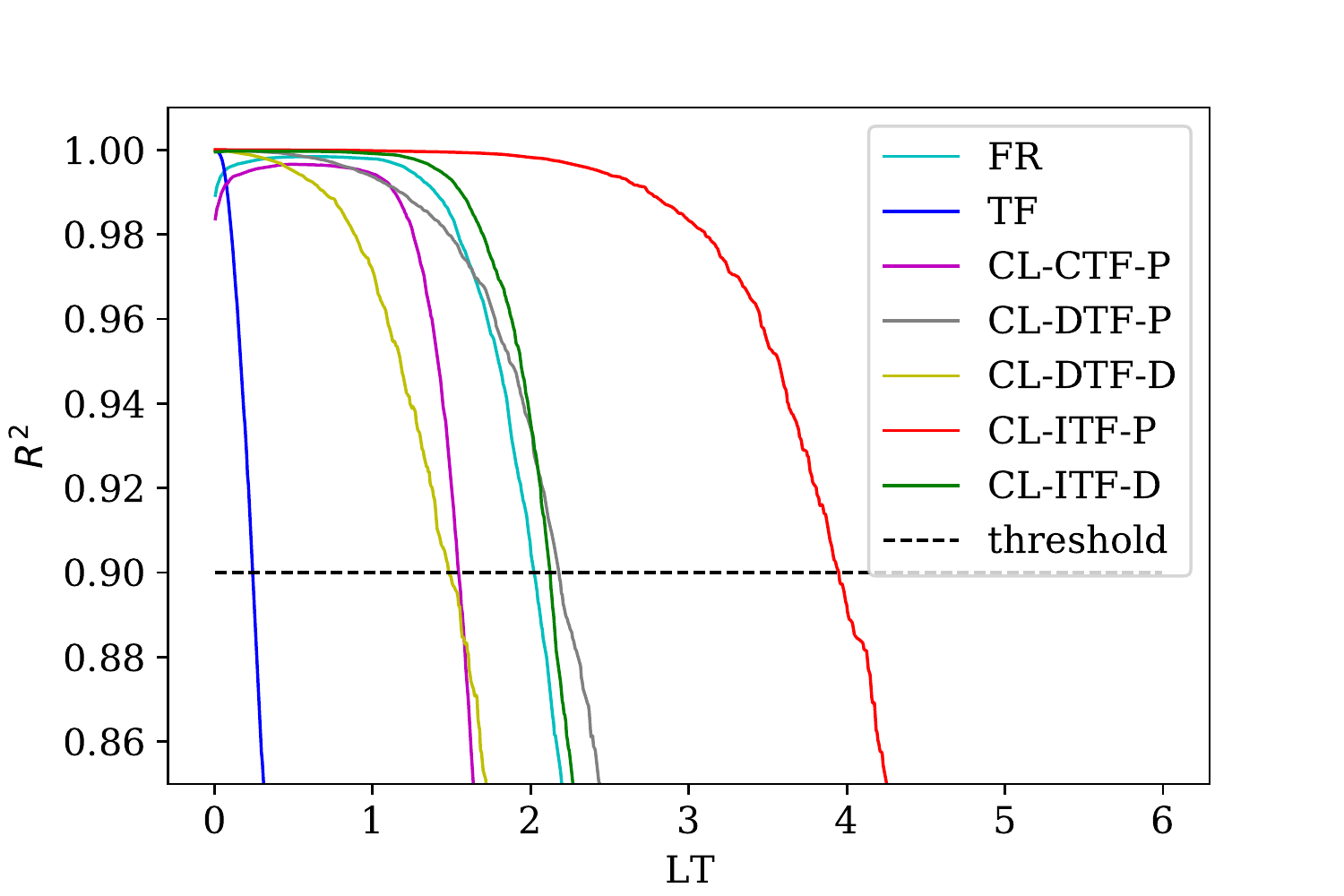}
		\caption{Thomas}
	\end{subfigure}
	
	\begin{subfigure}{0.4\linewidth}
		\includegraphics[trim=2 2 40 33,clip,width=\linewidth]{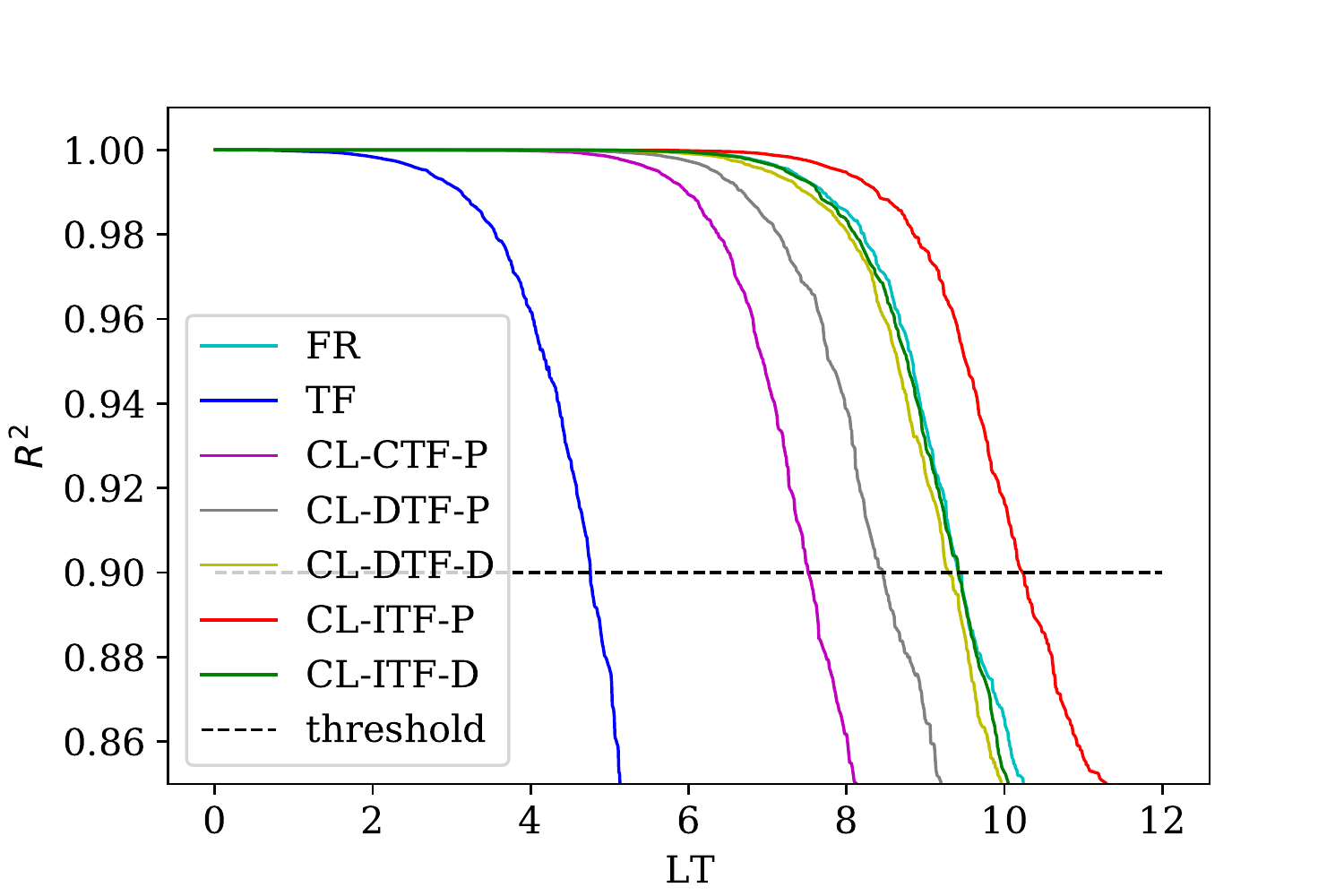}
		\caption{Rössler}
	\end{subfigure}\hspace{24pt}
	\begin{subfigure}{0.4\linewidth}
		\includegraphics[trim=2 2 40 33,clip,width=\linewidth]{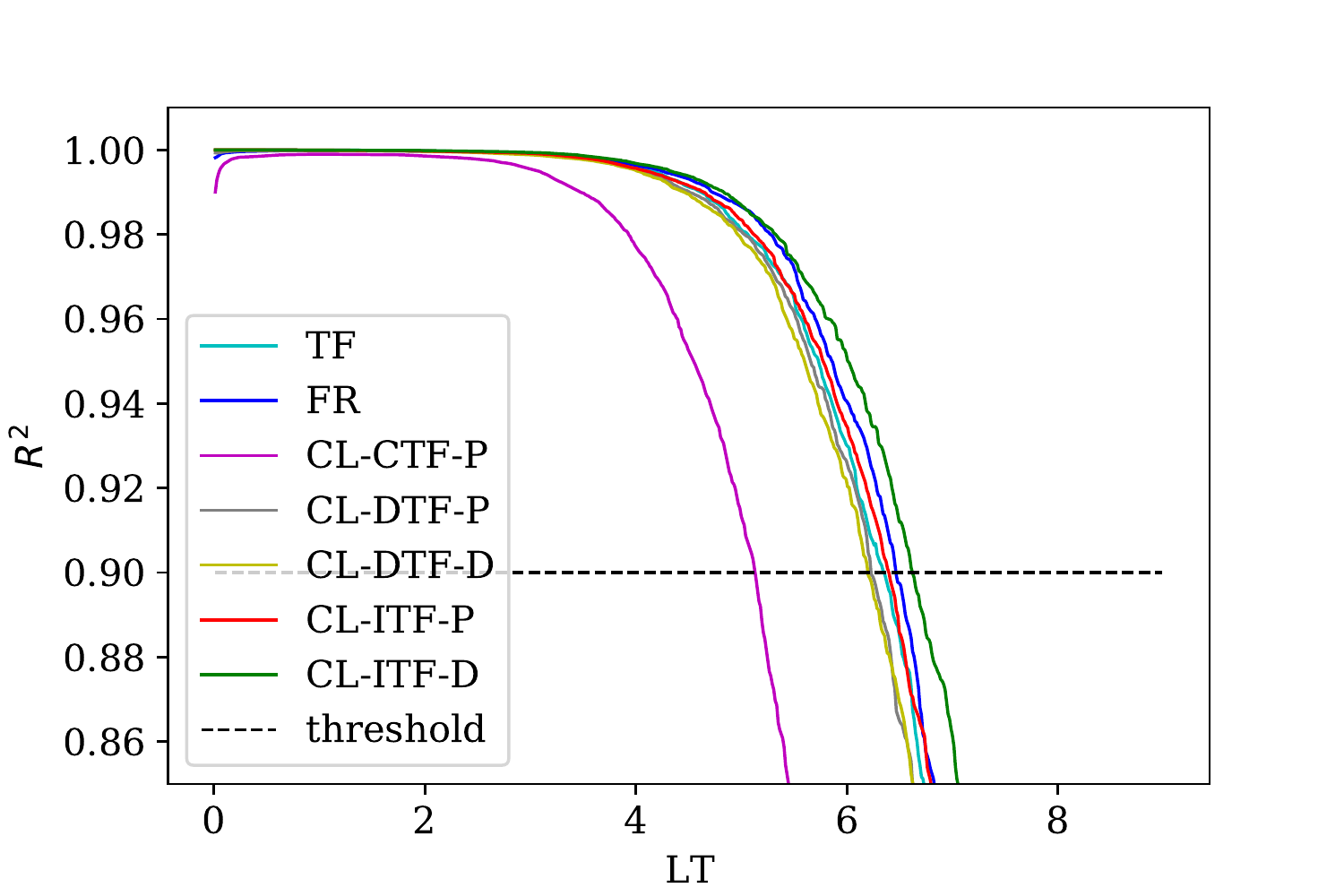}
		\caption{Hyper Rössler}
	\end{subfigure}
	
	\begin{subfigure}{0.4\linewidth}
		\includegraphics[trim=2 2 40 33,clip,width=\linewidth]{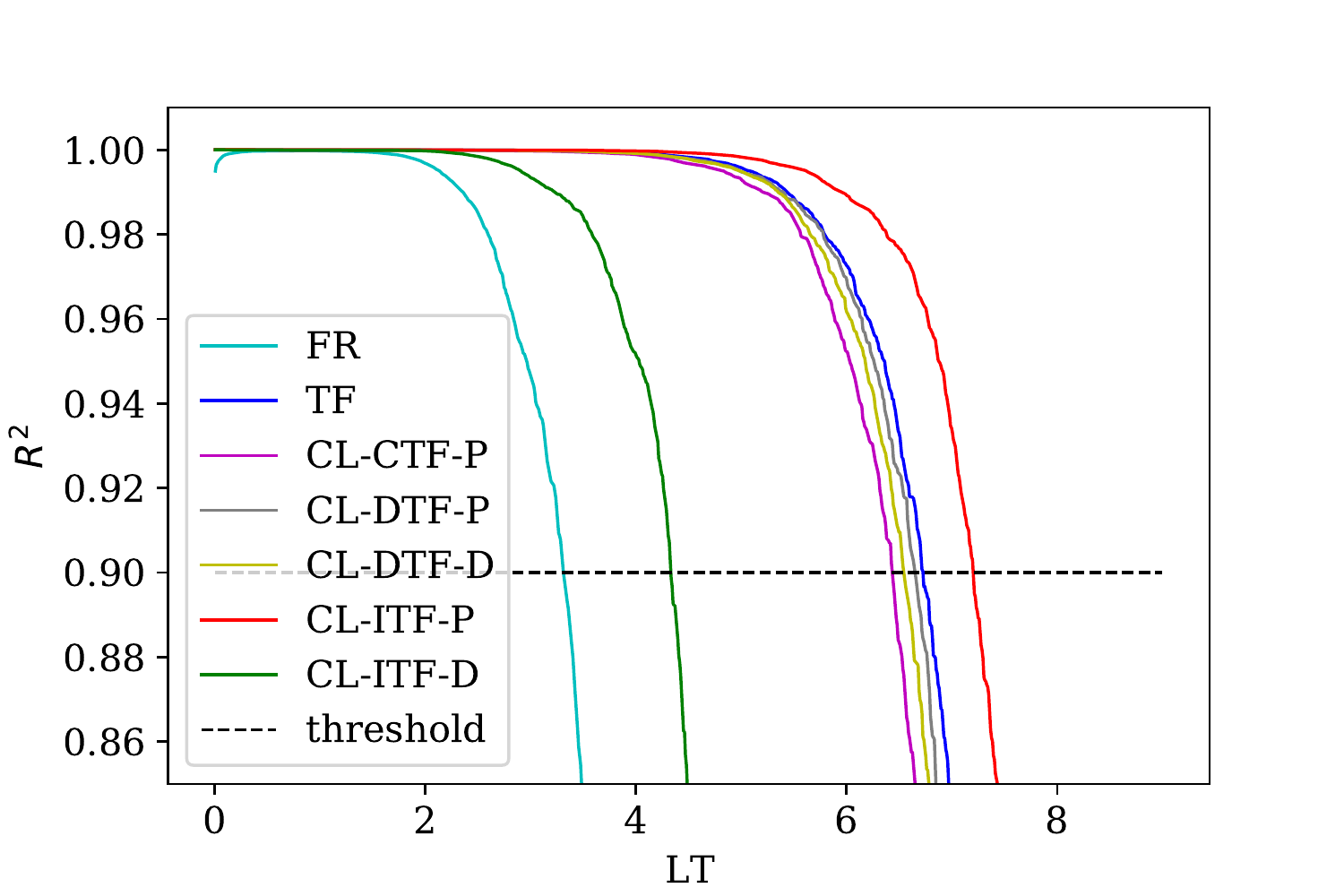}
		\caption{Lorenz}
	\end{subfigure}\hspace{24pt}
	\begin{subfigure}{0.4\linewidth}
		\includegraphics[trim=2 2 40 33,clip,width=\linewidth]{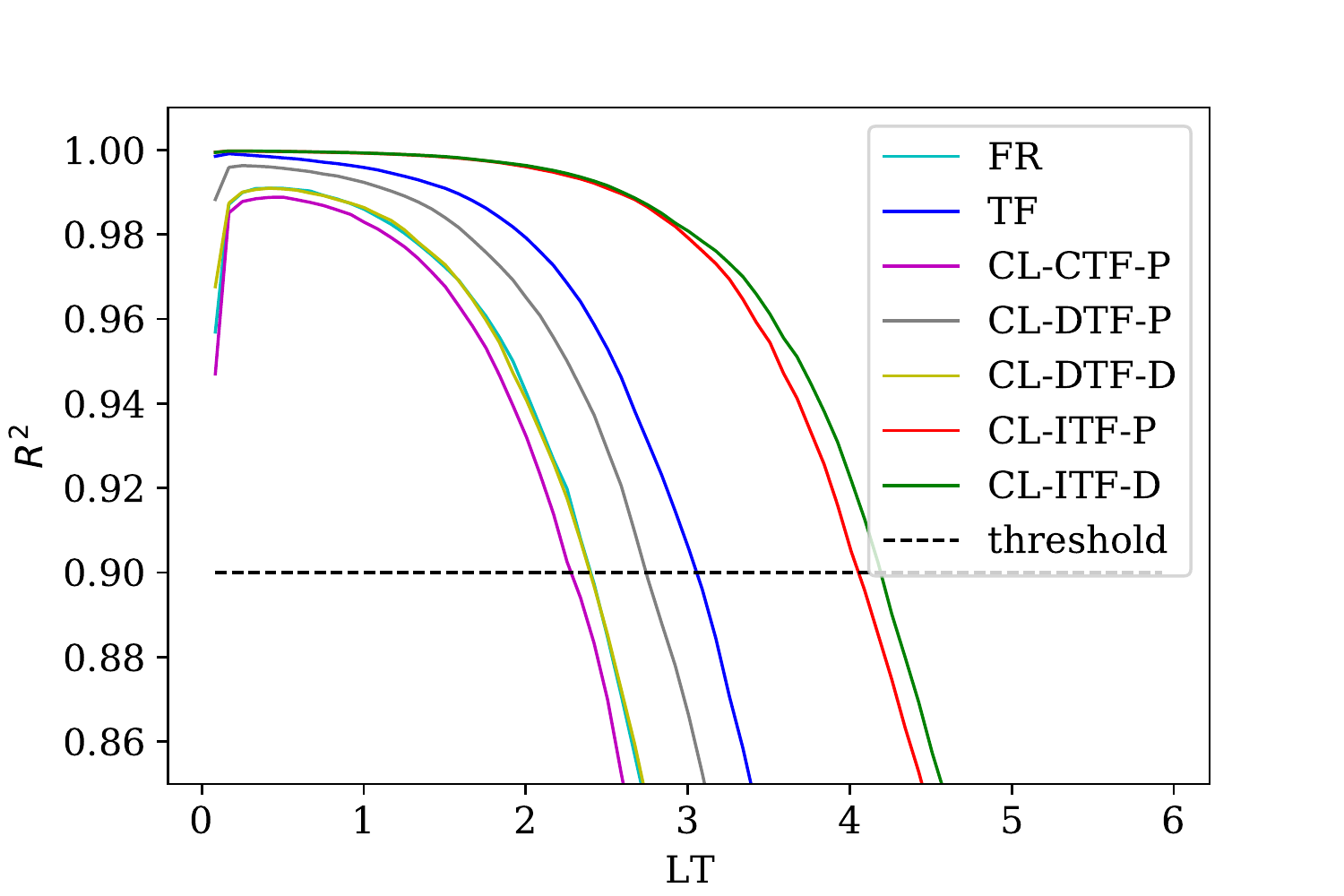}
		\caption{Lorenz' 96}
	\end{subfigure}

	\caption{$R^2$ score over multiple LTs for the six studied datasets using eight different training strategies}
	\label{fig:multiple_lt_cut}
\end{figure*}

However, this improvement comes at the cost of an intensive parameter optimization of the respective curriculum. Therefore, we run a second series of \textit{essential} experiments in which we simplify the parametrization of the curriculum by utilizing a linear transition from either $0.0 \rightarrow1.0$ (CL-ITF-x) or $1.0 \rightarrow0.0$ (CL-DTF-x), that is solely parametrized by the length of this transition in terms of training epochs $\textrm{\L}$. Table~\ref{tab:all_results_best_curriculum_length} reports results in terms of the previously introduced performance metrics again, measured over a prediction horizon of $1$ \gls*{lt} and across the same teaching strategies for six datasets, including those studied for the \textit{exploratory} experiments. Since the changes of the \textit{essential} over the \textit{exploratory} experiments solely affect teaching strategies with a training-iteration-dependent curriculum, they have no effect on the baseline strategies \gls*{fr} and \gls*{tf} as well as the constant curriculum CL-CTF-P, which we still report in Table~\ref{tab:all_results_best_curriculum_length} for direct comparison. Overall, we observe that CL-ITF-P outperforms all other strategies for four out of six datasets, while it performs second best for the remaining two datasets where the deterministic version CL-ITF-D performs best. These strategies yield relative improvements ranging from $25.06$ -- $80.61\%$ and are, thus, even higher than those observed for the \textit{exploratory} experiments. Beyond that we observe that for all datasets treated in both experimental sets, the training curricula used in the \textit{essential} experiments yield better performing models. For three out of four datasets, the training even requires substantially less training iterations than in the explorative experiments. Additionally, we report in column ``\#LT with $R^{2}>0.9$'' the prediction horizon in terms of \gls*{lt} that a trained model can predict while constantly maintaining an $R^{2}>0.9$. We observe that the ranking between the different strategies mostly remains the same as those observed when predicting 1 LT. That means that the best-performing strategy  consistently remains the same for the longer horizon. Figure~\ref{fig:multiple_lt_cut} more concretely depicts how the $R^{2}$ score develops across the prediction horizon for the different teaching strategies and datasets.

\subsection{Additional Experiments}
Judging from the essential experiments, CL-ITF-x are our winning strategies on the datasets we tested. However, as mentioned in Section~\ref{sec:related_work}, there are many other approaches that predict chaotic systems with adapted \glspl*{rnn} architectures that take theoretical insights of dynamical systems into account. \gls*{stf} \citep{monfared2021train} does not require any architectural modifications but instead provides an adapted training strategy. It determines a time interval $\tau=\frac{ln 2}{LLE}$ that denotes how many \gls*{fr} steps are processed before the next \gls*{tf} value is used within one sequence. We find this strategy comparable to the \gls*{cl} approaches we study. Therefore, we execute another set of experiments where we use \gls*{stf} during the training of our encoder-decoder \gls*{gru} for all chaotic systems in Table~\ref{tab:systems_details}. Since our data is sampled with different $dt$, we have to redefine the time interval as $\tau=\frac{ln 2}{LLE\cdot dt}$.

\begin{table*}[htb]
	\caption{Results of \gls*{stf} with those of the baseline and the CL-ITF-x strategies}
	\label{tab:sparse_tf_results}
		\centering
		\begin{tabular}{l|l|rrr}
			\toprule
			\textbf{System} & \textbf{Strategy} &   \textbf{Epochs}    & \textbf{NRMSE} $\bm{\downarrow}$ & \textbf{Rel. impr.} $ \bm{\uparrow}$ \\
			\midrule
			\multirow{6}{*}{\rotatebox[origin=c]{0}{Mackey-Glass ($0.006$)}}
			& FR                             & $ 4\,713 $          &  $ \underline{0.00391} $   & --                               \\
			& TF                             & $ 44 $             &  $ 0.09535 $                       & --                                \\
			\cmidrule{2-5}
			& CL-ITF-P                 & $ 1\,566 $          &    $ 0.00104 $                    & $ \good{73.40\%} $    \\
			& CL-ITF-D                 & $ 1\,808 $          &    $ \bm{0.00211} $                    & $ \good{46.03\%} $    \\
			\cmidrule{2-5}
			& STF                          & $ 5\,517 $          &    $ 0.00254 $                    & $ \good{35.04\%} $    \\
			\cmidrule{1-5}
			\multirow{6}{*}{\rotatebox[origin=c]{0}{Thomas ($0.055$)}}
			& FR                             & $ 427 $           &  $ \underline{0.03416} $   & --                               \\
			& TF                             & $ 163 $            &  $ 0.34545 $                       & --                                \\
			\cmidrule{2-5}
			& CL-ITF-P                 & $ 677 $             &    $ \bm{0.00930} $                    & $ \good{72.78\%} $    \\
			& CL-ITF-D                 & $ 649 $            &    $ 0.01819 $                    & $ \good{46.75\%} $    \\
			\cmidrule{2-5}
			& STF                          & $ 432 $             &    $ 0.03655 $                    & $ \bad{-7.00\%} $    \\
			\cmidrule{1-5}
			\multirow{6}{*}{\rotatebox[origin=c]{0}{Rössler ($0.069$)}}
			& FR                             & $ 3\,863 $          &  $ \underline{0.00098} $   & --                               \\
			& TF                             & $ 500 $             &  $ 0.00743 $                       & --                                \\
			\cmidrule{2-5}
			& CL-ITF-P                 & $ 4\,523 $           &    $ \bm{0.00019} $                    & $\good{80.61\%}  $    \\
			& CL-ITF-D                 & $ 7\,267 $           &    $ 0.00027 $                    & $ \good{72.24\%} $    \\
			\cmidrule{2-5}
			& STF                          & $ 4\,796 $           &    $ 0.00065 $                    & $ \good{33.67\%} $    \\
			\cmidrule{1-5}
			\multirow{6}{*}{\rotatebox[origin=c]{0}{Hyper-Rössler ($0.14$)}}
			& FR                             & $ 6\,461 $           &  $ 0.00599 $                           & --                               \\
			& TF                             & $ 2\,788 $           &  $ \underline{0.00435} $                       & --                                \\
			\cmidrule{2-5}
			& CL-ITF-P                 & $ 2\,802 $            &    $ 0.00366 $                    & $  \good{15.86\%} $    \\
			& CL-ITF-D                 & $ 3\,317 $            &    $ 0.00326 $                    & $ \good{25.06\%} $    \\
			\cmidrule{2-5}
			& STF                          & $ 2\,645 $            &    $ \bm{0.00321} $                    & $ \good{26.21\%} $    \\
			\cmidrule{1-5}
			\multirow{6}{*}{\rotatebox[origin=c]{0}{Lorenz ($0.905$)}}
			& FR                             & $ 9\,18 $              &  $ 0.01209 $   & --                               \\
			& TF                             & $ 4\,67 $              &  $ \underline{0.00152} $                       & --                                \\
			\cmidrule{2-5}
			& CL-ITF-P                 & $ 1\,137 $              &    $ \bm{0.00060} $                    & $ \good{60.53\%} $    \\
			& CL-ITF-D                 & $ 5\,78 $               &    $ 0.00135 $                    & $  \good{11.18\%}  $    \\
			\cmidrule{2-5}
			& STF                          & $ 1\,853 $             &    $ 0.00511 $                    & $ \bad{-236.18\%} $    \\
			\cmidrule{1-5}
			\multirow{6}{*}{\rotatebox[origin=c]{0}{Lorenz'96 ($1.67$)}}
			& FR                             & $ 8\,125 $             & $ 0.07273 $  & --                                     \\
			& TF                             & $ 4\,175 $             & $ \underline{0.03805} $      & --                               \\
			\cmidrule{2-5}
			& CL-ITF-P                 & $ 3\,886 $             &    $ 0.01680 $                    & $ \good{55.85\%} $    \\
			& CL-ITF-D                 & $ 3\,379 $             &    $ \bm{0.01628} $            & $ \good{57.21\%} $    \\
			\cmidrule{2-5}
			& STF                          & $ 1\,478 $              & $ 0.09030 $                      & $ \bad{-137.32\%}  $          \\
			\bottomrule
		\end{tabular}
\end{table*}

The results (cp. Tab.~\ref{tab:sparse_tf_results}) show that \gls*{stf} provides improved performance compared to the best baseline for three of six datasets ranging from $26.21$ -- $46.75\%$ relative improvement. For this it requires no additional hyper-parameters if the system's LLE is known. It also beats the best-performing CL strategy on the Hyper-Rössler dataset by a margin of $1.15\%$. For the rest of the datasets, the results stay behind those of the CL-ITF-x strategies, showing a worse, i.e., increased, \gls*{nrmse} by $7.00$ -- $236.18\%$. We assume that where \gls*{stf} systematically induces \gls*{tf} to catch chaos-preventing exploding gradients before they appear, using knowledge about the processed data, \gls*{cl} helps the model to find more consistent minima while not taking the degree of chaos into account. Originally, \gls*{stf} was not proposed to work for the kind of sequence-2-sequence architecture we are using, which may limit its effectiveness here. We further hypothesize that the \gls*{gru} is able to keep the risk of exploding gradients low in many cases, due to its gating mechanism and thus prevents \gls*{stf} to really show its full strength here.

For further investigation on CL for non-chaotic systems and to enrich our experiments, we conduct additional experiments that include the application of the baseline strategies \gls*{tf} and \gls*{fr} together with CL-ITF-P and CL-ITF-D on a periodic system and a measured real-world dataset. We use CL-ITF-P and CL-ITF-D, since they provide the most consistent relative improvements in the essential experiments. As periodic system, we study the Thomas attractor \citep{thomas1999deterministic} with parameter $b=0.32899$, which ensures a periodic behavior. Extending our evaluation to empirical data, we selected a time series used in the Santa Fe Institute competition \citep{weigend1993results}\footnote{https://github.com/tailhq/DynaML/blob/master/data/santafelaser.csv}.

\begin{table*}[htb]
	\caption{Comparing baseline strategies and CL-ITF-P on periodic Thomas and measured Santa Fe laser dataset}
	\label{tab:extra_results_best}
	 \centering
		\begin{tabular}{ll|rrrr}
			\toprule
			& \textbf{Strategy} & \textbf{\L}  &  \textbf{Epochs}    & \textbf{NRMSE} $\bm{\downarrow}$ & \textbf{Rel. impr.} $ \bm{\uparrow}$ \\
			\midrule
			\multirow{3}{*}{\rotatebox[origin=c]{0}{Per. Thomas}}
			& FR                            & -- & $ 542 $            & $ \underline{0.00057} $   & --                                \\
			& TF                             & -- & $ 542 $            & $ 0.00107 $                       & --                                \\
			& CL-ITF-P                  &  $ 8\,000$  & $ 794 $     &    $ \bm{0.00033} $ & $\good{42.11\%}$    \\
			& CL-ITF-D                  & $ 125 $ & $ 326 $           &    $ 0.00045 $          & $\good{21.05\%}$    \\
			\cmidrule{1-6}
			\multirow{3}{*}{\rotatebox[origin=c]{0}{Santa Fe ($m=20$)}}
			& FR                             & -- & $ 500 $       & $ \underline{0.02170} $  & --                                     \\
			& TF                             & -- & $ 22 $           & $ 0.04793 $                       & --                                   \\
			& CL-ITF-P                 & $ 32\,000 $ & $ 469 $        & $ \bm{0.02042} $  & $\good{5.90\%}$          \\
			& CL-ITF-D                 & $ 4\,000 $ & $ 536 $        & $ 0.02232 $             & $\bad{-2.86\%}$          \\
			\cmidrule{1-6}
			\multirow{3}{*}{\rotatebox[origin=c]{0}{Santa Fe ($m=200$)}}
			& FR                             & -- & $ 208 $       & $ \underline{0.2786} $  & --                                     \\
			& TF                             & -- & $ 1 $           & $ 0.7104 $                       & --                                   \\
			& CL-ITF-P                 & $ 64\,000 $ & $ 1506 $        & $ \bm{0.2054} $  & $\good{26.27\%}$      \\
			& CL-ITF-D                 & $32\,000 $ & $ 201 $        & $ 0.2950 $               & $\bad{-5.89\%}$       \\
			\bottomrule
		\end{tabular}
\end{table*}

\begin{figure}[htb]
	\centering
	\includegraphics[width=0.7\linewidth]{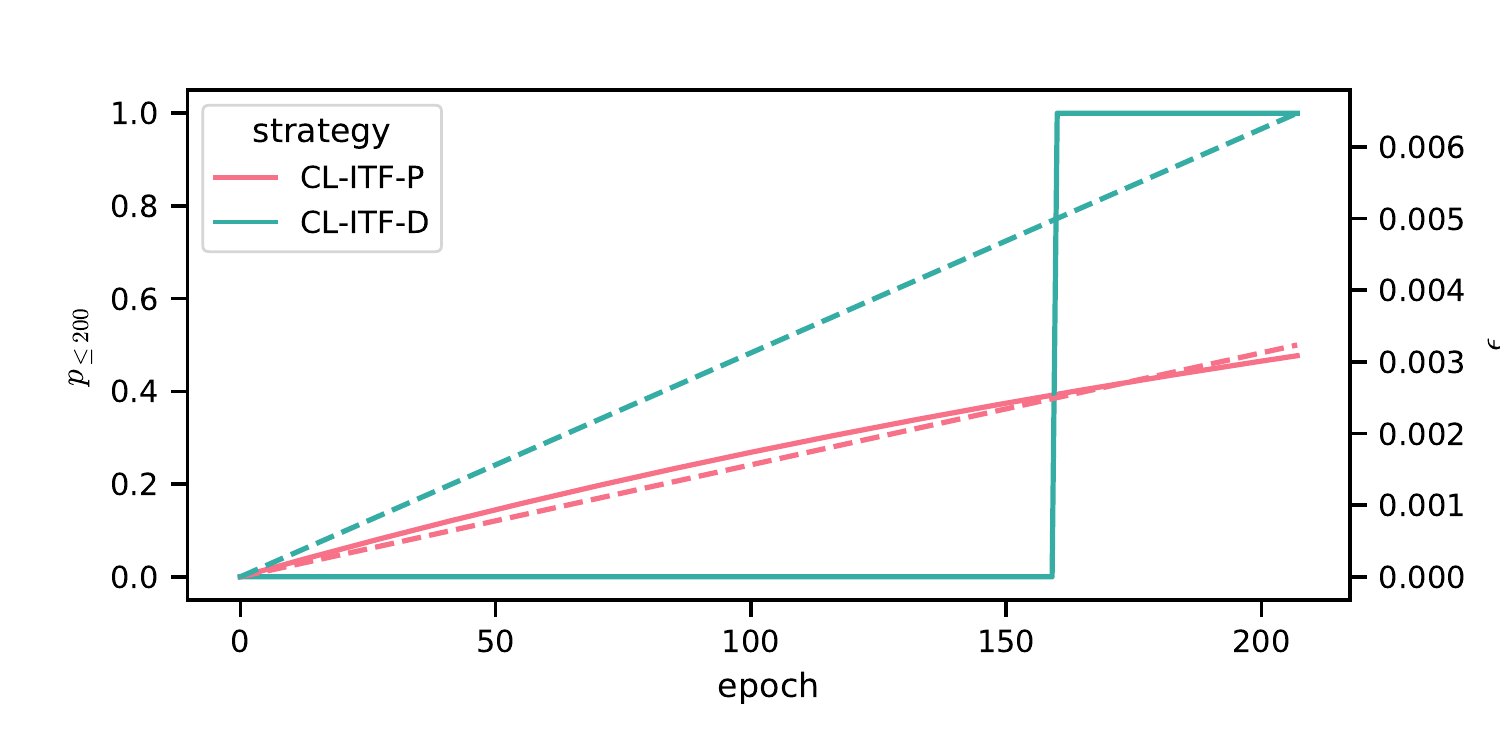}
	
	\caption{The probability course of a sample to include at least one TF step (solid) and the corresponding TF probability $\epsilon$ (dashed). Exemplary shown for the Santa Fe setup with $200$ output steps (cp.~Tab.~\ref{tab:extra_results_best}).}
	\label{fig:santafe_probabilities}
\end{figure}

The results in Table~\ref{tab:extra_results_best} support our assumption that CL-ITF-x strategies are also successfully applicable for time series data originating dynamical systems with periodic behavior, achieving relative improvements of $21.05$ -- $42.11\%$. Regarding the Santa Fe dataset we observe less impact by our strategies. Only having an improvement by $5.90\%$ for CL-ITF-P and a worsening by $2.86\%$ for CL-ITF-D on the empirical real-world data. When increasing the prediction horizon from $20$ to $200$ steps, we observe an increased relative improvement of $26.27\%$ for CL-ITF-P. At the same time, we observe a relative decrease of $-5.89\%$ for CL-ITF-D. To investigate the radical difference between these results, we study the $\epsilon$ in relation to the appearance of \gls*{tf} steps in the respective training curriculum. Figure~\ref{fig:santafe_probabilities} shows two different probabilities along the training epochs. There is the \gls*{tf} probability $\epsilon$ that is directly determined by the curriculum function and $p_{\leq200}$, which denotes the probability of a training sample to contain at least one \gls*{tf} step. It is computed as $p_{\leq200} = \epsilon \sum_{i=1}^{200} (1 - \epsilon)^{i - 1}$ for the probabilistic strategy. For CL-ITF-D, this probability is either $0$ or $1$, meaning that it only has an effect on training epochs $\geq160$. If we would use the same $\textrm{\L}$ as for CL-ITF-P here, the training would converge after $208$ epochs just as in the plain \gls*{fr} case. CL-ITF-P though, due to its probabilistic nature, can sporadically induce \gls*{tf} steps for some of the batches in the earlier phase of training and thus provides a much smoother transition towards \gls*{tf} phase. We argue that this smoothness is crucial for the given task, which appears to be very sensitive to \gls*{tf} training steps.

We also conducted experiments for other \gls*{rnn} architectures, i.e., we selected a vanilla \gls*{rnn}, an \gls*{lstm}, a \gls*{urnn} \citep{arjovsky2016unitary} and a \gls*{lrnn} \citep{erichson2020lipschitz}, in the same encoder-decoder setup as applied for the previous experiments with the \gls*{gru} architecture. In the Table~\ref{tab:other_rnn_results}, we compare the \gls*{nrmse} and relative improvement of these architectures on the four chaotic datasets from the exploratory experiments (cp. Tab.~\ref{tab:all_results_best}). We compare \gls*{tf}, \gls*{fr} and previously best-performing training strategies CL-ITF-P and CL-ITF-D (cp. Tab.~\ref{tab:all_results_best_curriculum_length}). Except for one case, we observe that both \gls*{cl} strategies outperform the respective best-performing baseline strategy on the vanilla \gls*{rnn} as well as the \gls*{lstm} architecture. The only exception is a vanilla \gls*{rnn} trained to forecast the Lorenz’96 system using CL-ITF-P. This setup suffers a performance decrease by $83.45\%$, while the \gls*{nrmse} in all other experiments improves by $37.83$ -- $75.28\%$ \gls*{rnn} and $26.04$ -- $69.75\%$ \gls*{lstm} respectively when using the \gls*{cl} strategies. The \gls*{urnn} and the \gls*{lrnn} do not adapt their weight matrices directly while training, but rather update several structured building block matrices to construct the actual weight matrices. We observe improvements from $0.4$ to $77.75\%$ and from $0.21$ to $83.74\%$ for the two architectures respectively. We provide more detailed information about these experiments in appendix \ref{apx:different_rnns}.

\begin{figure*}[htbp]
	\centering
	\begin{subfigure}{0.65\linewidth}
		\includegraphics[width=\linewidth]{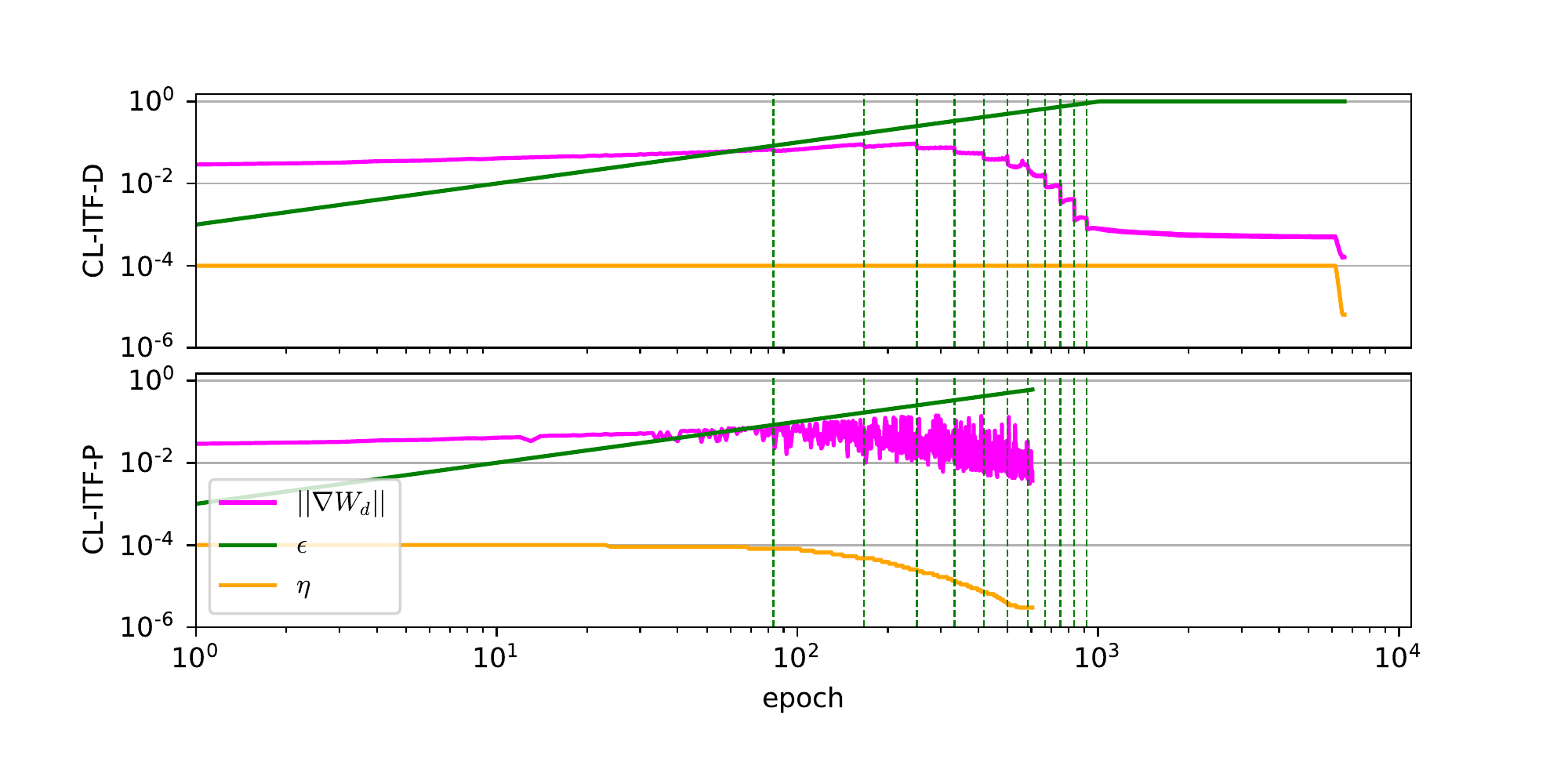}
		\caption{RNN}
		\label{fig:rnn_l96_gradnorm}
	\end{subfigure}

	\begin{subfigure}{0.49\linewidth}
		\includegraphics[width=\linewidth]{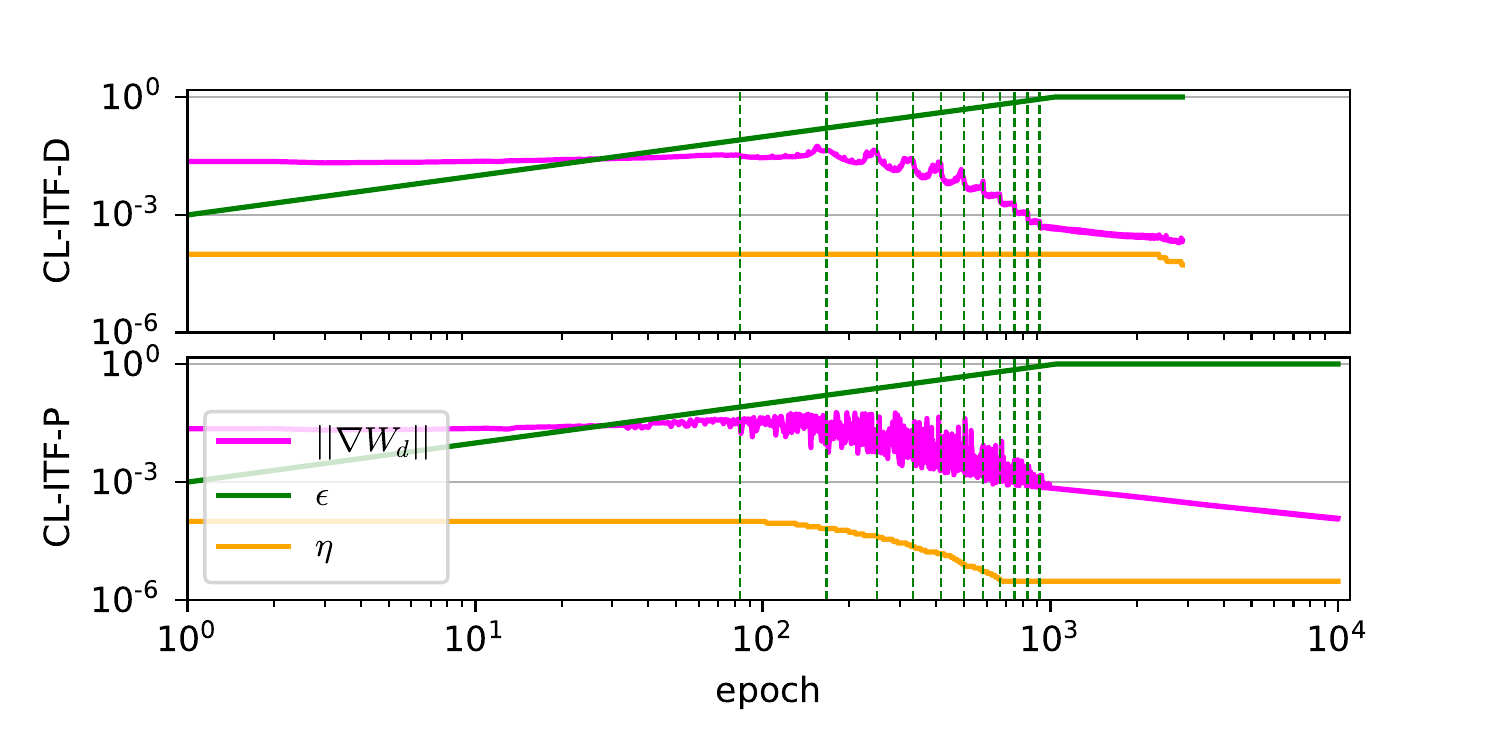}
		\caption{LSTM}
		\label{fig:lstm_l96_gradnorm}
	\end{subfigure}
	\begin{subfigure}{0.49\linewidth}
		\includegraphics[width=\linewidth]{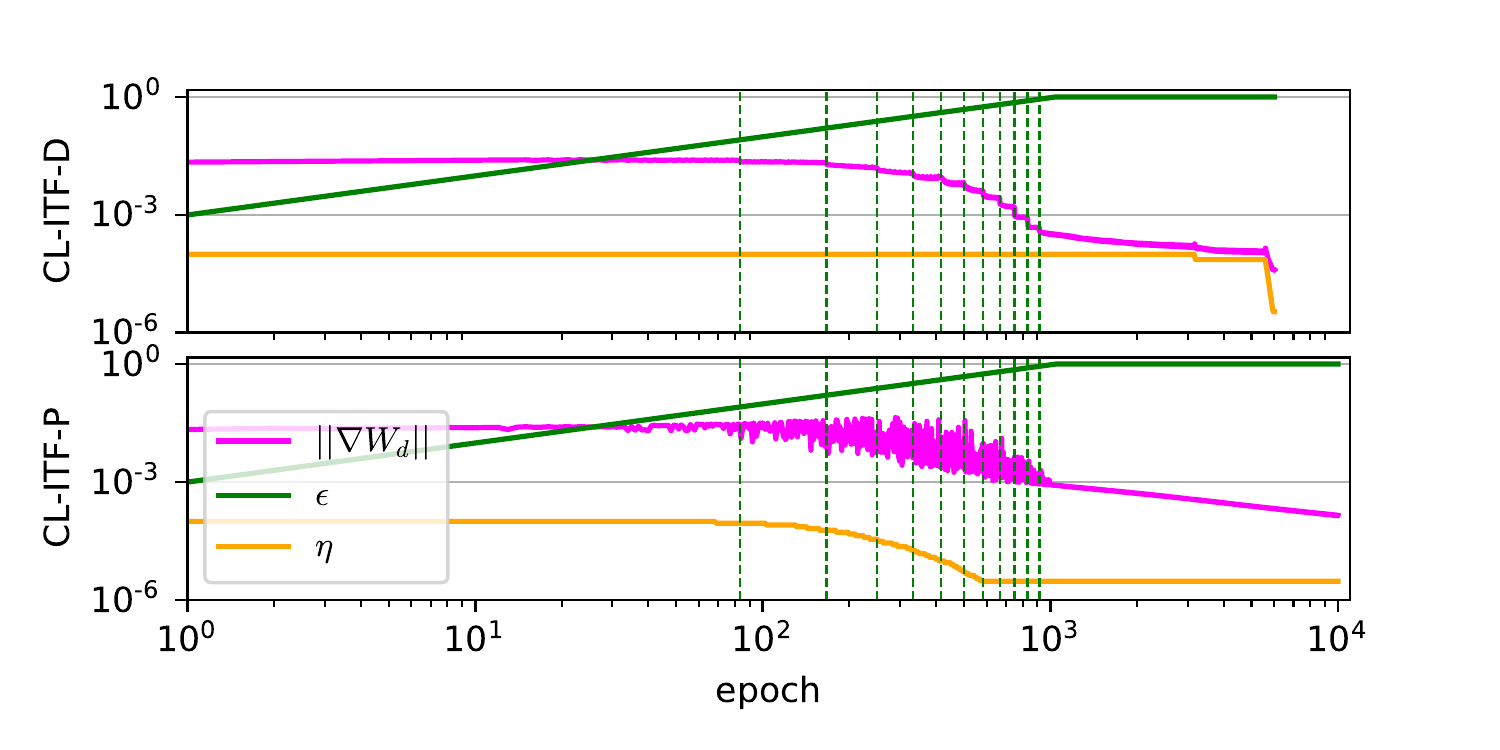}
		\caption{GRU}
		\label{fig:gru_l96_gradnorm}
	\end{subfigure}
	\caption{Course of the average decoder gradient norm $||\nabla W_d||$ during training on Lorenz'96 data together with TF probability $\epsilon$ and learning rate $\eta$.}
\end{figure*}

In our experiments on the Lorenz'96 data, the performance of the encoder-decoder \gls*{rnn} remains an outlier. Especially when we consider that for the deterministic variant CL-ITF-D there is a clear NRMSE improvement by $75.28\%$. We use this setup to take a closer look on the training gradients and process. In Figure~\ref{fig:rnn_l96_gradnorm}, we plot the mean of the decoder \gls*{rnn} gradient norm ($||\nabla W_d||$) over the training epochs for the CL-ITF-P and the CL-ITF-D case. We also add the \gls*{tf} probability $\epsilon$ and the learning rate $\eta$ to the plot. For comparison with the two successful setups, we also give an overview of the gradient norm of the \gls*{gru} and the \gls*{lstm}-based model under the same conditions (cp.~Figs~\ref{fig:gru_l96_gradnorm} and \ref{fig:lstm_l96_gradnorm}).

For all models, we observe an oscillation of the gradient norm during CL-ITF-P training. CL-ITF-D avoids this behavior with a constant iteration scale curriculum that keeps the TF-FR-pattern fix for several epochs between the changes. The crucial epochs are indicated by green dashed vertical lines. In contrast, CL-ITF-P provides a probabilistic iteration scale curriculum that leads to intra-epoch changes. The resulting variation in the gradient norm only stops when the training scale curriculum hits the final $\epsilon=1$. Preceding that, the oscillating gradient norm originates from an unsteady training loss. Most of the time, the unsteady loss can be compensated due to the increasing amount of \gls*{tf} over the epochs. For the Lorenz'96 data, however, it seems the unsteady loss together with the used scheduler lead to a rapid decrease of the learning rate. This effect is observable during the curriculum training of the basic \gls*{rnn}, the \gls*{lstm} and the \gls*{gru}. The latter two models seem to handle the resulting minimal learning rate of $3e^{-6}$ much better and manage to continue their training successfully.

\begin{figure*}[htbp]
	\centering
	\includegraphics[width=0.65\linewidth]{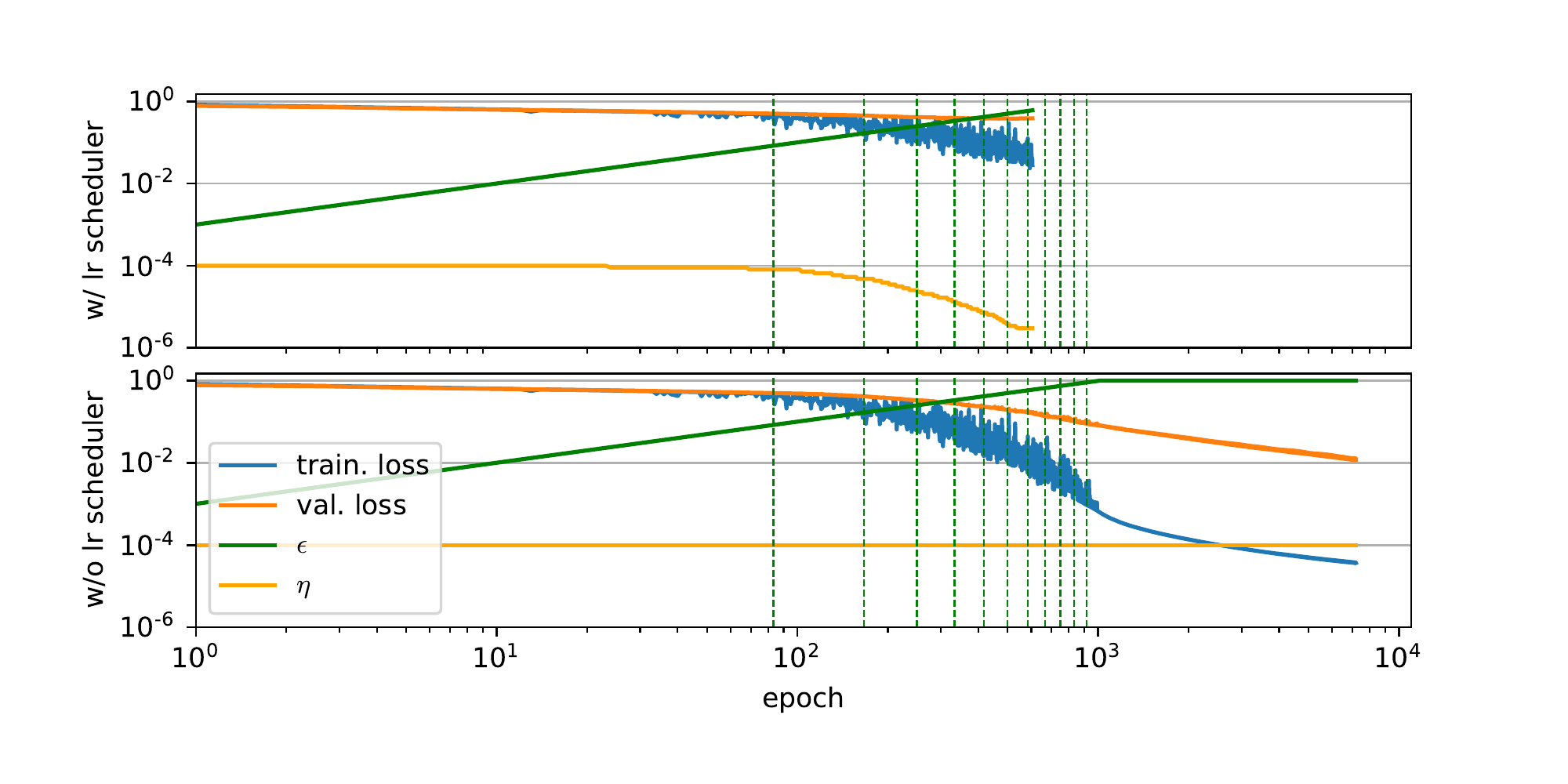}
	\caption{Course of the training and validation loss for the RNN on Lorenz'96 data using CL-ITF-P with and without learning rate scheduler.}
	\label{fig:rnn_l96_loss_val_loss}
\end{figure*}

To get an impression of the \gls*{rnn}'s behavior without the unwanted learning rate drop, we conducted the same CL-ITF-P experiments again without applying any learning rate scheduler. This time we plot the training loss, that exhibits the same oscillating behavior as the gradient norm, together with the validation loss, to see if the change leads to improved loss values. The plot in Figure~\ref{fig:rnn_l96_loss_val_loss} shows that without the unwanted learning rate drop, the basic \gls*{rnn} remains stable during the transient phase towards $\epsilon=1$ and ends up with a validation loss of $0.012$ instead of  $0.373$.
For the more sophisticated \glspl*{rnn}, we see clearly improved validation losses after much less epochs as well. The corresponding plots are shown in appendix \ref{apx:lr_scheduler}.

\begin{table*}[htb]
	\caption{Forecasting performance of additional RNNs (vanilla RNN, LSTM, unitary evolution RNN (uRNN) and Lipschitz RNN (LRNN)) on the different chaotic datasets. Given is the NRMSE with relative improvement in parenthesis for the CL strategies.}
	\label{tab:other_rnn_results}
		\centering
		\begin{tabular}{ll|cccc}
			\toprule
			\textbf{System} & \textbf{Strategy} & \textbf{RNN} &  \textbf{LSTM}    & \textbf{uRNN} & \textbf{LRNN} \\
			\midrule
			\multirow{6}{*}{\rotatebox[origin=c]{0}{Thomas}}
			& FR                                                                               & $ 0.48117 $                                        & $ 0.43698 $                    &  $ \underline{0.74991} $  &  $ 0.11697 $    \\
			& TF                                                                               & $ \underline{0.41274} $                      & $ \underline{0.05265} $  & $ 0.85679 $ &  $ \underline{0.05714} $     \\
			\cmidrule{2-6}
			& \multirow{2}{*}{CL-ITF-P}                                           & $ \bm{0.17955} $                                & $ 0.01892 $                     & $ \bm{0.61889} $  &  $ 0.05237 $  \\
			&                                                                                    & $ \good{(56.50\%)} $                          & $ \good{(64.06\%)} $       & $ \good{(17.47\%)} $   & $ \good{(8.35\%)} $  \\
			\cmidrule{3-6}
			& \multirow{2}{*}{CL-ITF-D}                                           & $ 0.25659 $                                        &  $ \bm{0.01181} $            & $ 0.73331 $   & $ \bm{0.03926} $   \\
			&                                                                                    & $ \good{(37.83\%)} $                          & $ \good{(77.57\%)} $       & $ \good{(2.21\%)} $  & $ \good{(31.29\%)} $   \\
			\cmidrule{1-6}
			\multirow{6}{*}{\rotatebox[origin=c]{0}{Rössler}}
			& FR                                                                              & $ \underline{0.00747} $                     & $ 0.00210 $                      & $ 0.03726 $  & $ \underline{0.27122} $   \\
			& TF                                                                              & $ 0.50174 $                                        & $ \underline{0.00139} $   & $ \underline{0.03317} $  & $ 0.90844 $   \\
			\cmidrule{2-6}
			& \multirow{2}{*}{CL-ITF-P}                                           & $ \bm{0.00283} $                               & $ \bm{0.00063}$              & $ 0.02473 $  & $ 0.05185 $  \\
			&                                                                                    & $ \good{(62.12\%)} $                         &  $ \good{(54.68\%)} $       & $ \good{(25.44\%)} $ & $ \good{(80.88\%)} $  \\
			\cmidrule{3-6}
			& \multirow{2}{*}{CL-ITF-D}                                          & $ 0.00319 $                                        & $ 0.00085 $                      & $ \bm{0.02420} $  & $ \bm{0.04409} $  \\
			&                                                                                   & $ \good{(57.30\%)} $                          & $ \good{(38.85\%)} $        & $ \good{(27.04\%)} $ & $ \good{(83.74\%)} $   \\
			\cmidrule{1-6}
			\multirow{6}{*}{\rotatebox[origin=c]{0}{Lorenz}}
			& FR                                                                              & $ 0.11389 $                                        & $ 0.06526  $                    & $ \underline{0.13021} $ & $ 0.06675 $  \\
			& TF                                                                              &  $\underline{0.00913} $                      &  $ \underline{0.00169} $  & $ 0.96357 $ & $ \underline{0.02830} $  \\
			\cmidrule{2-6}
			& \multirow{2}{*}{CL-ITF-P}                                           & $ 0.00603 $                                        & $ \bm{0.00075} $              & $ \bm{0.03098} $ & $ \bm{0.01732} $   \\
			&                                                                                    & $ \good{(33.95\%)} $                          & $ \good{(55.62\%)} $        & $ \good{(76.21\%)} $ & $ \good{(38.80\%)} $   \\
			\cmidrule{3-6}
			& \multirow{2}{*}{CL-ITF-D}                                           & $ \bm{0.00378} $                                & $ 0.00125 $                      & $ 0.02897 $ & $ 0.02417 $  \\
			&                                                                                    &  $ \good{(58.60\%)} $                         & $ \good{(26.04\%)} $        & $ \good{(77.75\%)} $ & $ \good{(14.59\%)} $  \\
			\cmidrule{1-6}
			\multirow{6}{*}{\rotatebox[origin=c]{0}{Lorenz'96}}
			& FR                                                                              & $ \underline{0.31002}  $                     &  $ \underline{0.13010} $    & $ \underline{0.26359} $  & $ \underline{0.62710} $ \\
			& TF                                                                              & $ 0.57870 $                                         &  $ 0.22757 $                      & $ 0.61159 $ &  $ 0.65122 $ \\
			\cmidrule{2-6}
			& \multirow{2}{*}{CL-ITF-P}                                           & $ 0.56872 $                                        & $ \bm{0.03935} $               & $ 0.26253 $  & $ \bm{0.62170} $  \\
			&                                                                                    & $ \bad{(-83.45\%)} $                           &  $ \good{(69.75\%)} $        & $ \good{(0.4\%)} $ & $ \good{(0.86\%)} $ \\
			\cmidrule{3-6}
			& \multirow{2}{*}{CL-ITF-D}                                           & $ \bm{0.07663} $                                & $ 0.06855 $                       & $ \bm{0.18483} $ & $ 0.62581 $ \\
			&                                                                                    & $ \good{(75.28\%)} $                          & $ \good{(47.31\%)} $         & $ \good{(29.88\%)} $ & $ \good{(0.21\%)} $ \\
			\bottomrule
		\end{tabular}
\end{table*}
\section{Discussion}\label{sec:discussion}
\textbf{Baseline teaching strategies (RQ1).}
Considering the baseline teaching strategies \gls*{fr} and \gls*{tf}, we observe that per dataset, one of the strategies performs substantially better than the other. We also observe that for the upper, based on their LLE, less chaotic datasets in Table~\ref{tab:all_results_best_curriculum_length} \gls*{fr} performs better, while for the lower, more chaotic datasets, \gls*{tf} yields the better-performing model. However, a larger study with more datasets would be required to justify this claim. Our takeaway is that none of the methods can be universally recommended again motivating curriculum learning strategies.

\textbf{Curriculum learning strategies (RQ2).}
Among the curriculum learning strategies, we observe that blending \gls*{fr} with a constant ratio of \gls*{tf}, i.e., CL-CTF-P, almost consistently yields worse results than the best-performing baseline strategy and we therefore consider the strategy not relevant. The decreasing curricula CL-DTF-x that start the training with a high degree of \gls*{tf} and then incrementally reduce it to pure \gls*{fr} training partly perform better than the CL-CTF-P strategy and for a few datasets even substantially better than the baseline. We also proposed and studied increasing curricula CL-ITF-x that start the training with no or a low degree of TF, which is then incrementally increased over the course of the training. We observe that these strategies consistently outperform not only the baseline strategies, but all other tested curriculum learning strategies as well. However, it was not foreseeable when this would be the case, making their application not suitable for new datasets without a lot of experimentation and tuning.

\textbf{Training length (RQ3).}
Choosing an improper teaching strategy can result in an early convergence on a high level of generalization error, e.g., \gls*{tf} strategy for Mackey-Glass, Thomas, and Rössler. Models that yield better performance typically train for more iterations (cp. Tab.~\ref{tab:all_results_best} and \ref{tab:all_results_best_curriculum_length}). However, a longer training may not necessarily yield a better performance, e.g., \gls*{fr} vs. \gls*{tf} for Lorenz. When considering the best-performing CL-ITF-x strategies compared to the best-performing baseline strategy, we observe moderately increased training iterations for some datasets, i.e., Thomas, Rössler, Hyper Rössler, Lorenz, but also decreased training iterations for other datasets, i.e., Mackey Glass and Lorenz'96. To better understand whether the longer training is the true reason for the better-performing CL-ITF-x models, we compared their performance when only trained for as many iterations as the baseline model and still observe superior performance over it. In conclusion, we observe that the CL-ITF-x strategies facilitate a robust training, reaching a better generalizing model in a comparable training time on all six tested datasets.

\textbf{Prediction stability (RQ4).}
We evaluated the generalization as NRMSE for all models trained with the different training strategies per dataset, while forecasting a dataset-specific horizon of 1\,LT. However, this metric reflects only an average. When we strive for higher model performance on a multi-step prediction task, we often aim for a longer prediction horizon at an acceptable error. To compare prediction stability, we report the NRMSE metric separately, solely computed on the last $10\%$ of the $1$ LT horizon. Additionally, we computed how many LTs per datasets can be predicted before falling below an $R^{2}$ of $0.9$. We found that the CL-ITF-x strategies yielded the lowest NRMSE of the last $10\%$ predicted values across all datasets and even more promising that these strategies facilitated the longest prediction horizon without falling below $R^{2}=0.9$. We conclude that the CL-ITF-x strategies may help models find more stable minima in training, substantially improving their long-term forecasting performance.

\textbf{Curriculum parametrization (RQ5).}
Initially, we evaluated curriculum learning strategies with a variety of different transition functions and individual start and end \gls*{tf} rate (cp. \textit{exploratory} experiments). In these experiments, we observed high prediction performance of the CL-ITF-x strategies, but with a diverse dataset-specific best-performing curriculum, meaning that the application of these strategies for new datasets would have necessitated an extensive hyper-parameter search. In a second set of \textit{essential} experiments, we therefore explored whether we could identify curricula with less parametrization and a similar performance. We found those by using a linear transition that is solely configured by a single parameter $\textrm{\L}$ that determines the pace with which the \gls*{tf} increases or decreases over the course of the training. We found that these curricula were not only comparable to the previous transition functions and their parametrization, but performed better for all four datasets that we evaluated in both experimental sets and yielded the best-performing model across all six datasets in the second experiment.	

\textbf{Iteration scale curriculum (RQ6).}
Having the CL-ITF-x strategies outperforming the CL-DTF-x strategies leads to rethinking the hitherto common intuition of supporting the early phases of training by \gls*{tf} and moving towards \gls*{fr} in the later stages of training. Rather, we hypothesize that this lures the model into regions of only seemingly valid and stable minima in which they stay even during the more \gls*{fr}-heavy epochs, resulting in a premature termination of the training. This hypothesis is supported by the resulting loss curves for the respective experiments. For example, for the increasing curriculum case, the training duration and the course of the $\epsilon$ (c.p., Fig.~\ref{fig:roessler_loss}) show that a training that starts off with \gls*{fr} can enable the model to train much longer in the pure \gls*{tf} zone later. This is compared to the case where solely \gls*{tf} is used. The difference between the two CL-ITF-x strategies is how the prescribed amount of \gls*{tf} is distributed across the prediction steps of one training iteration (aka epoch). While the CL-ITF-D strategy distributes them as one cohesive sequence, the CL-ITF-P strategy distributes them randomly across the training sequence. We found that in the \textit{essential} experiments with the linear transition, the CL-ITF-P strategy performed overall best for four of the six datasets and would have also been a good choice with a substantial gain over the best-performing baseline training strategy for the other two datasets. In conclusion, we observe that the CL-ITF-P strategy trains models that yield 16--81\% higher performance than a conventional training with \gls*{fr} or \gls*{tf}. Apart from that, the \textit{essential} results do not lead to a clear conclusion whether to use CL-ITF-P or CL-ITF-D in a given case. The above-mentioned most obvious difference in the distribution of \gls*{tf} steps firstly may lead to a more coherent backpropagation in the deterministic variant, but it also results in a different behavior regarding maximum number of consecutive \gls*{fr} steps (TF-gap) for a given $\epsilon$. Having the same curriculum function applied for CL-ITF-P and CL-ITF-D makes the TF-gap decrease much faster in the early training stage for the probabilistic variant. Further, it changes the TF-gap in a logarithmic, rather than a linear fashion as for CL-ITF-D. This difference cannot be compensated by parametrizing the curriculum length, demonstrating the need for both strategies. Plus, this only affects the mean TF-gap produced by CL-ITF-P, which has a variance of $\frac{1-\epsilon}{\epsilon^2}$ due to its geometric distribution. Therefore, the TF-gap also varies a lot in the early training stage. In the context of the result Tables~\ref{tab:extra_results_best} and \ref{tab:other_rnn_results}, we see the probabilistic and the deterministic type of CL-ITF-x can exhibit disfunctions in some specific cases. More thorough investigation of such cases is required to further improve \gls*{cl} strategies, making them more generally applicable and fully identifying their limitations.

\textbf{Limitations of this work.} 
Our observations allow us to draw conclusions regarding appropriate curricula for the training of sequence-to-sequence \gls*{rnn} on continuous time series data. More precisely, the data in our study originates from dynamical systems that predominantly impose chaotic behavior. Consequently, all observations made here are currently limited to these kind of models and data. We acknowledge that more research is necessary to clarify several uncertain points. First, regarding the question why an increasing curriculum tends to find more stable minima throughout all studied datasets. Leading to the question, what determines a proper curriculum and parametrization. A first step could be to shift the focus on the baseline strategies again to extract primary conditions under which trainings get unstable. Therefore, a closer look at the weights and behavior of the model gradients during training, the statistics of the gradient of the processed time series and the used sampling rate is required at least. Finally, a more thorough investigation on empirical real-world data improving on the early and inconclusive results on the Santa Fe dataset (cp. Tab.~\ref{tab:extra_results_best}) is needed to reveal the capabilities and limitations of different CL strategies in this regard.

\textbf{Application in other domains.}
In this paper we focus exclusively on the field of time series forecasting. However, the proposed strategies could easily be applied to \gls*{ml} models exhibiting autoregressive predictions trained for other domains. For example, sequence-to-sequence \glspl*{rnn} and transformers for \gls*{amt}. As mentioned before, scheduled sampling, that is included in the CL-DTF-x strategies, was originally proposed, and successfully used to improve the training of \glspl*{rnn} for \gls*{nlp}, but, on the other hand, performed rather poorly for time series forecasting in our experiments. Thus, we cannot draw any conclusions about the usefulness of CL-ITF-x strategies in other domains based on our results either. Therefore, evaluating the effectiveness of increasing curricula in other domains remains a task for future work and, at this point, would go beyond the focus of our study.

\section{Conclusions}\label{sec:conclusion}
While training encoder-decoder \glspl*{rnn} to forecast time series data, strategies like \gls*{tf} produce a discrepancy between training and inference mode, i.e., the exposure bias. Curriculum learning strategies like scheduled sampling or training directly in \gls*{fr} mode are used to reduce this effect. We run an extensive series of experiments using six chaotic dynamical systems as benchmark and observed that not for all of them the exposure bias yields a crucial problem. We see that even an \gls*{fr} training that is per definition free of exposure bias may not be optimal. In fact, we observed that for half of our datasets, \gls*{tf} performs better, while for the other half, \gls*{fr} yields better results. With the focus on curriculum learning strategies like scheduled sampling, we proposed two novel strategies and found that those yield models that consistently outperform the best-performing baseline model by a wide margin of 15--80\% NRMSE in a multi-step prediction scenario over one Lyapunov time. We found that these models are robust in their prediction, allowing to forecast longer horizons with a higher performance. We found it sufficient to parametrize the strategy with a single additional parameter adopting the pace of the curriculum.

\subsubsection*{Acknowledgments}
\subsubsubsection{This research was funded by the project P2018-02-001 ``DeepTurb – Deep Learning in and of Turbulence'' of the Carl Zeiss Foundation; the German Federal Ministry for the Environment, Nature Conservation, Nuclear Safety and Consumer Protection (BMUV) grant: 67KI2086A; and the German Ministry of Education and Research (BMBF) grant: 01IS20062B.}

\bibliography{tmlr}
\bibliographystyle{tmlr}

\newpage
\appendix
\section{Training and Validation Loss Curves}
\label{apx:loss_curves}

\begin{figure*}[ph]
	\centering
	\begin{subfigure}{0.40\linewidth}
		\includegraphics[width=\linewidth]{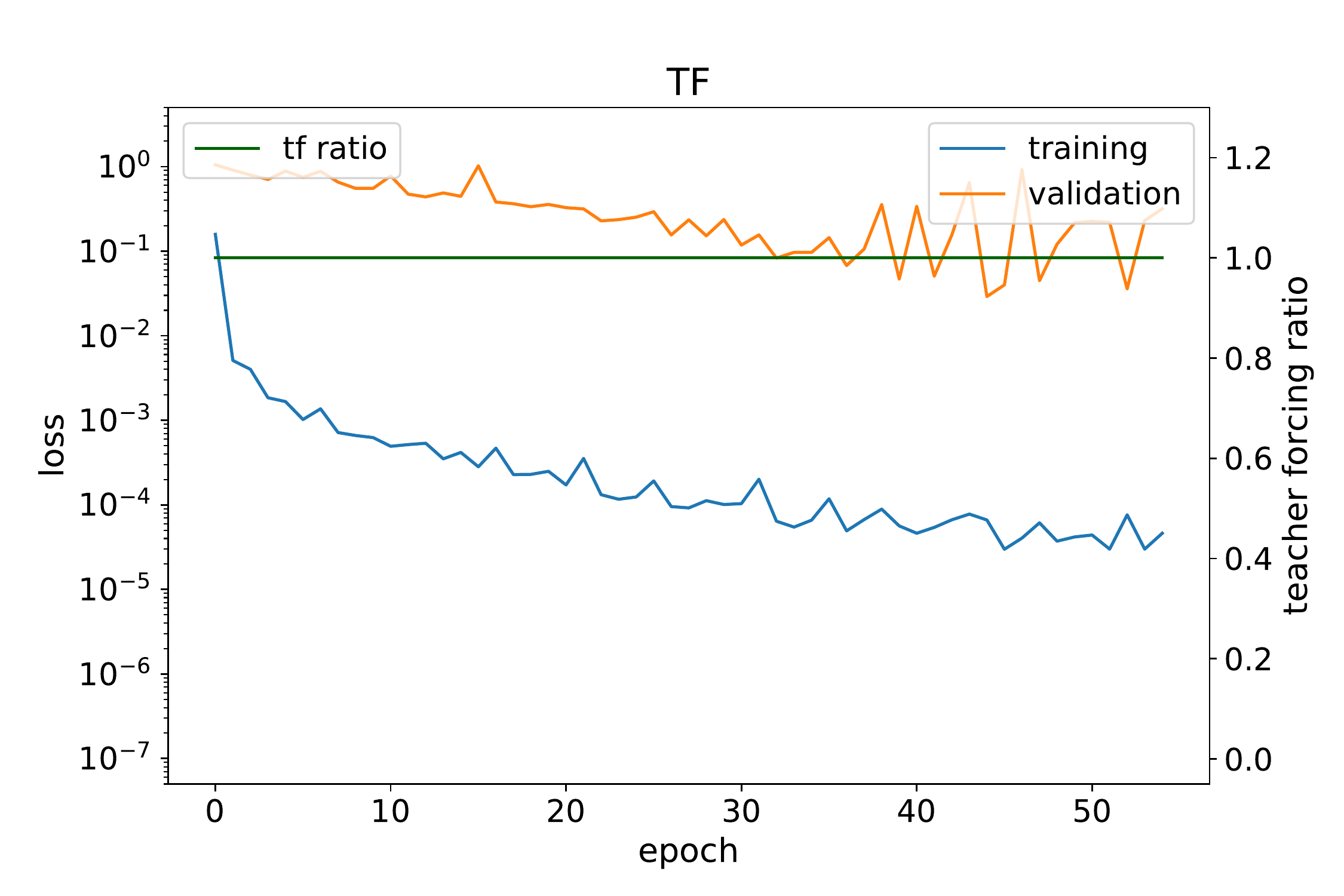}
		\caption{}
	\end{subfigure}
	\begin{subfigure}{0.40\linewidth}
			\includegraphics[width=\linewidth]{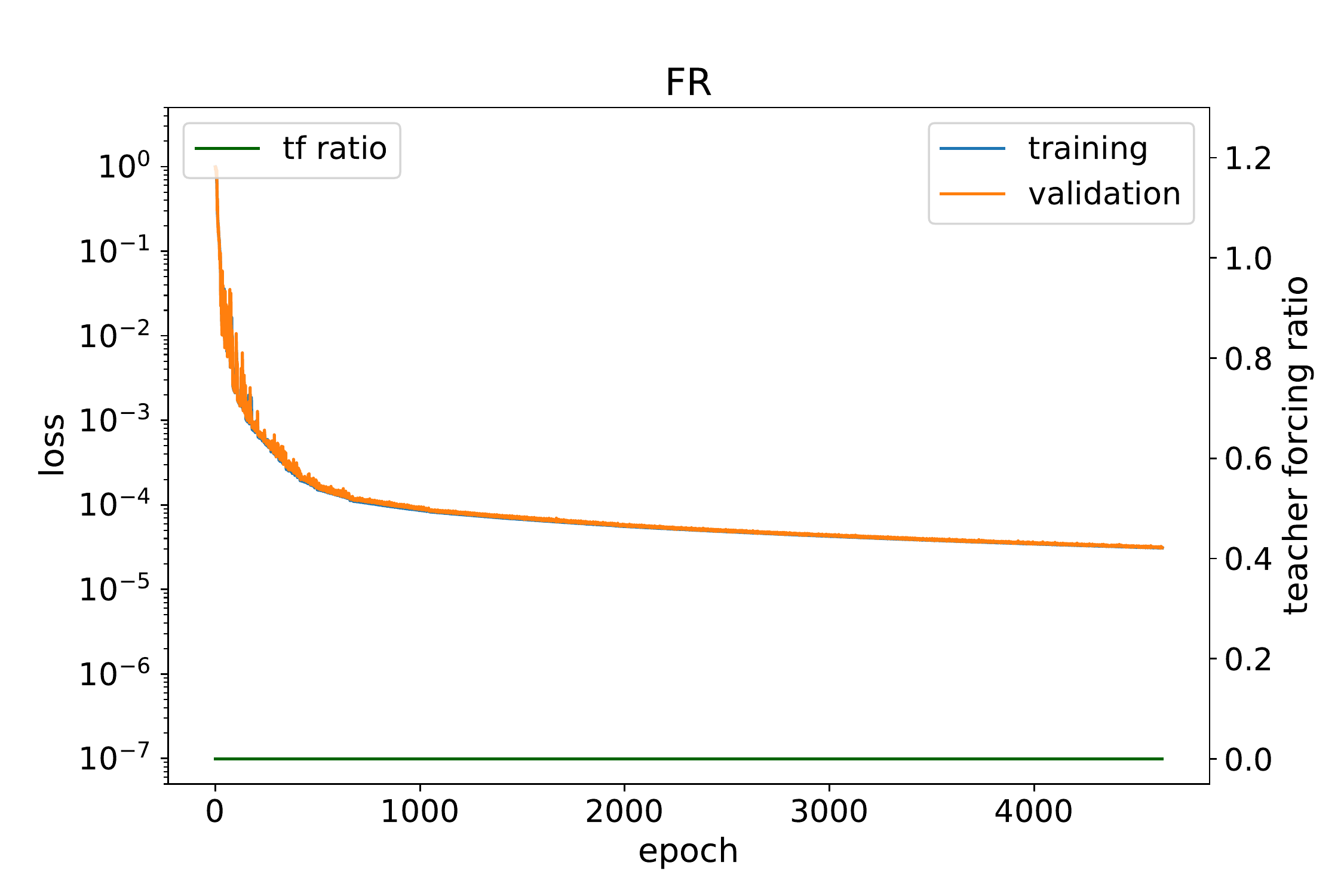}
		\caption{}
	\end{subfigure}
	
	\begin{subfigure}{0.40\linewidth}
		\includegraphics[width=\linewidth]{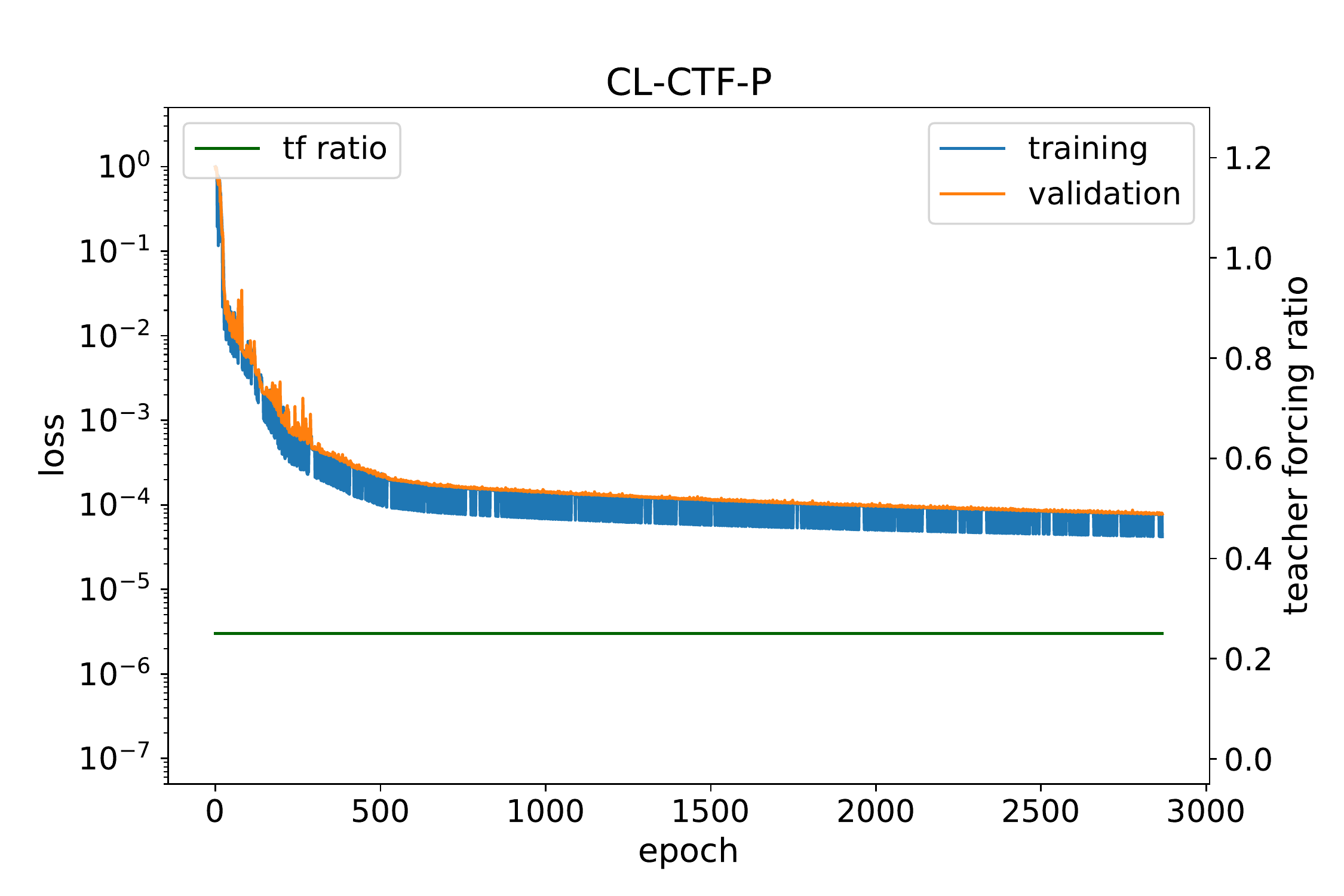}
		\caption{}
	\end{subfigure}

	\begin{subfigure}{0.40\linewidth}
		\includegraphics[width=\linewidth]{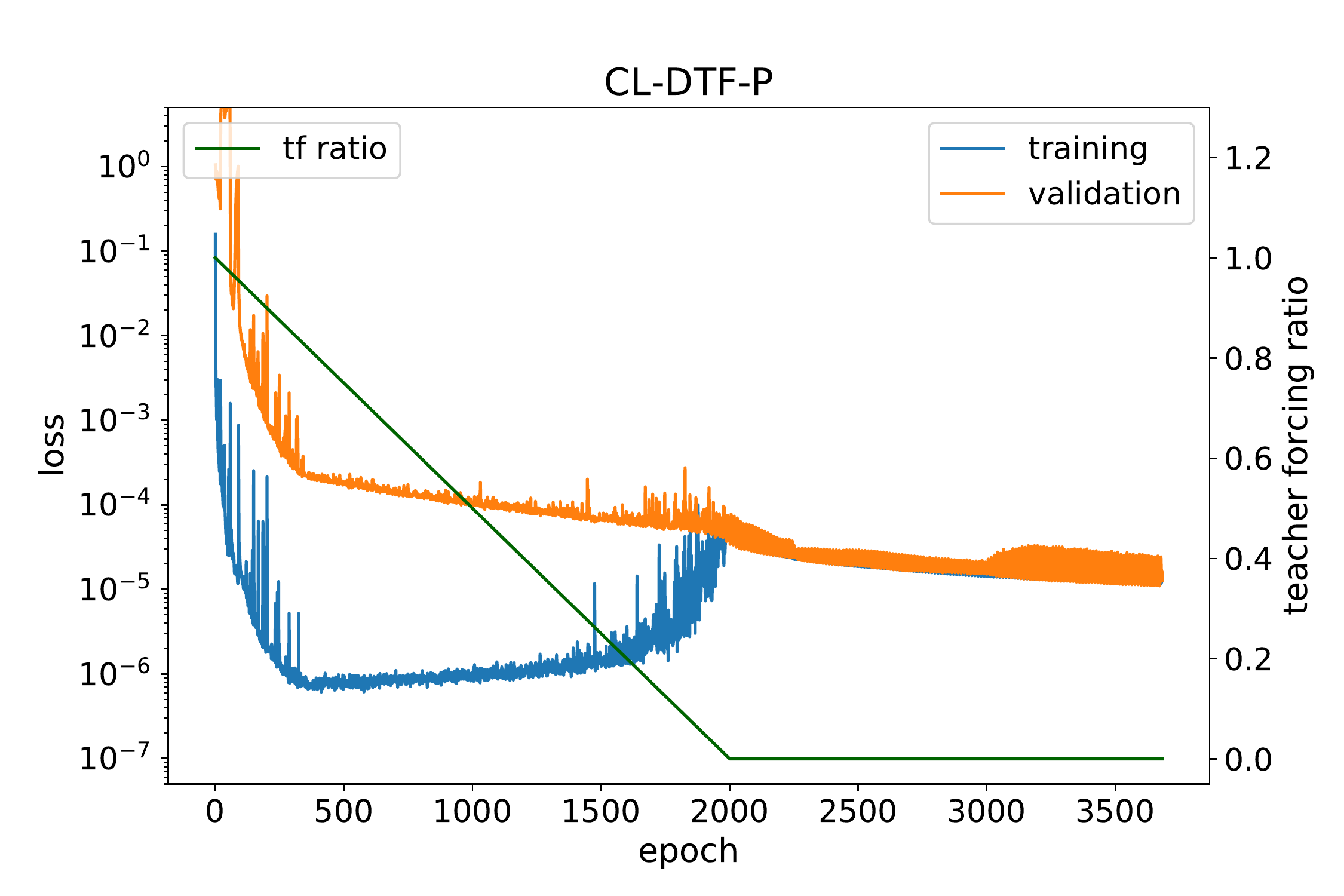}
		\caption{}
	\end{subfigure}
	\begin{subfigure}{0.40\linewidth}
	\includegraphics[width=\linewidth]{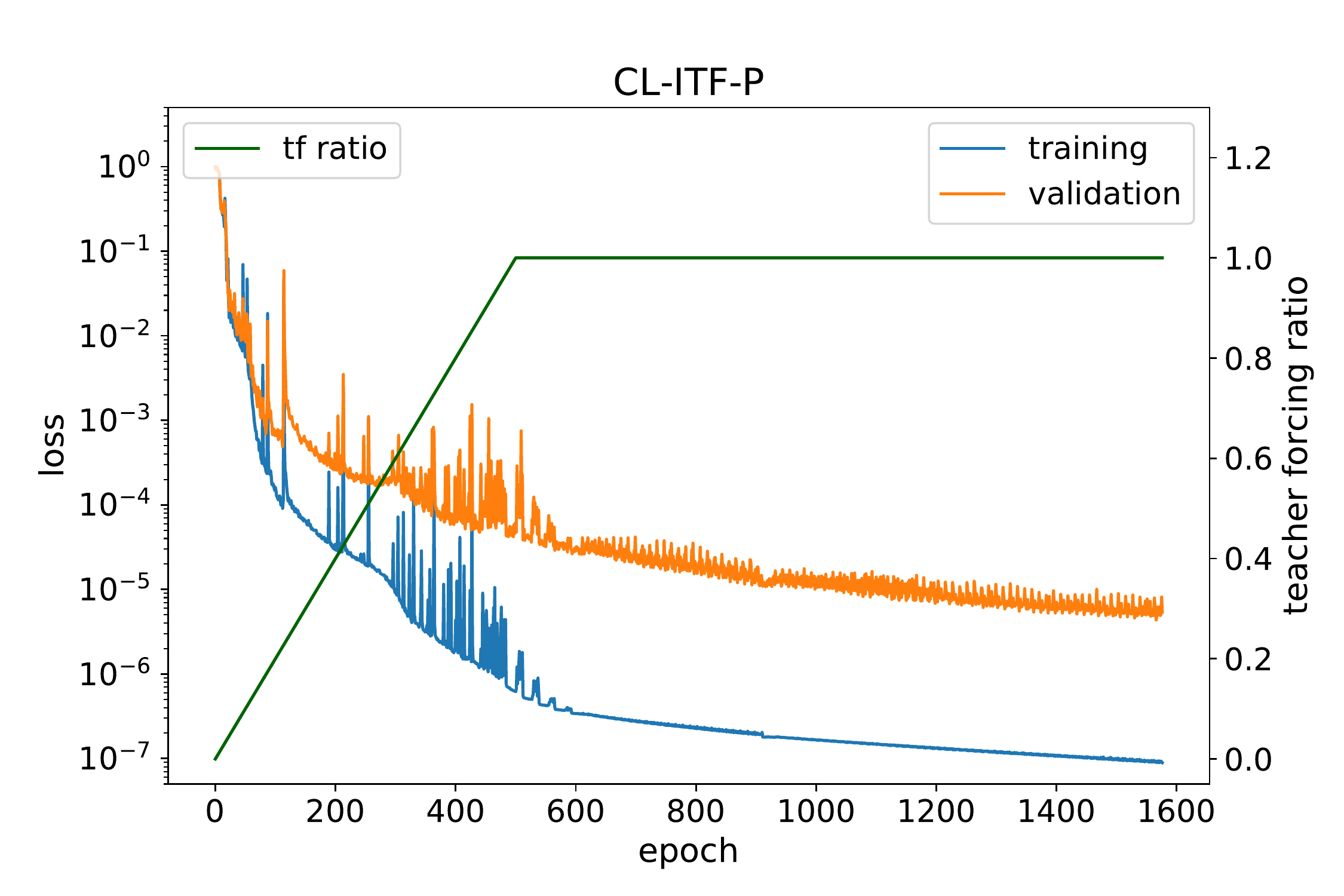}
	\caption{}
	\end{subfigure}

	\begin{subfigure}{0.40\linewidth}
		\includegraphics[width=\linewidth]{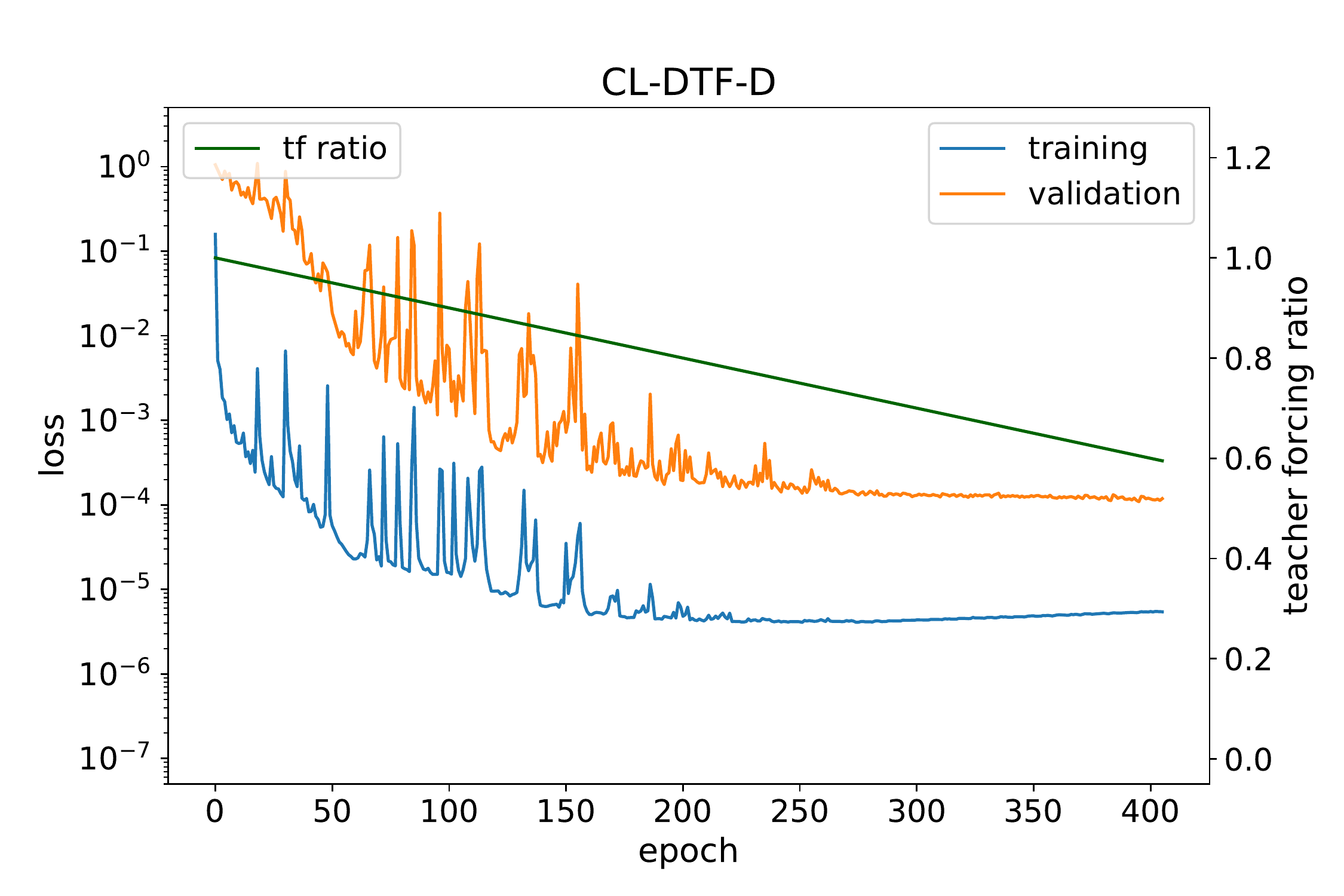}
		\caption{}
	\end{subfigure}
	\begin{subfigure}{0.40\linewidth}
		\includegraphics[width=\linewidth]{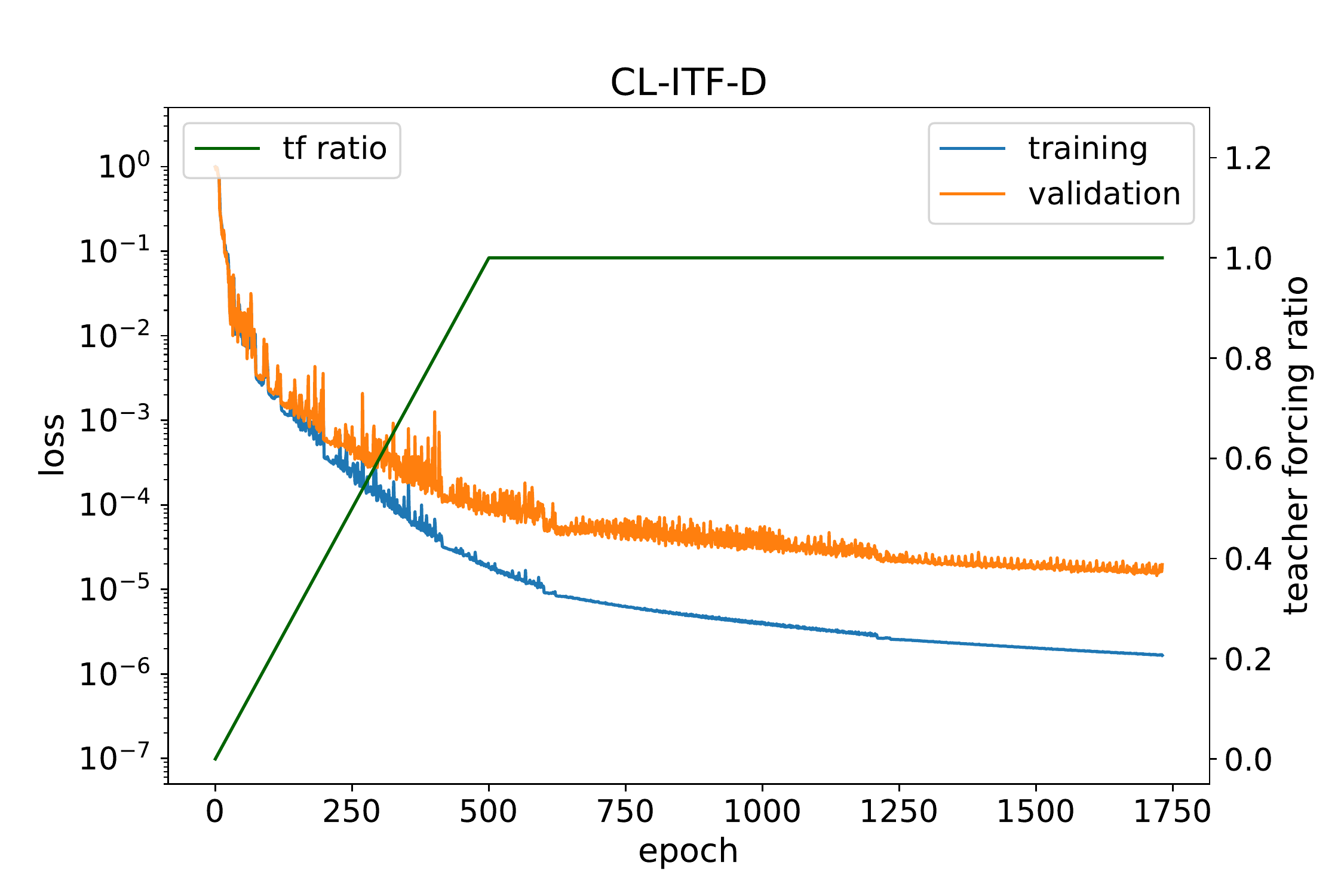}
		\caption{}
	\end{subfigure}
	
	\caption{Training and validation loss for Mackey-Glass}
	\label{fig:apx_mackey_loss_curves}
\end{figure*}

\begin{figure*}[ph]
	\centering
	\begin{subfigure}{0.40\linewidth}
		\includegraphics[width=\linewidth]{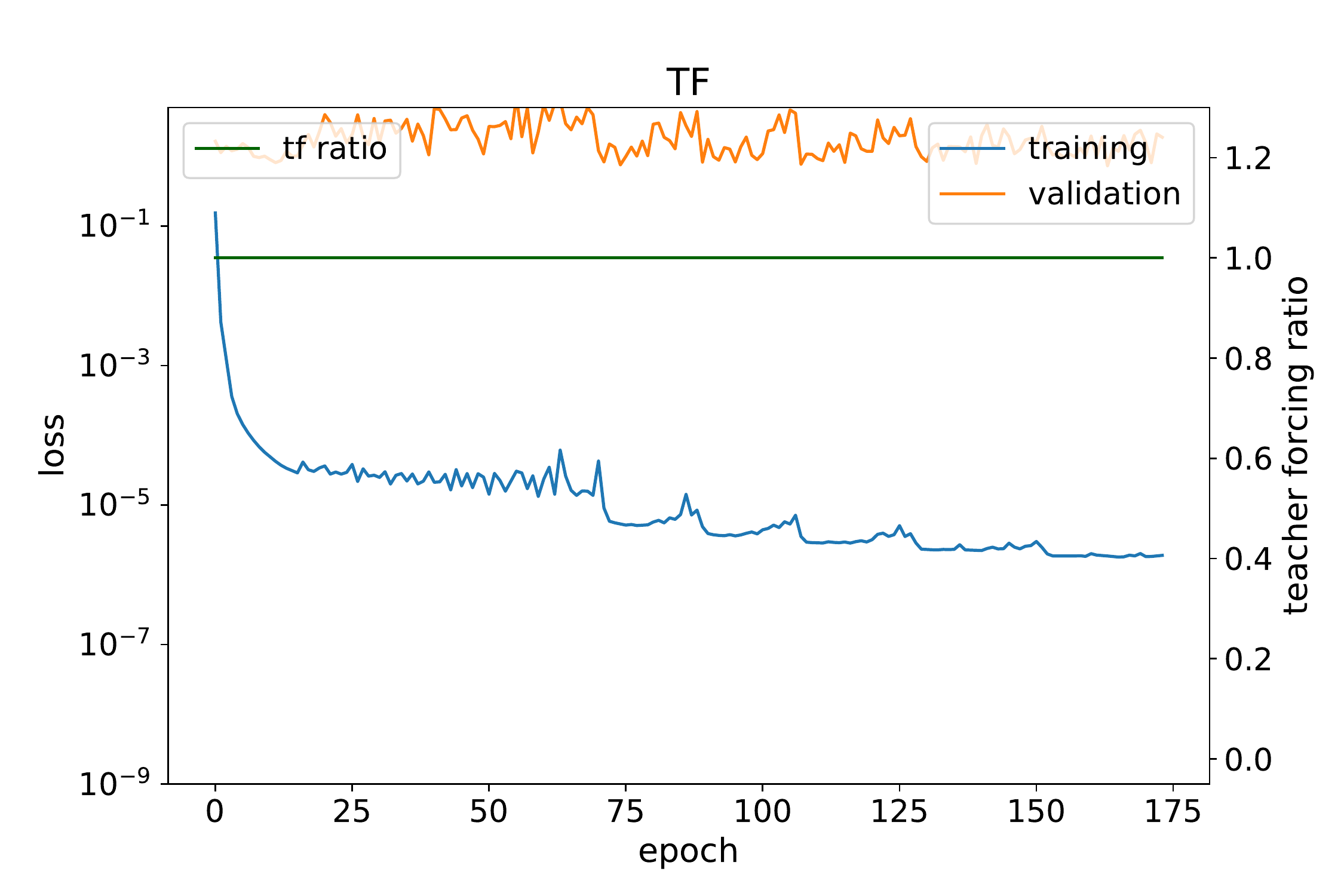}
		\caption{}
	\end{subfigure}
	\begin{subfigure}{0.40\linewidth}
			\includegraphics[width=\linewidth]{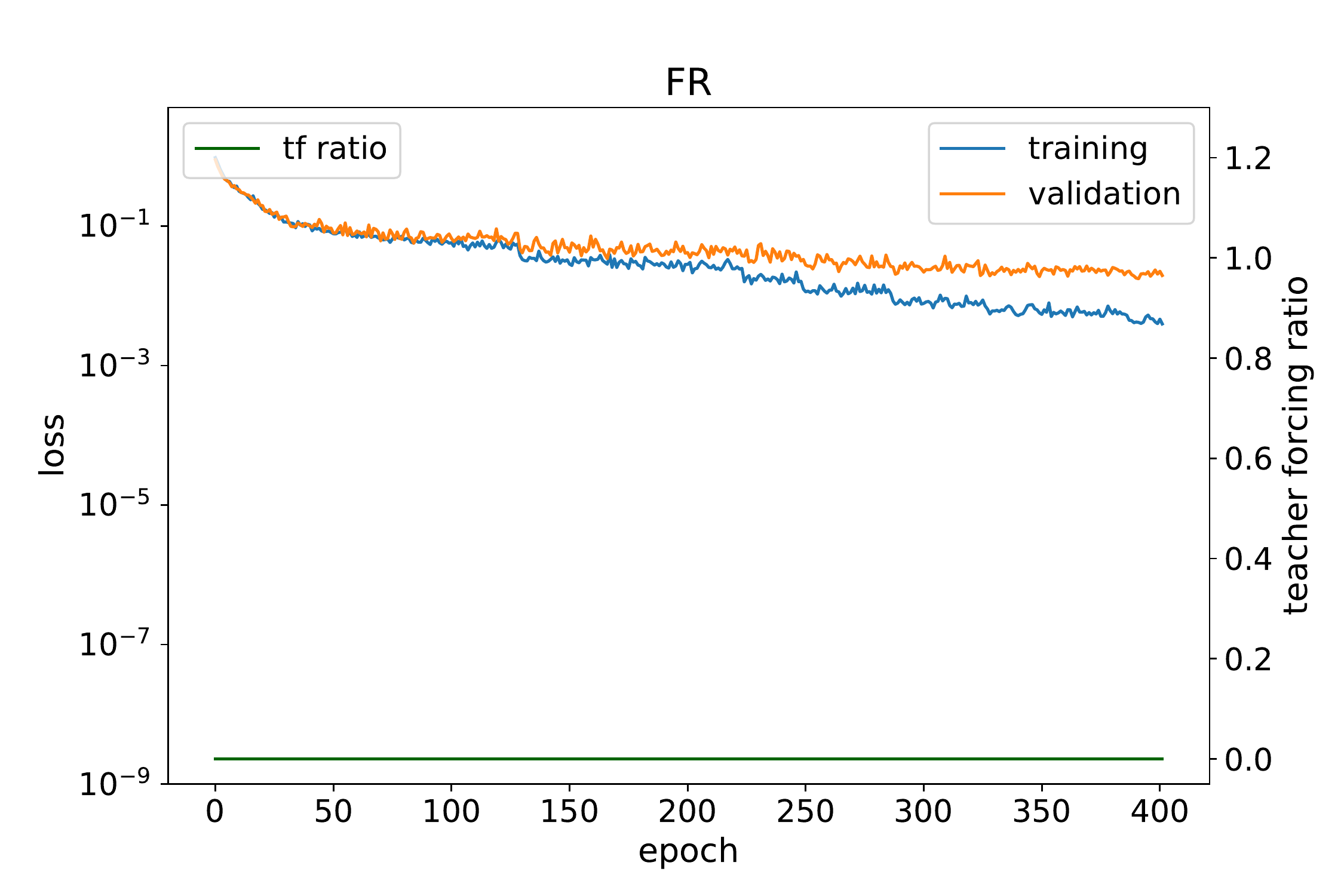}
		\caption{}
	\end{subfigure}
	
	\begin{subfigure}{0.40\linewidth}
		\includegraphics[width=\linewidth]{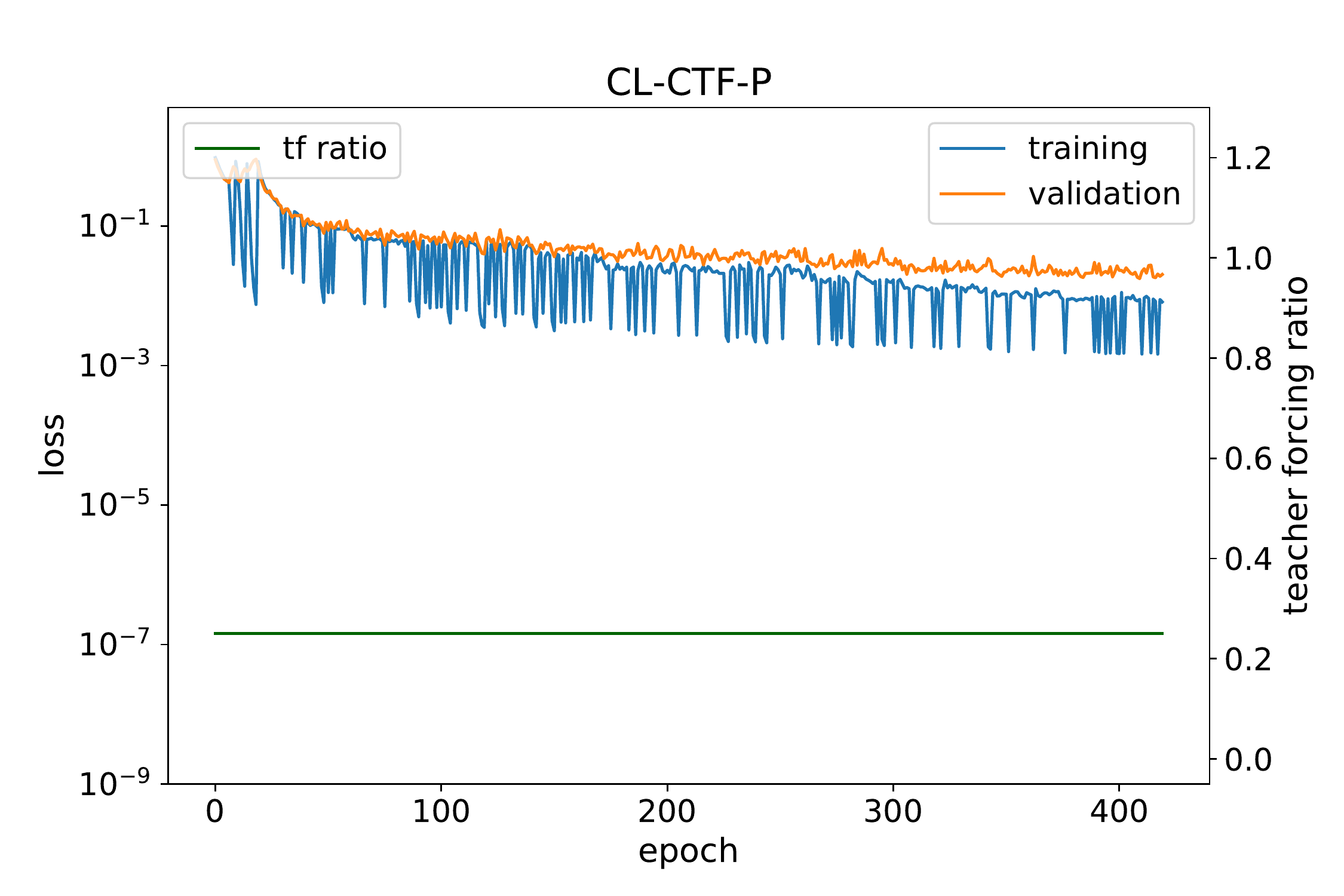}
		\caption{}
	\end{subfigure}

	\begin{subfigure}{0.40\linewidth}
		\includegraphics[width=\linewidth]{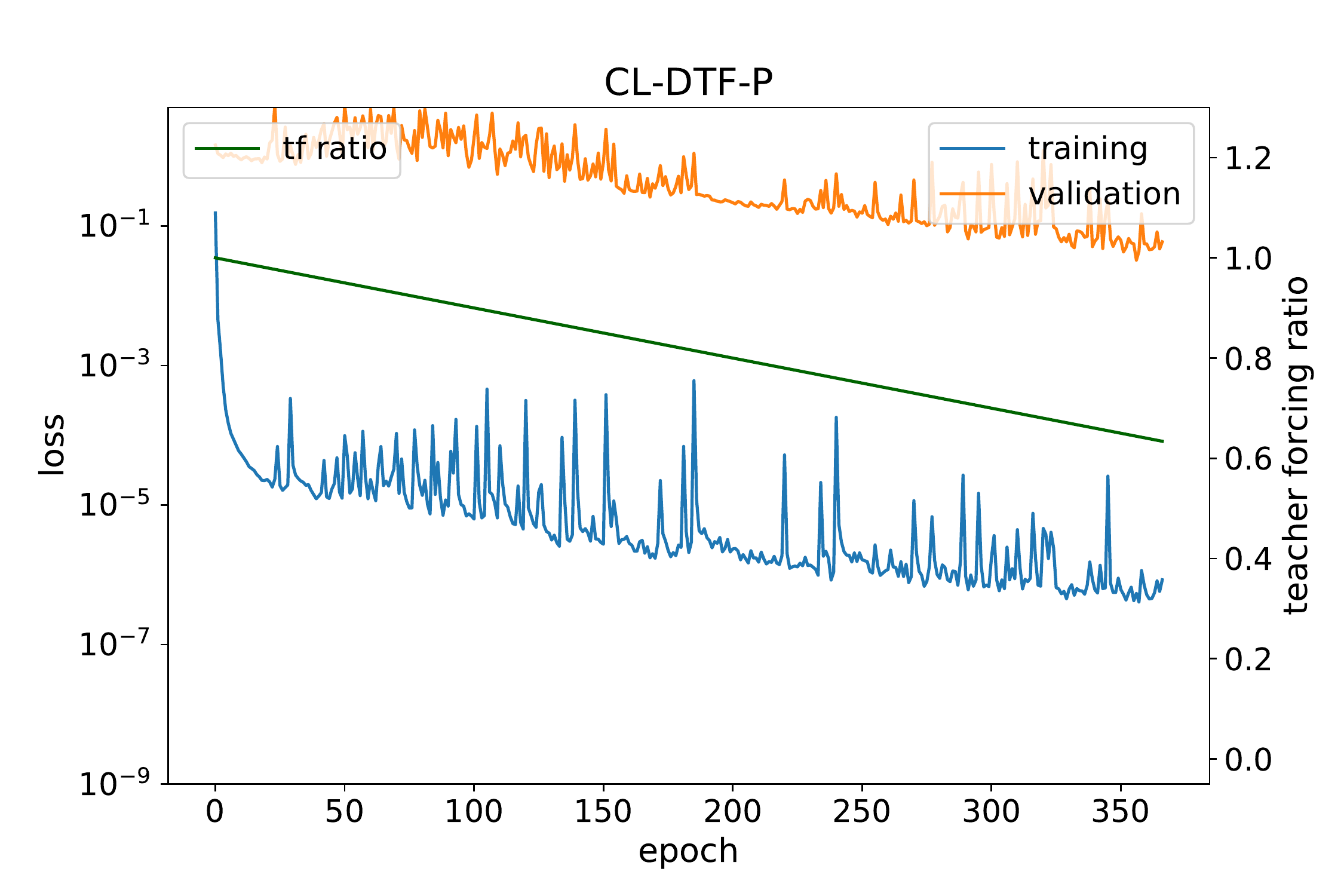}
		\caption{}
	\end{subfigure}
	\begin{subfigure}{0.40\linewidth}
		\includegraphics[width=\linewidth]{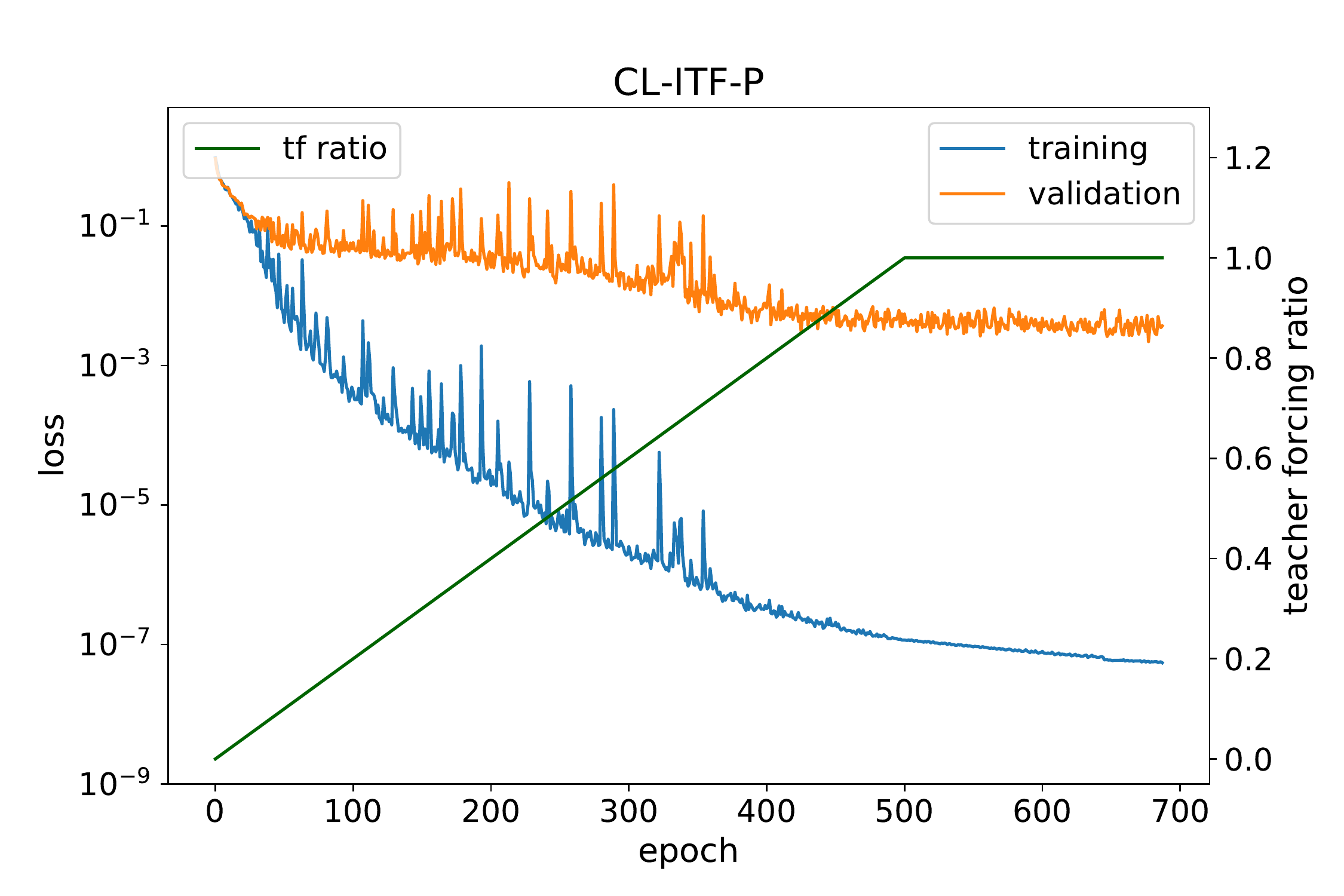}
		\caption{}
	\end{subfigure}

	\begin{subfigure}{0.40\linewidth}
	\includegraphics[width=\linewidth]{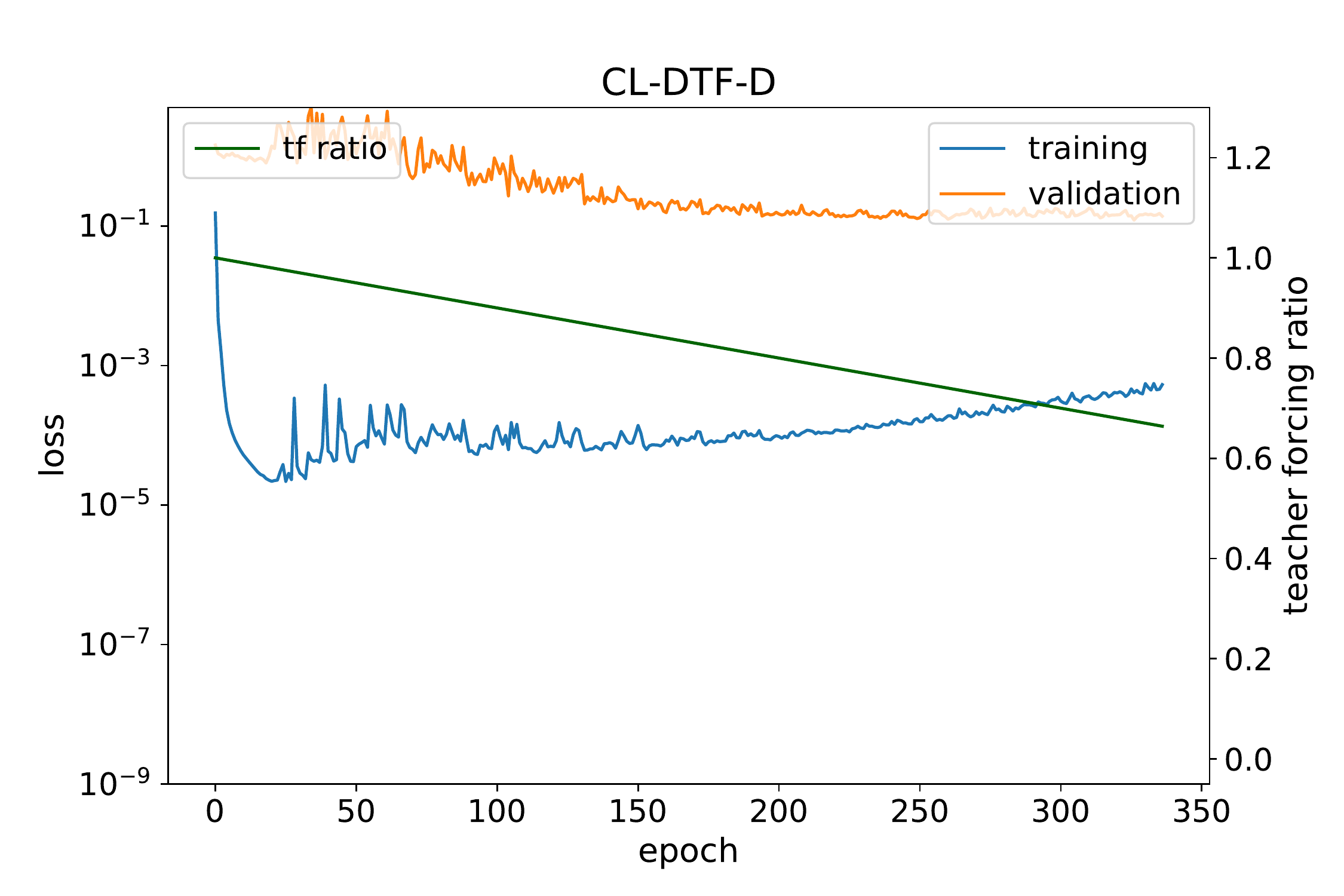}
	\caption{}
	\end{subfigure}
	\begin{subfigure}{0.40\linewidth}
		\includegraphics[width=\linewidth]{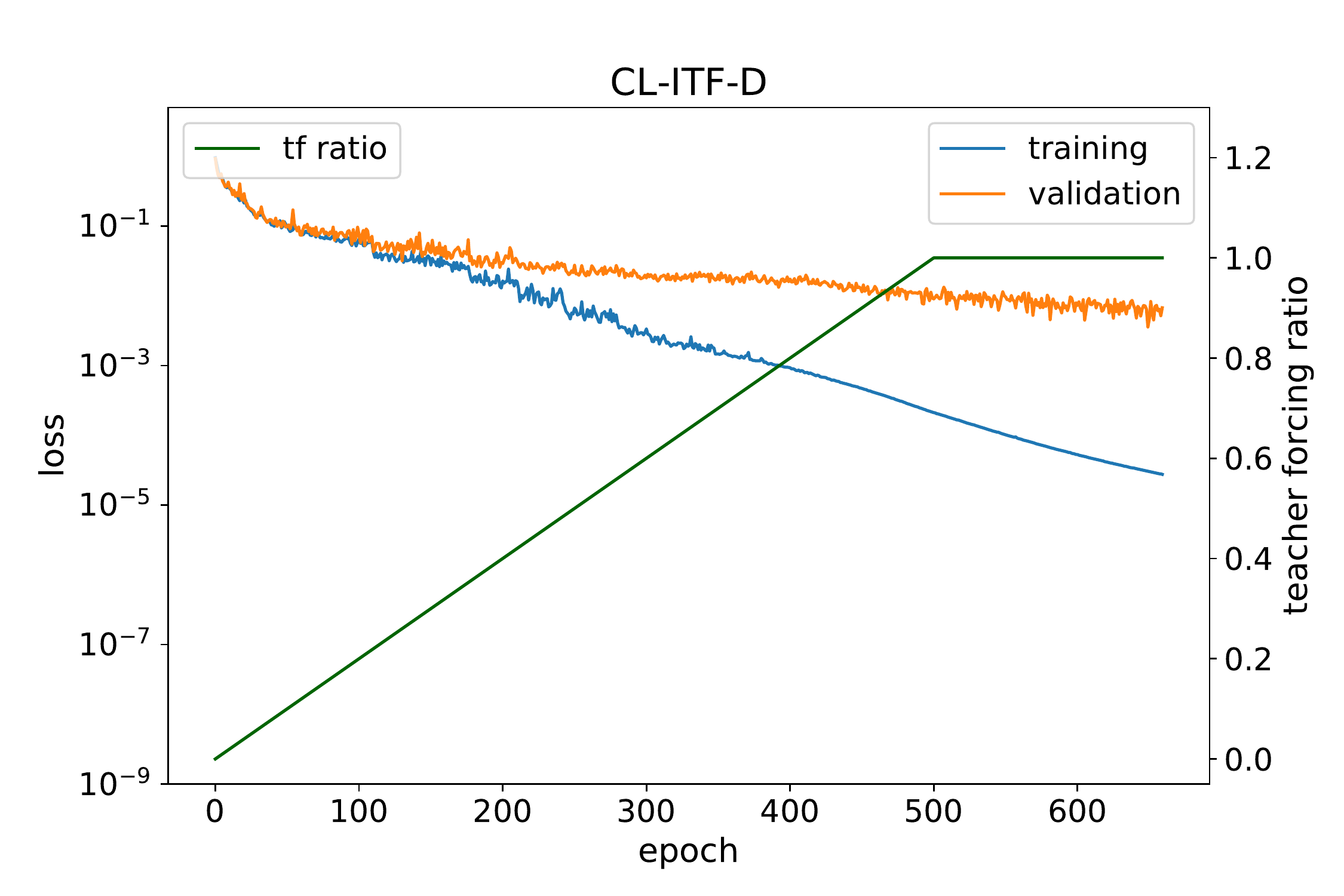}
		\caption{}
	\end{subfigure}
	
	\caption{Training and validation loss for Thomas}
	\label{fig:apx_thomas_loss_curves}
\end{figure*}

\begin{figure*}[ph]
	\centering
	\begin{subfigure}{0.40\linewidth}
		\includegraphics[width=\linewidth]{images/loss_plots/roessler_tf}
		\caption{}
	\end{subfigure}
	\begin{subfigure}{0.40\linewidth}
			\includegraphics[width=\linewidth]{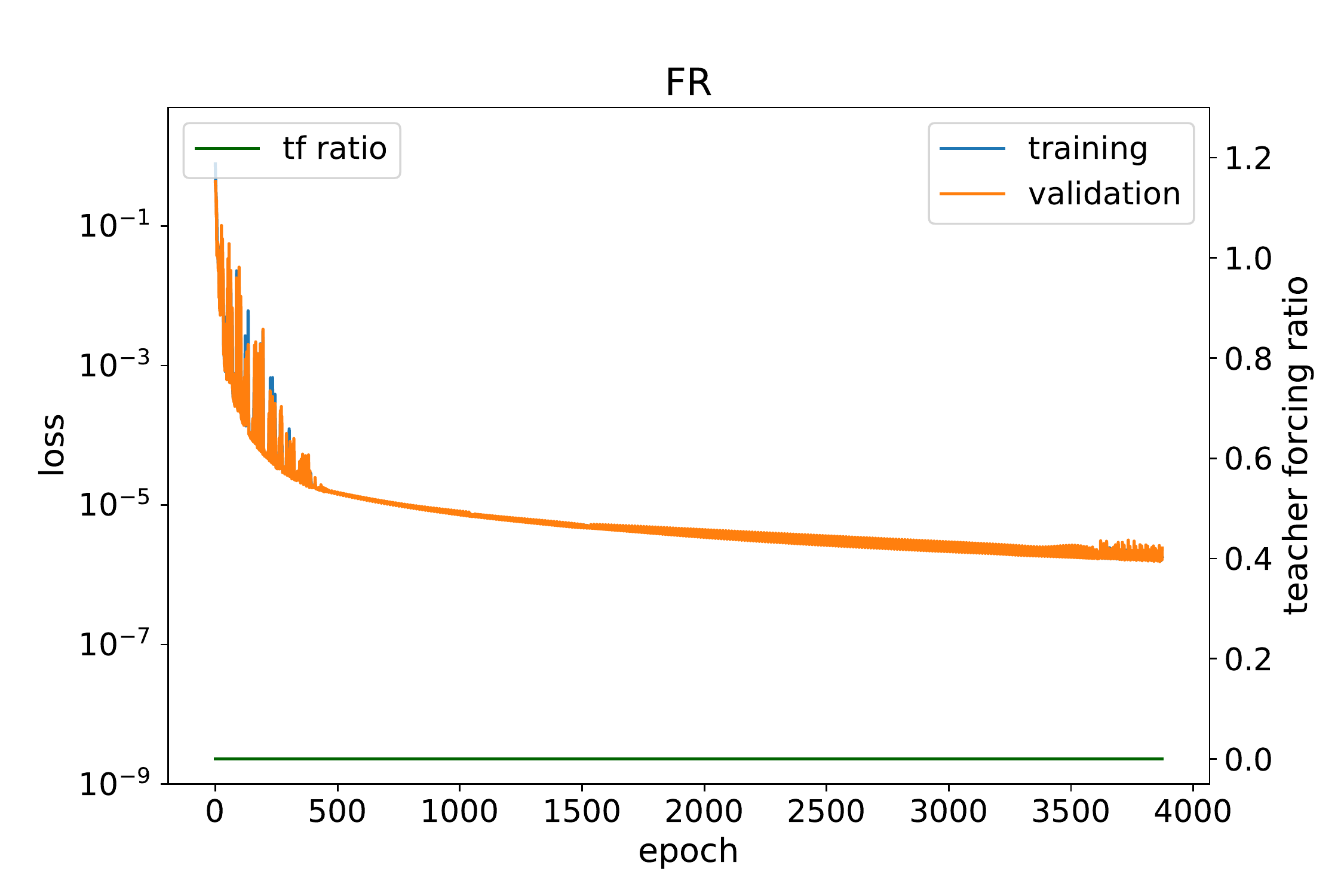}
		\caption{}
	\end{subfigure}
	
	\begin{subfigure}{0.40\linewidth}
		\includegraphics[width=\linewidth]{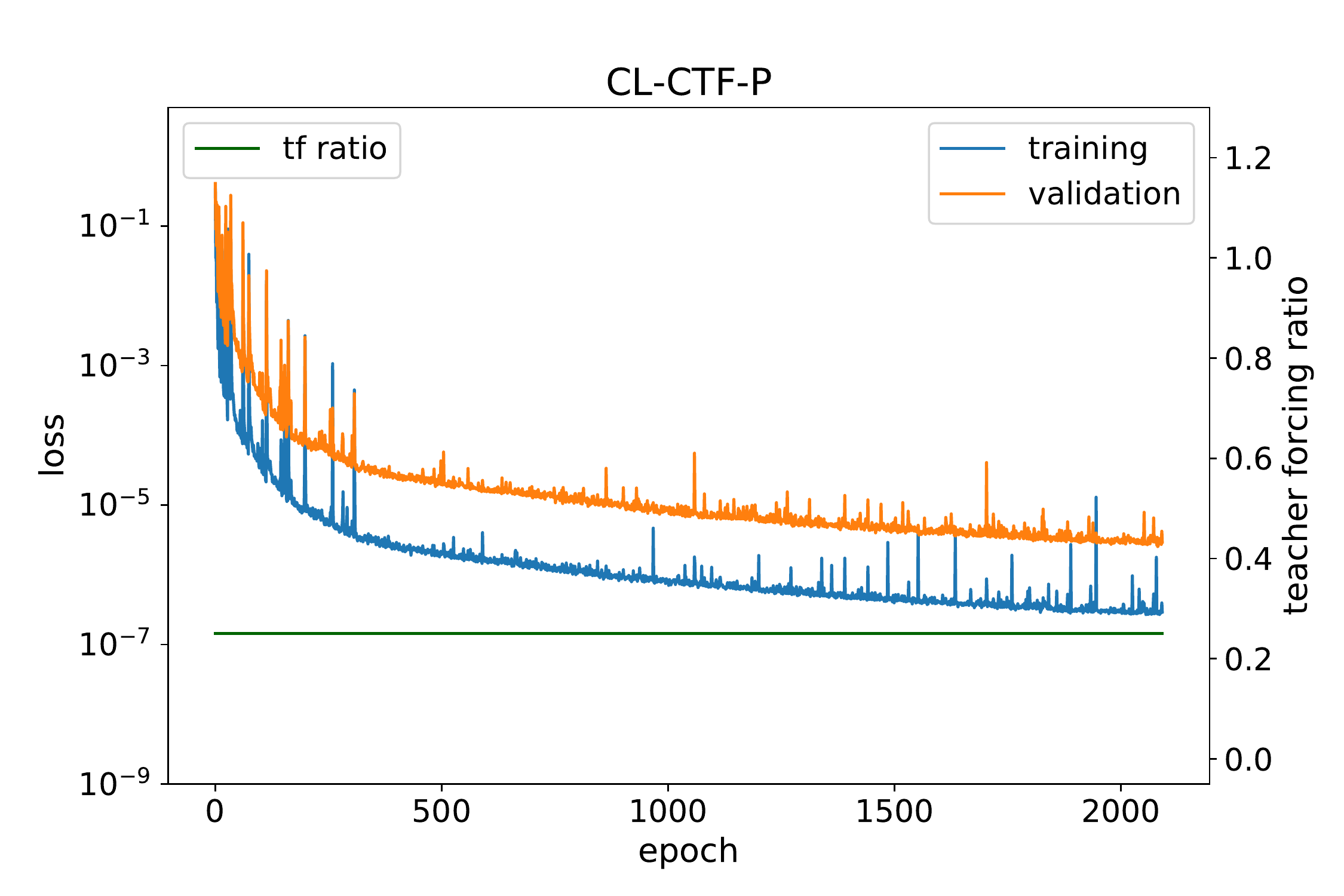}
		\caption{}
	\end{subfigure}

	\begin{subfigure}{0.40\linewidth}
		\includegraphics[width=\linewidth]{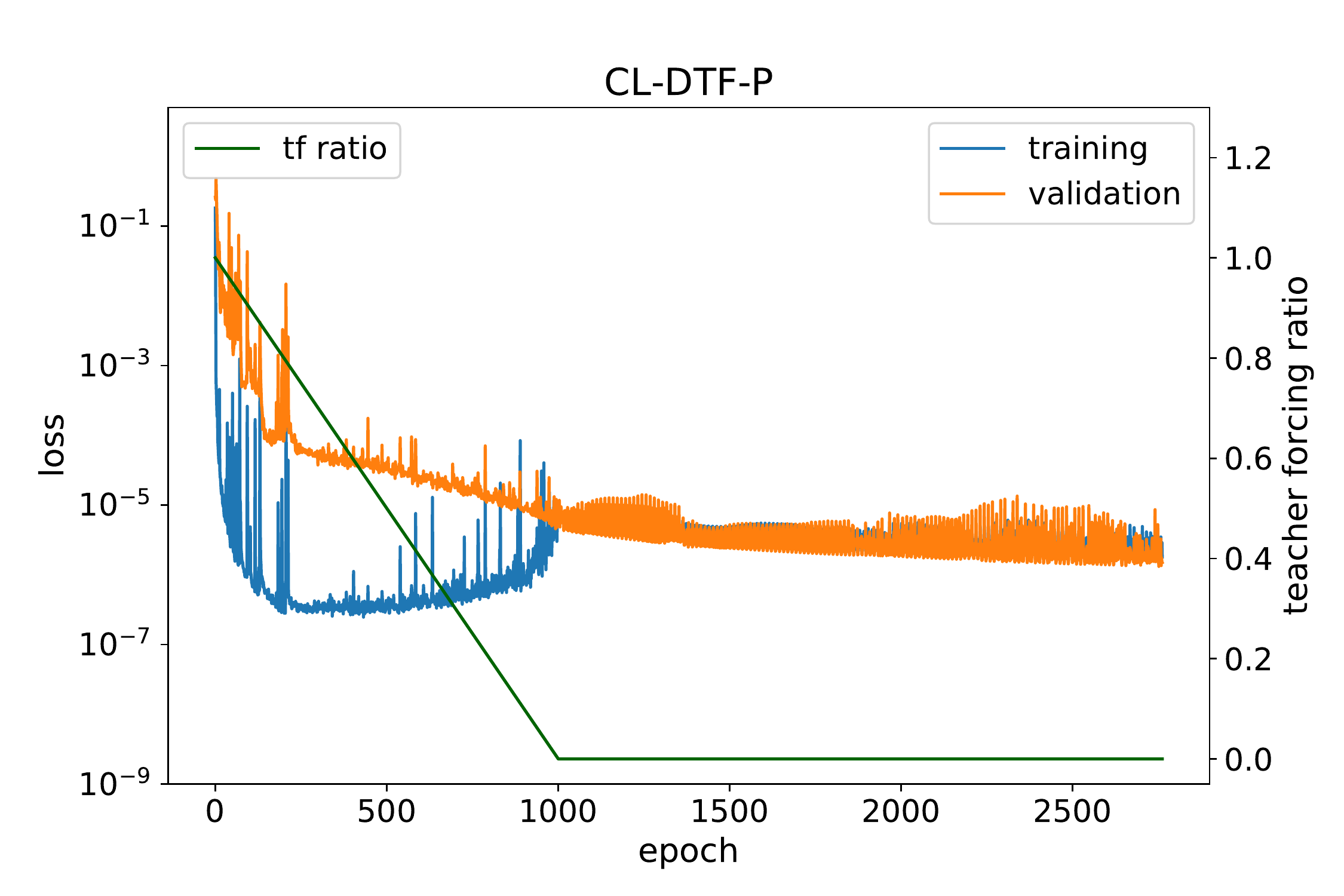}
		\caption{}
	\end{subfigure}
	\begin{subfigure}{0.40\linewidth}
		\includegraphics[width=\linewidth]{images/loss_plots/roessler_cl_itf_p}
		\caption{}
	\end{subfigure}

	\begin{subfigure}{0.40\linewidth}
		\includegraphics[width=\linewidth]{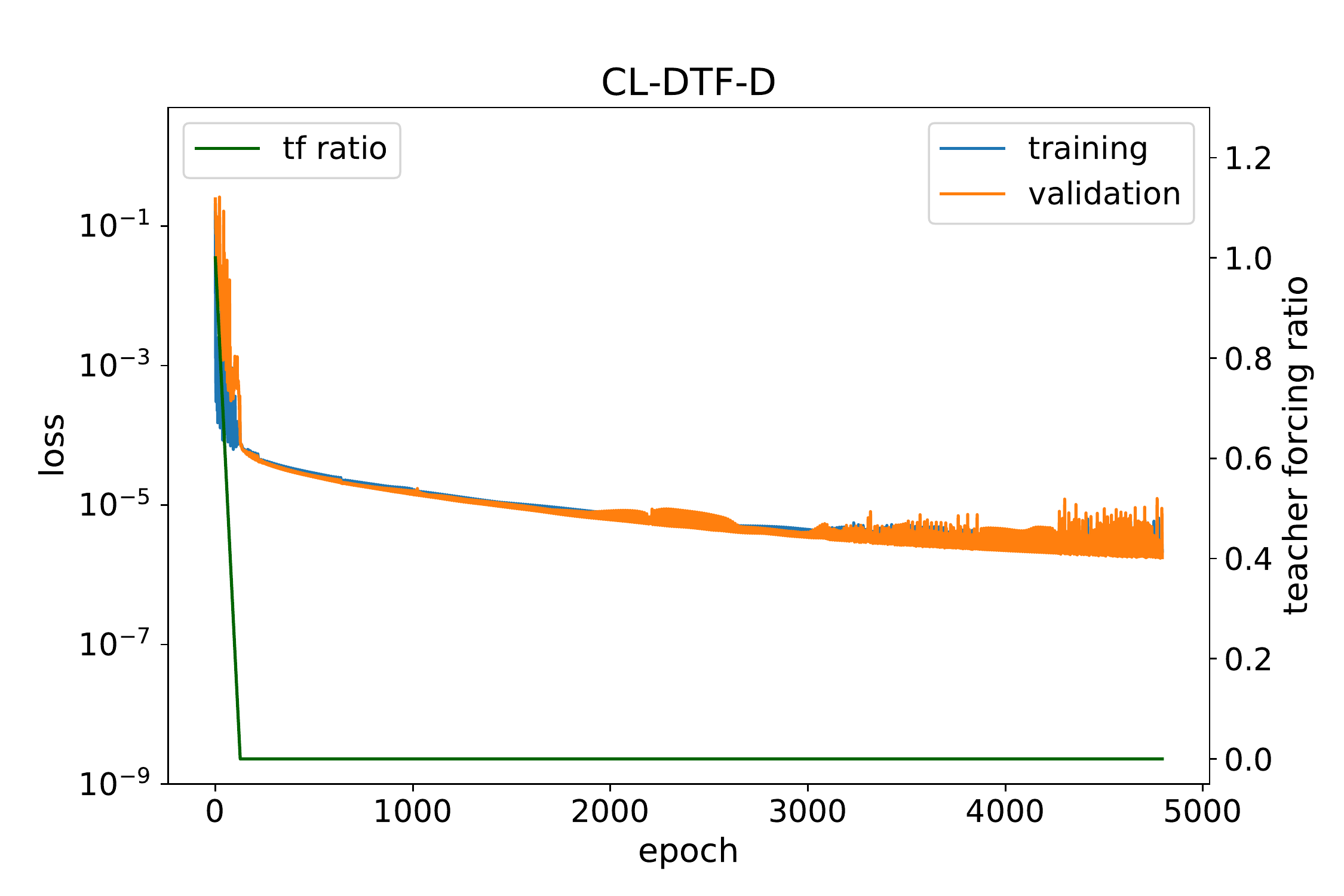}
		\caption{}
	\end{subfigure}
	\begin{subfigure}{0.40\linewidth}
		\includegraphics[width=\linewidth]{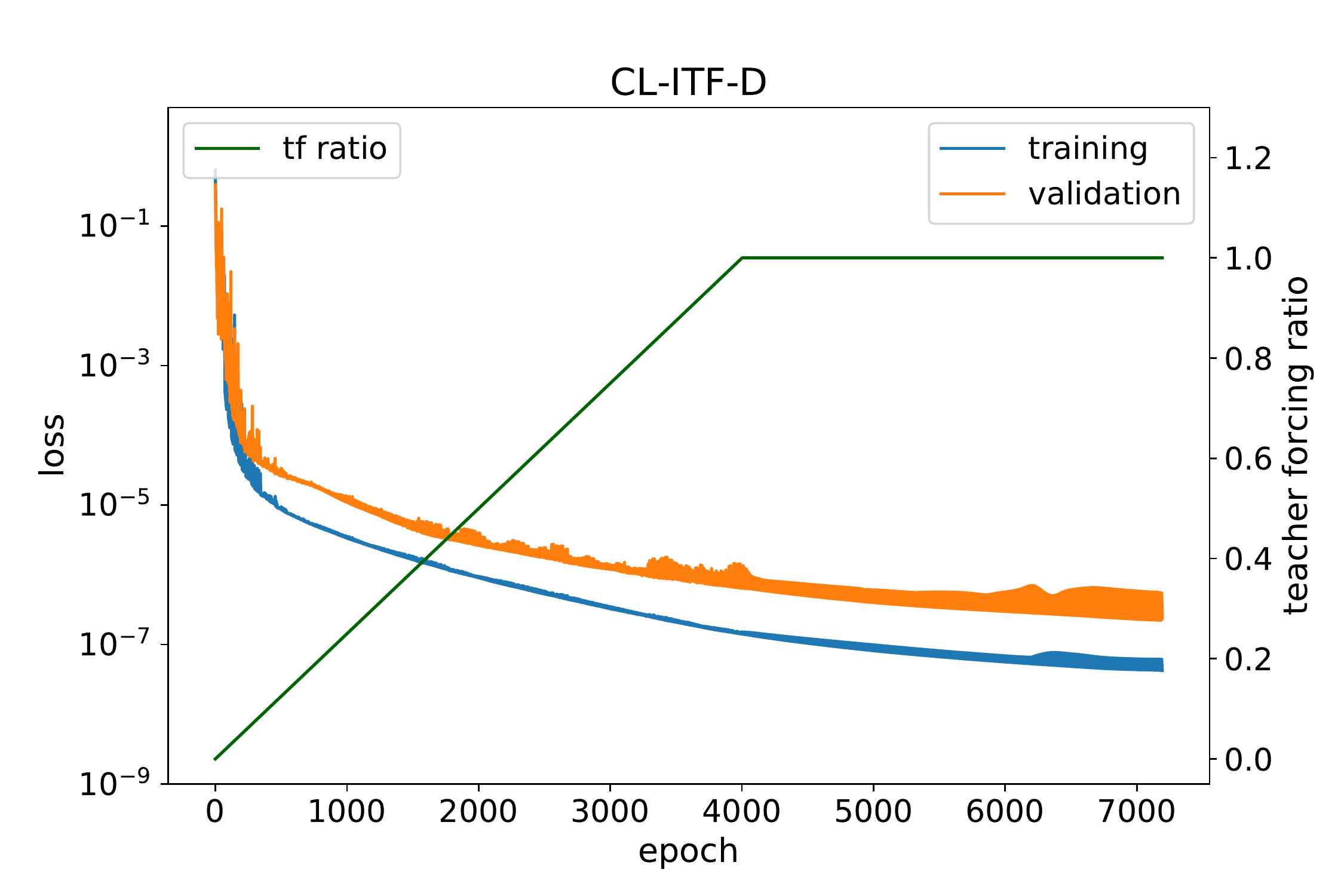}
		\caption{}
	\end{subfigure}
	
	\caption{Training and validation loss for Rössler}
	\label{fig:apx_roessler_loss_curves}
\end{figure*}

\begin{figure*}[ph]
	\centering
	\begin{subfigure}{0.40\linewidth}
		\includegraphics[width=\linewidth]{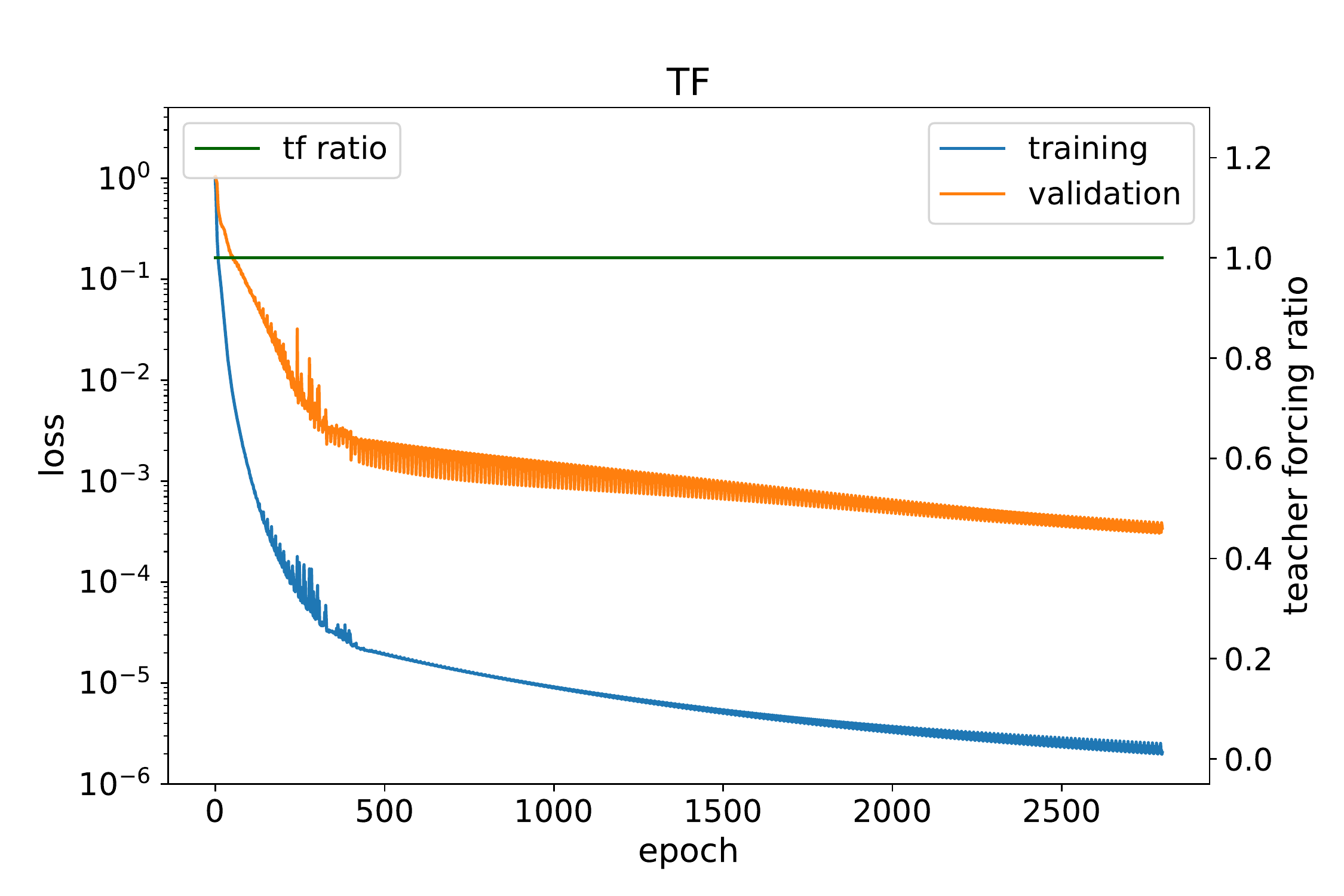}
		\caption{}
	\end{subfigure}
	\begin{subfigure}{0.40\linewidth}
			\includegraphics[width=\linewidth]{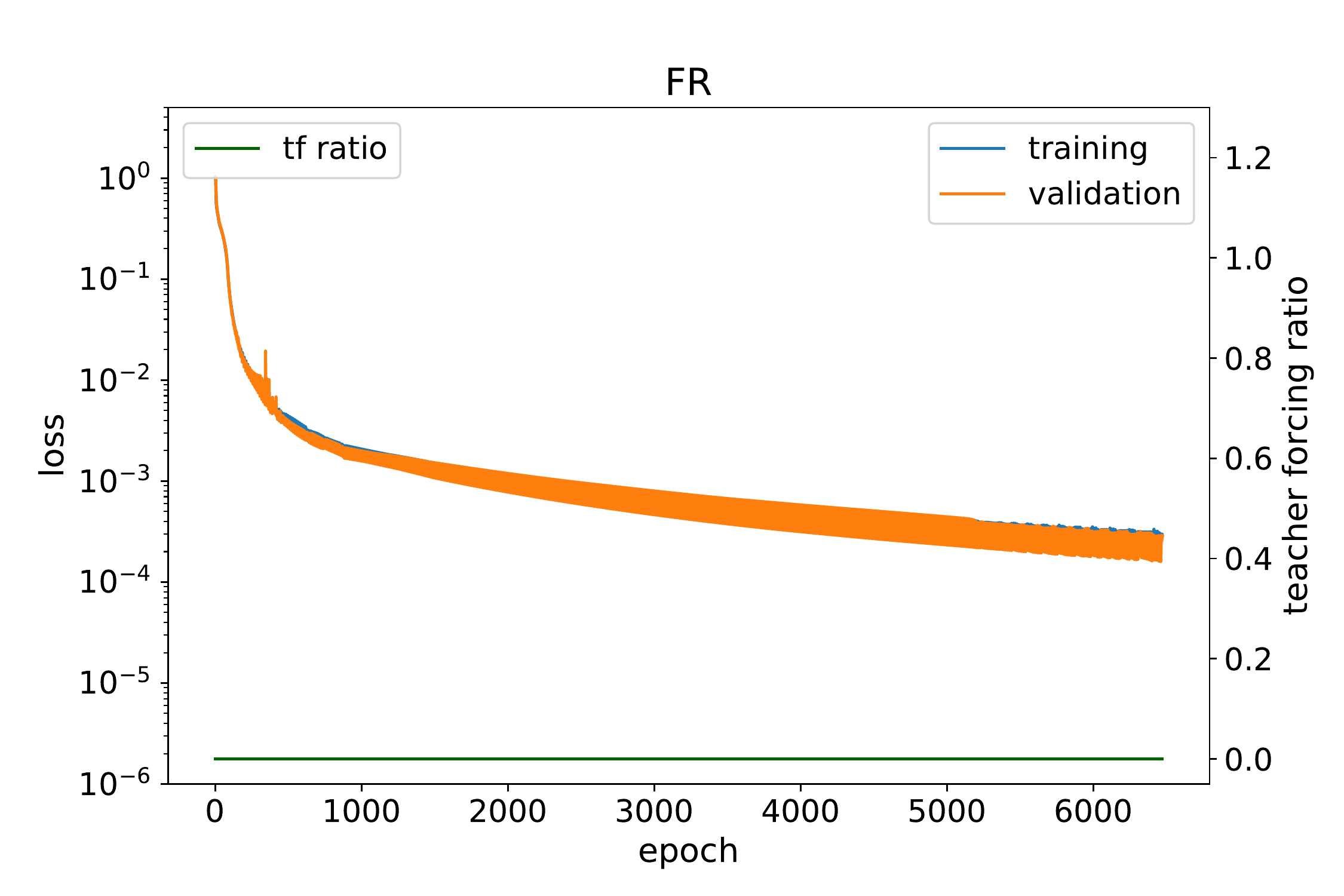}
		\caption{}
	\end{subfigure}
	
	\begin{subfigure}{0.40\linewidth}
		\includegraphics[width=\linewidth]{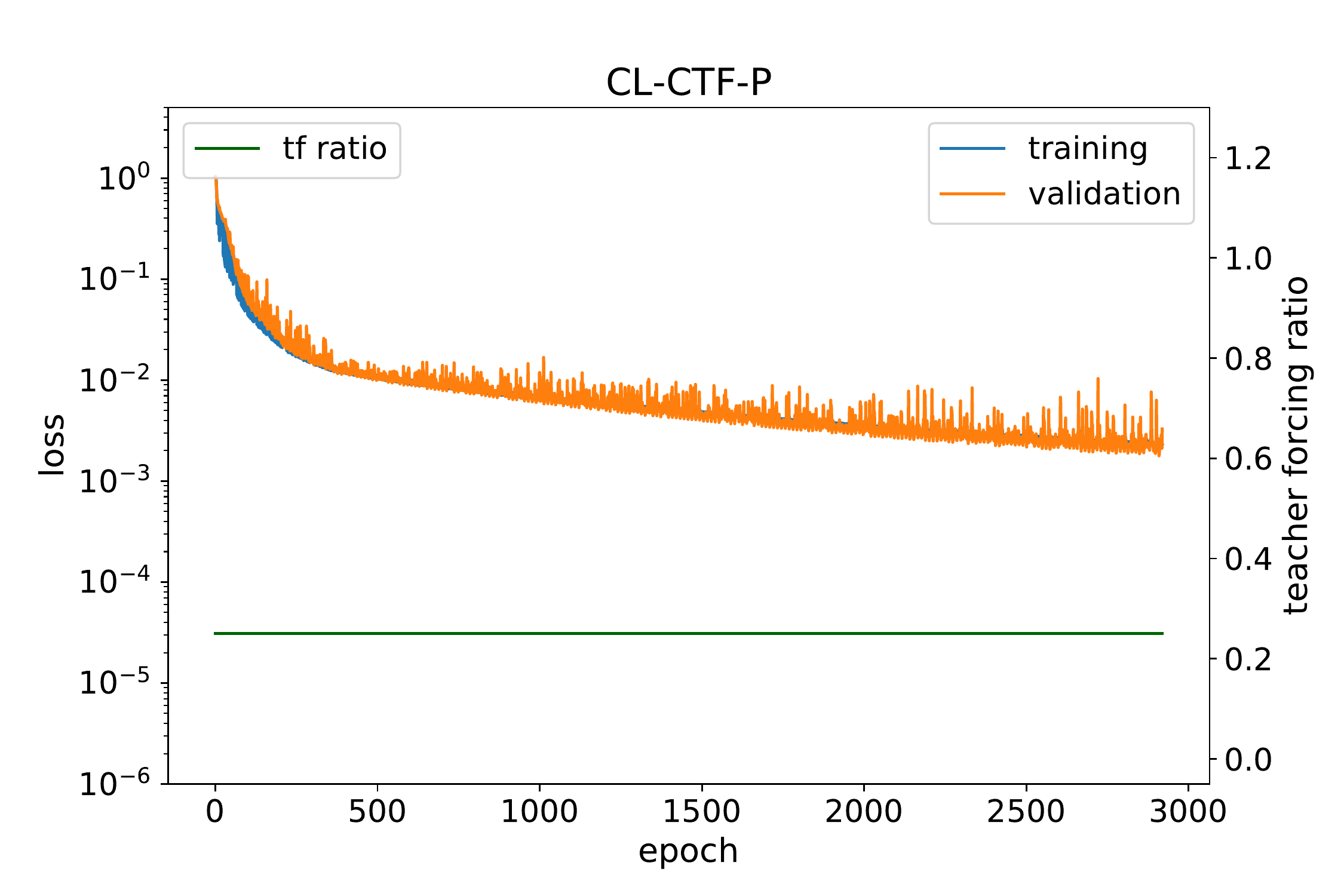}
		\caption{}
	\end{subfigure}

	\begin{subfigure}{0.40\linewidth}
		\includegraphics[width=\linewidth]{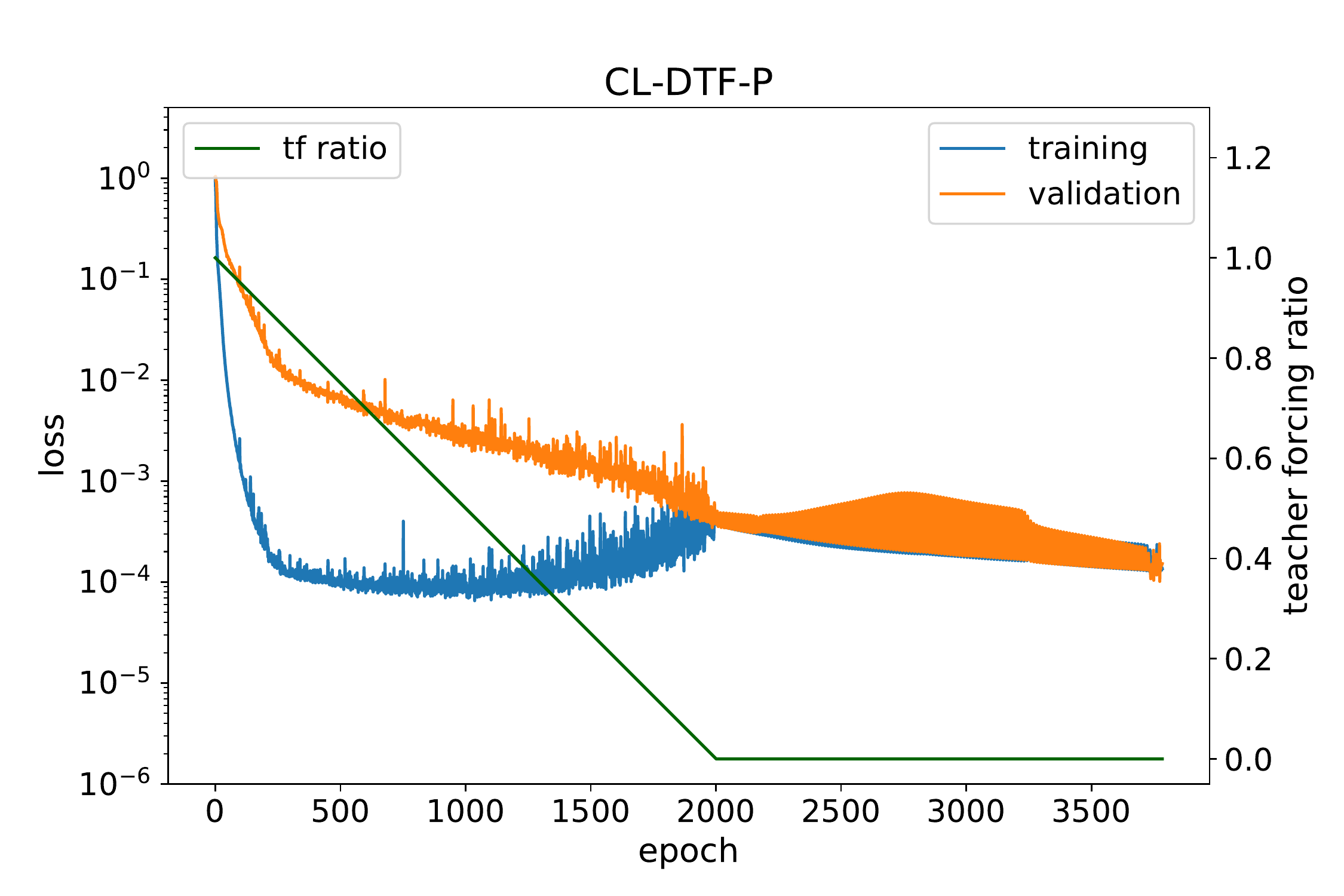}
		\caption{}
	\end{subfigure}
	\begin{subfigure}{0.40\linewidth}
		\includegraphics[width=\linewidth]{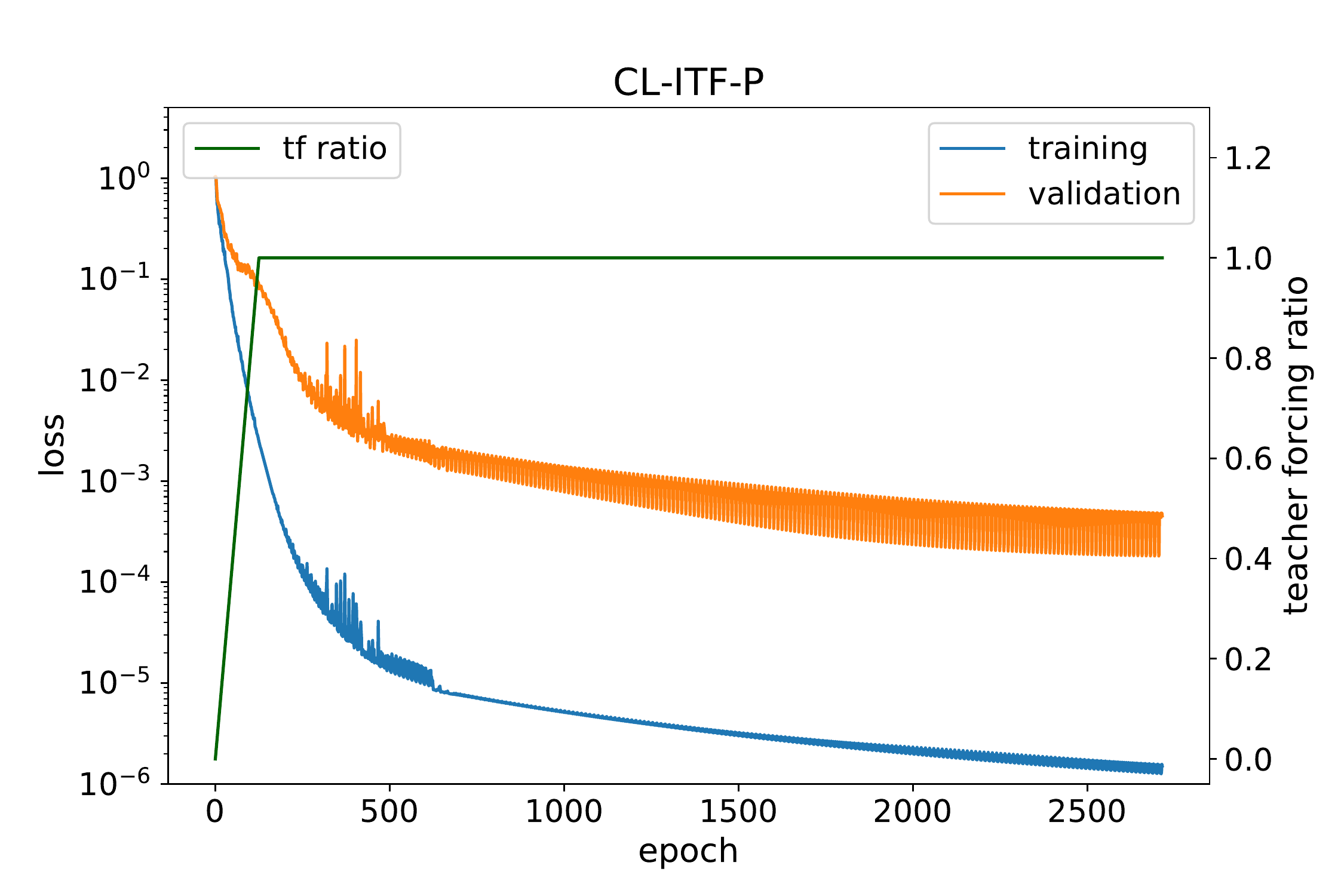}
		\caption{}
	\end{subfigure}

	\begin{subfigure}{0.40\linewidth}
		\includegraphics[width=\linewidth]{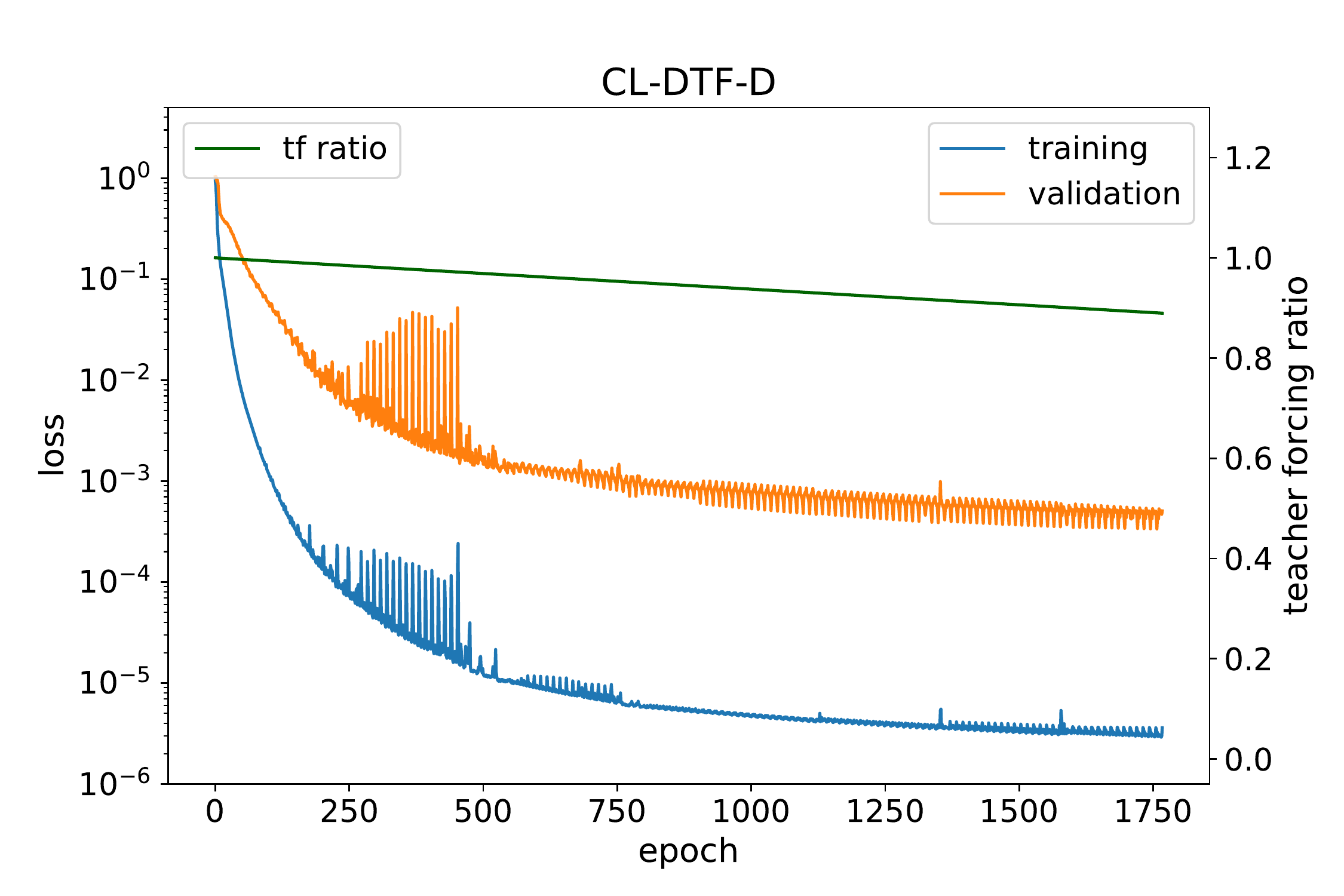}
		\caption{}
	\end{subfigure}
	\begin{subfigure}{0.40\linewidth}
		\includegraphics[width=\linewidth]{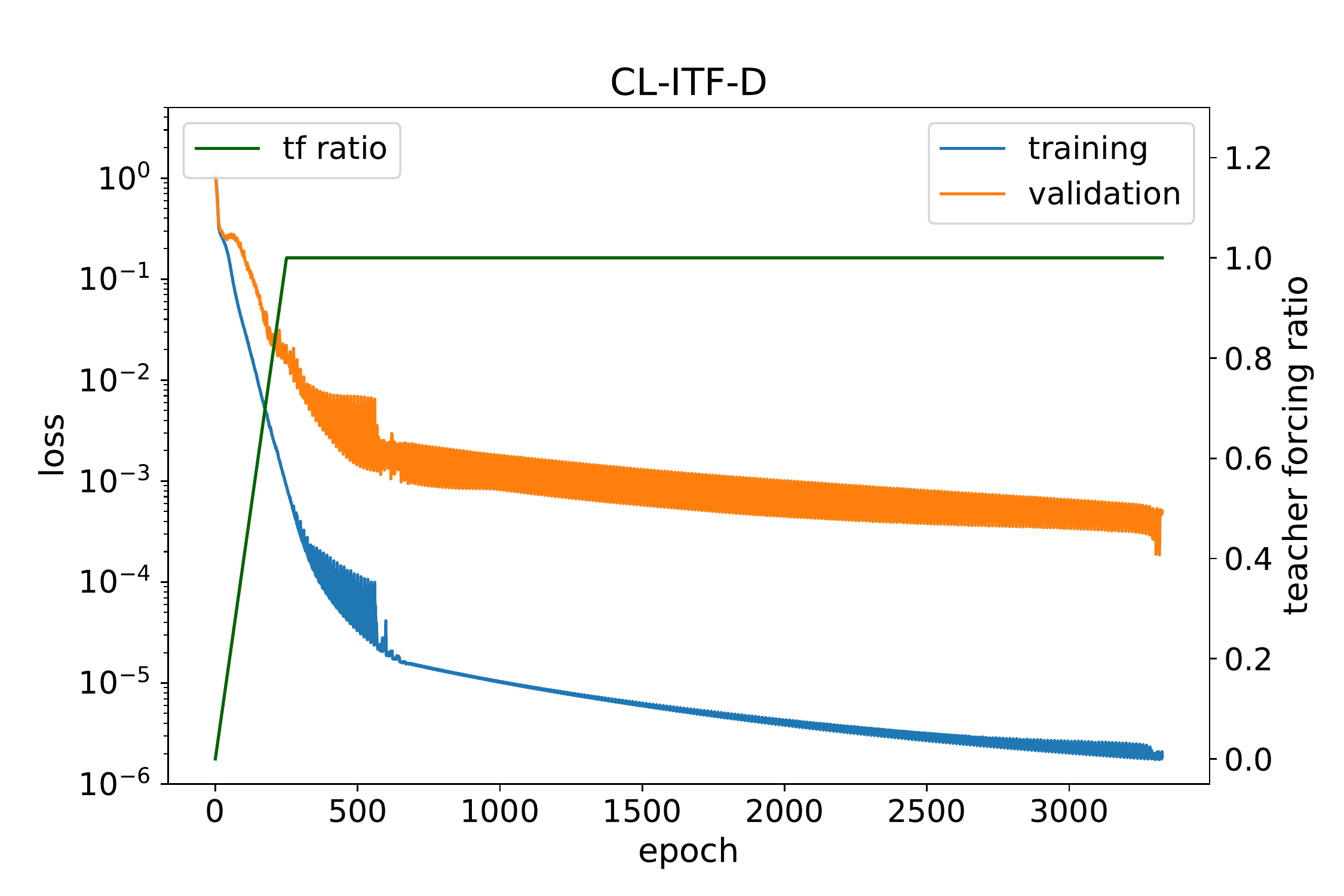}
		\caption{}
	\end{subfigure}
	
	\caption{Training and validation loss for Hyper-Rössler}
	\label{fig:apx_hyperroessler_loss_curves}
\end{figure*}

\begin{figure*}[ph]
	\centering
	\begin{subfigure}{0.40\linewidth}
		\includegraphics[width=\linewidth]{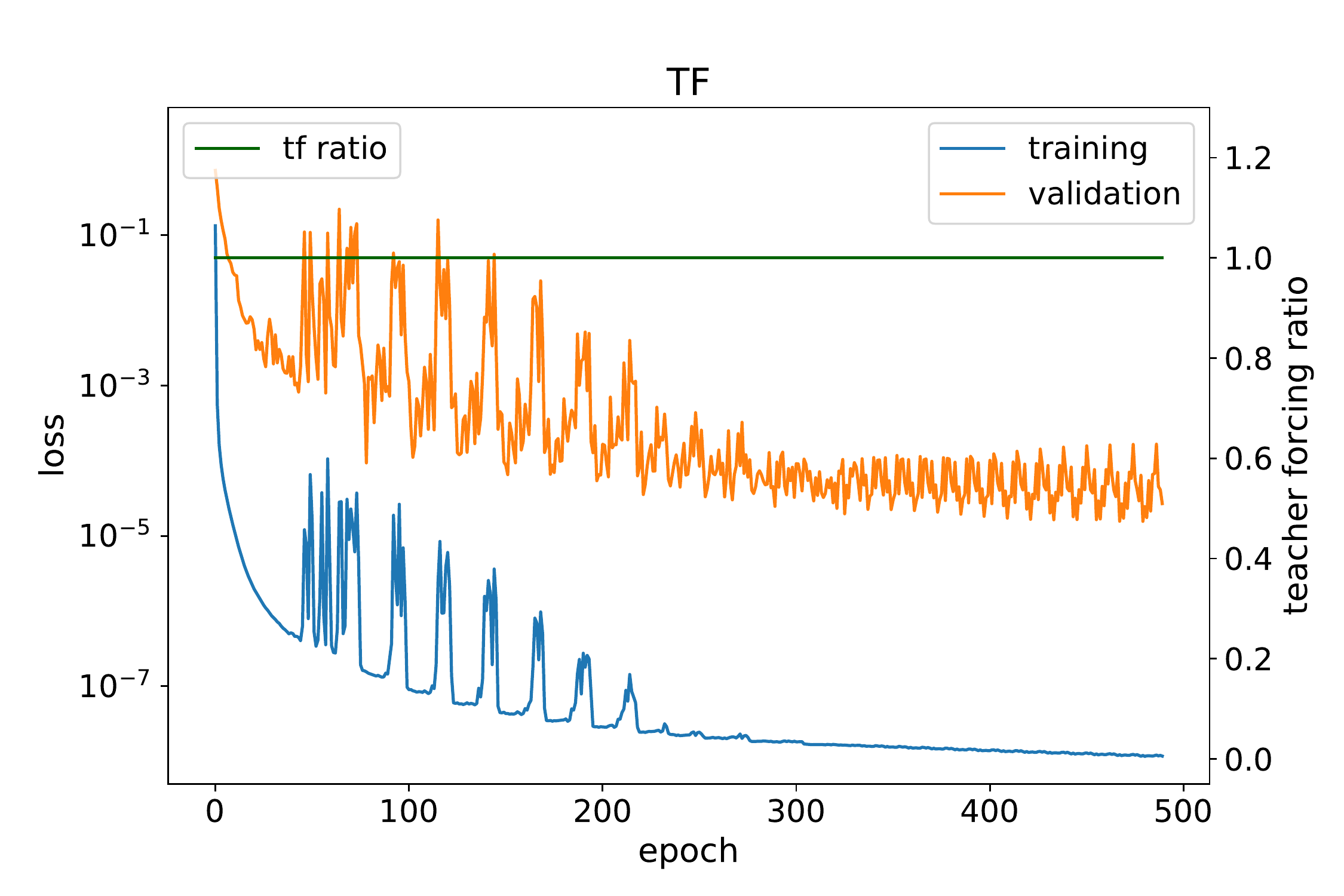}
		\caption{}
	\end{subfigure}
	\begin{subfigure}{0.40\linewidth}
			\includegraphics[width=\linewidth]{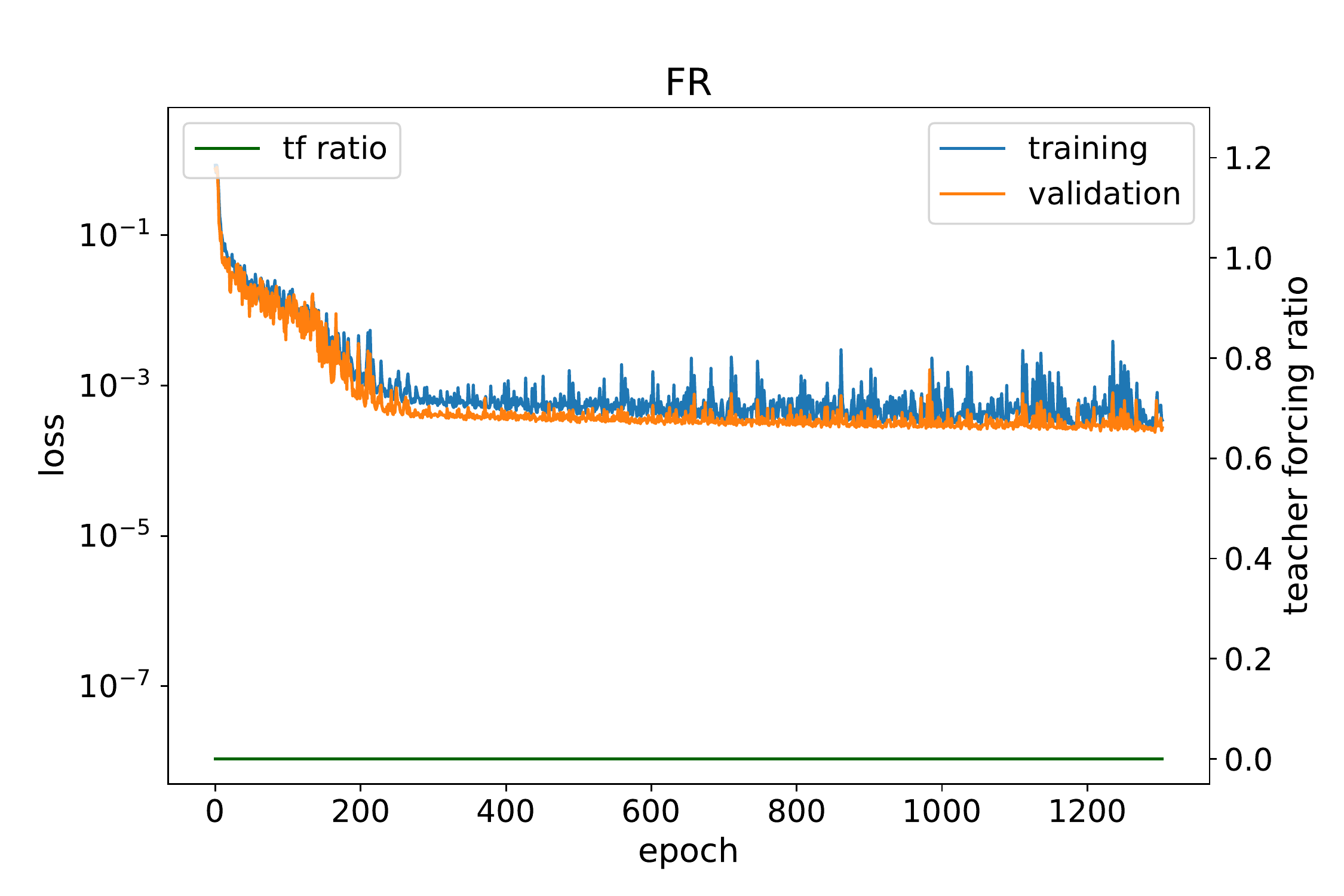}
		\caption{}
	\end{subfigure}
	
	\begin{subfigure}{0.40\linewidth}
		\includegraphics[width=\linewidth]{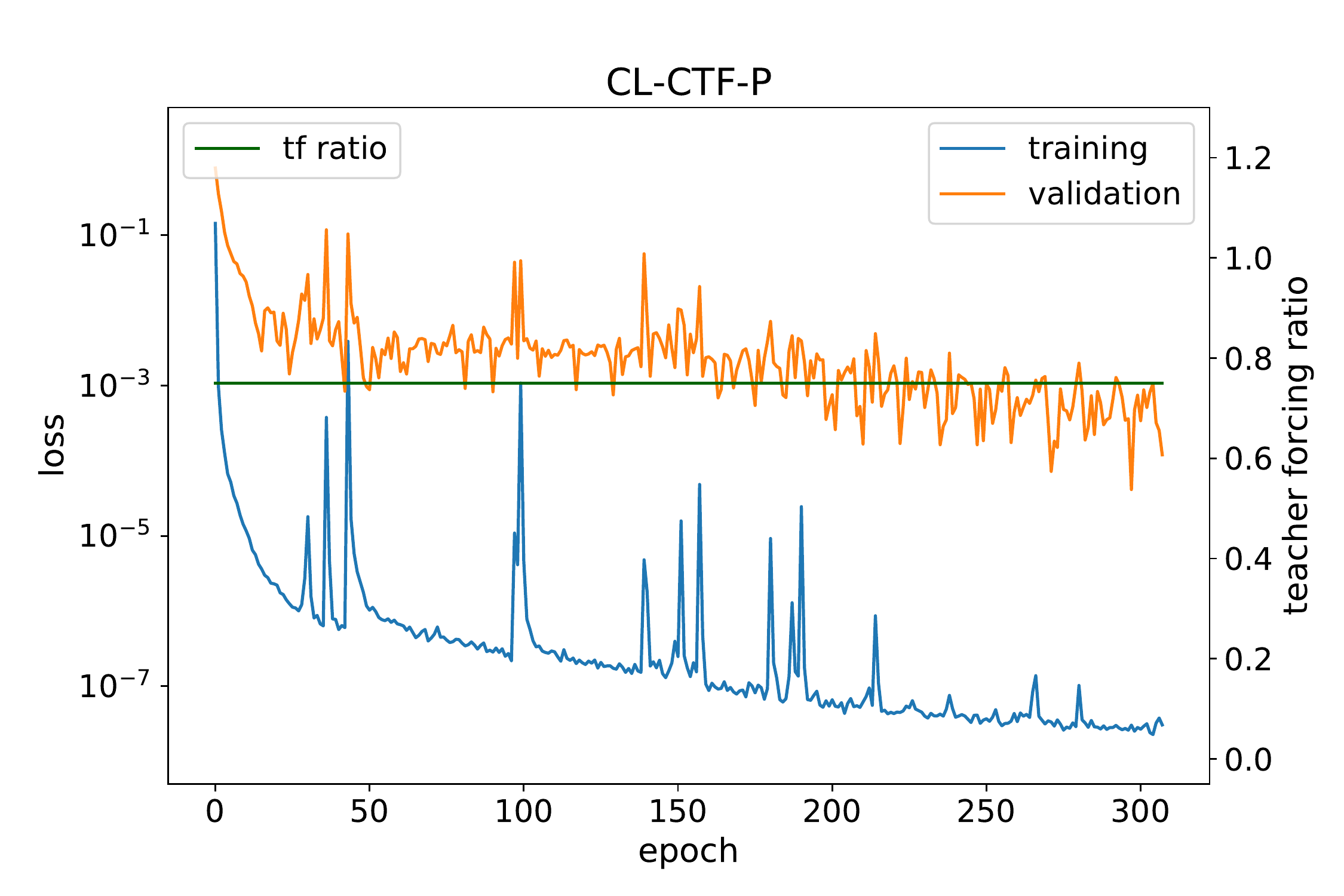}
		\caption{}
	\end{subfigure}

	\begin{subfigure}{0.40\linewidth}
		\includegraphics[width=\linewidth]{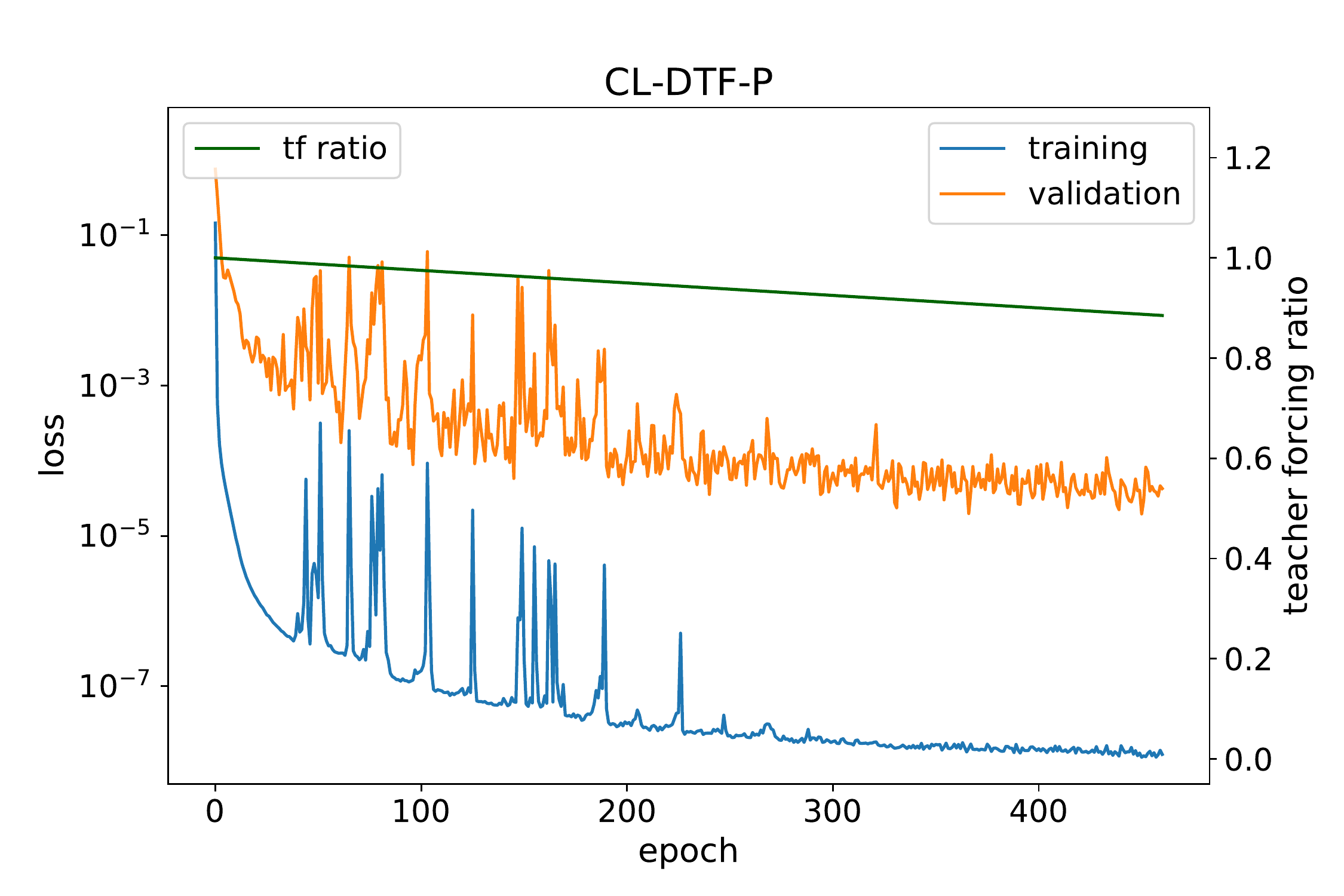}
		\caption{}
	\end{subfigure}
	\begin{subfigure}{0.40\linewidth}
		\includegraphics[width=\linewidth]{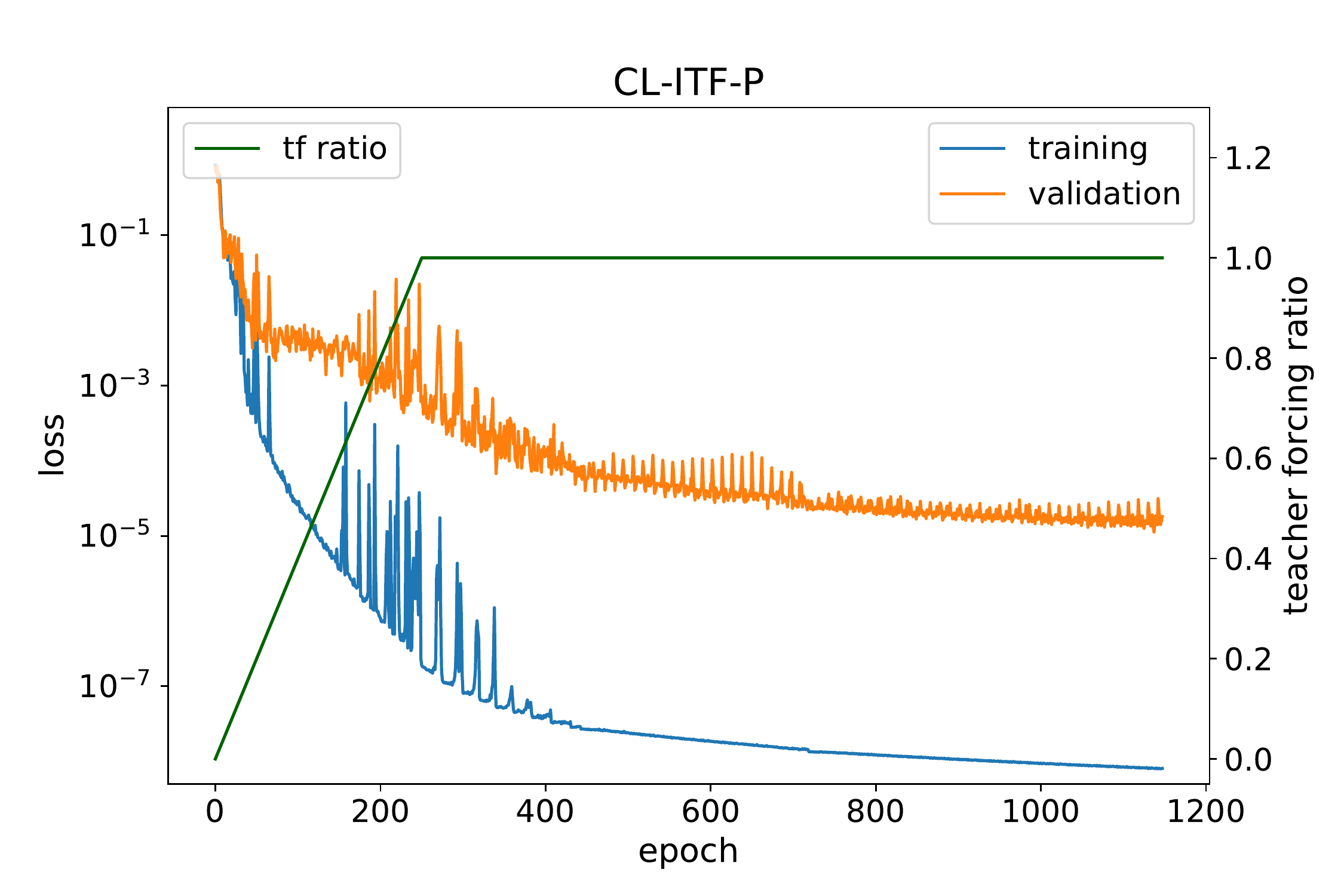}
		\caption{}
	\end{subfigure}

	\begin{subfigure}{0.40\linewidth}
		\includegraphics[width=\linewidth]{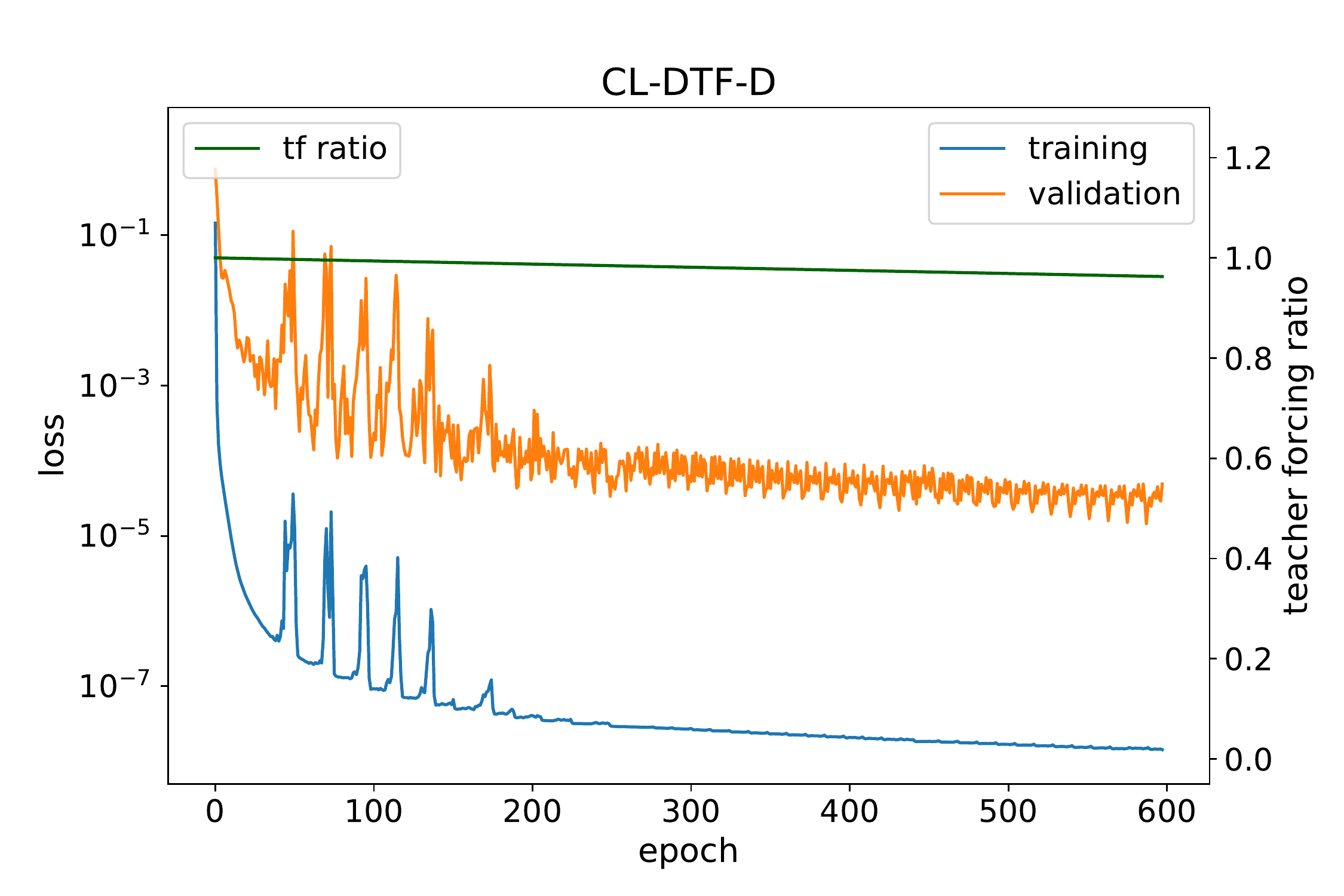}
		\caption{}
	\end{subfigure}
	\begin{subfigure}{0.40\linewidth}
		\includegraphics[width=\linewidth]{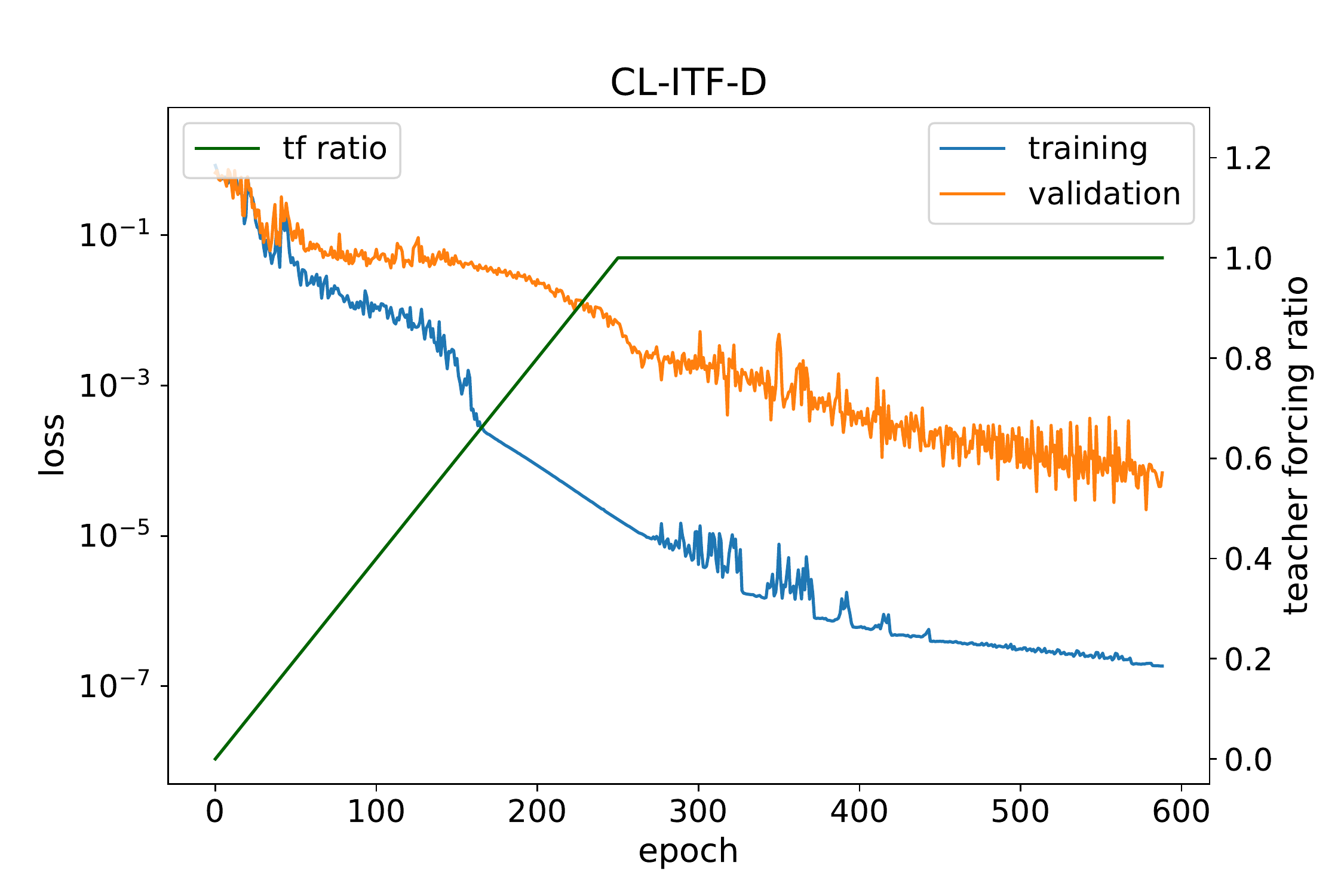}
		\caption{}
	\end{subfigure}
	
	\caption{Training and validation loss for Lorenz}
	\label{fig:apx_lorenz_loss_curves}
\end{figure*}

\begin{figure*}[ph]
	\centering
	\begin{subfigure}{0.40\linewidth}
		\includegraphics[width=\linewidth]{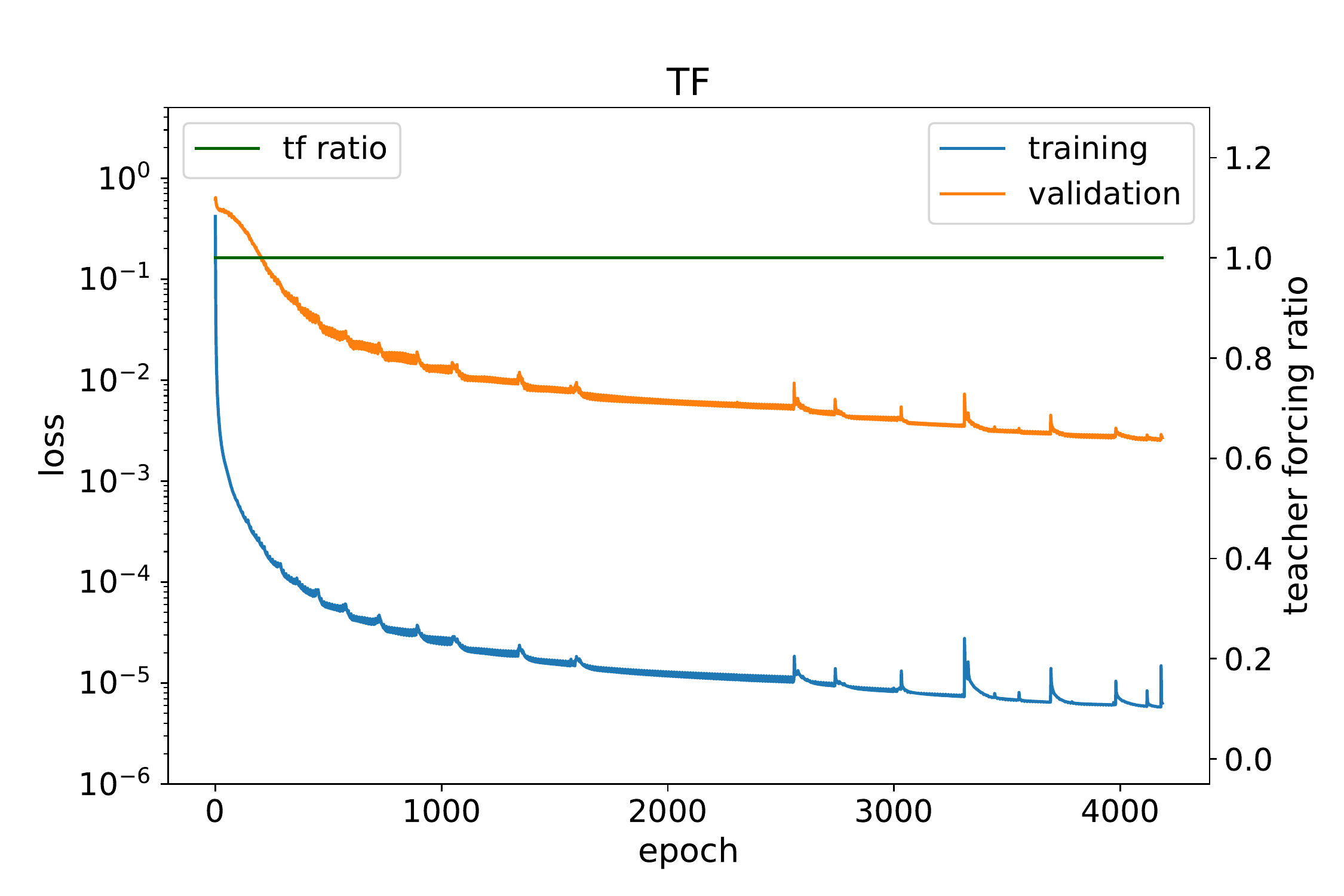}
		\caption{}
	\end{subfigure}
	\begin{subfigure}{0.40\linewidth}
			\includegraphics[width=\linewidth]{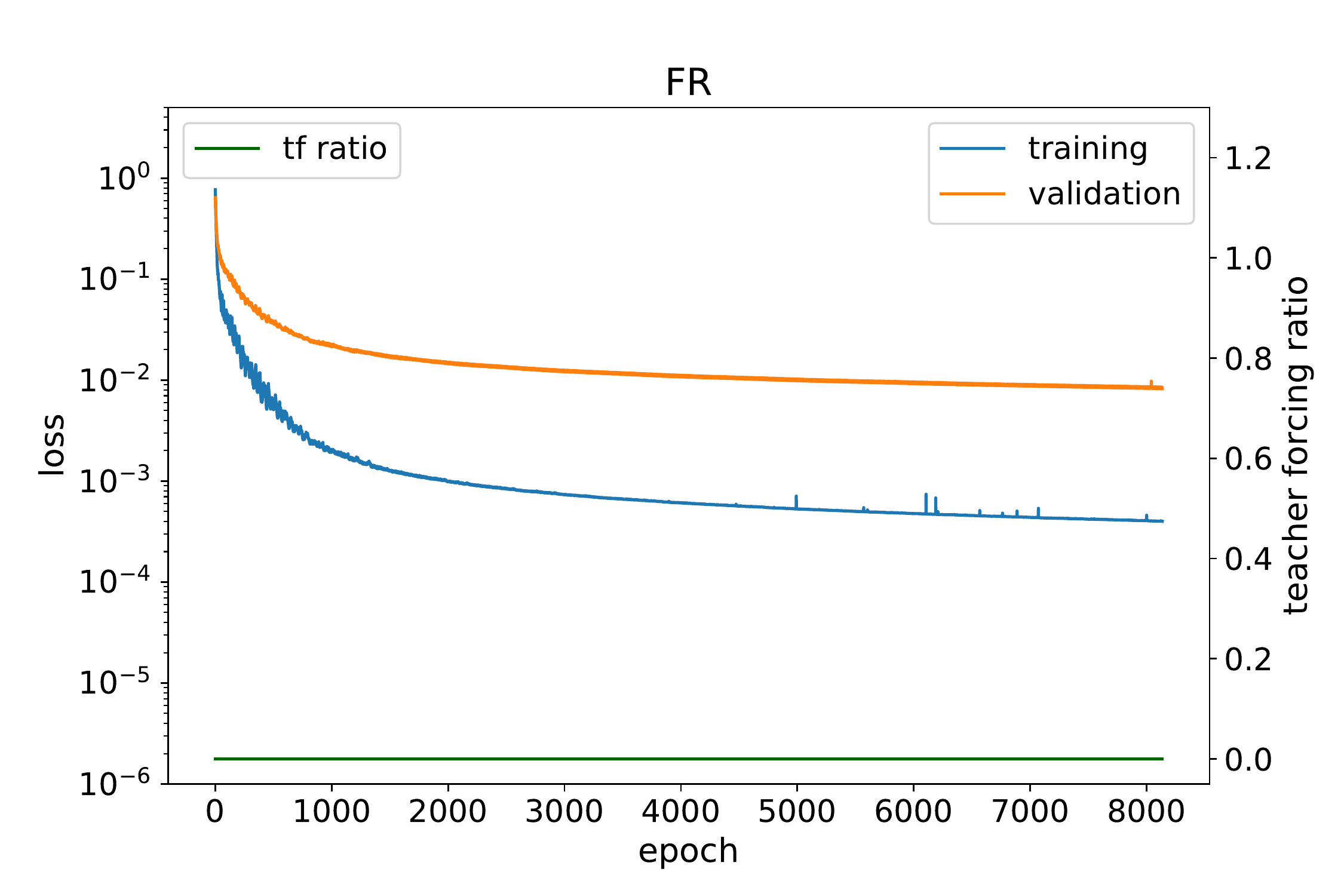}
		\caption{}
	\end{subfigure}
	
	\begin{subfigure}{0.40\linewidth}
		\includegraphics[width=\linewidth]{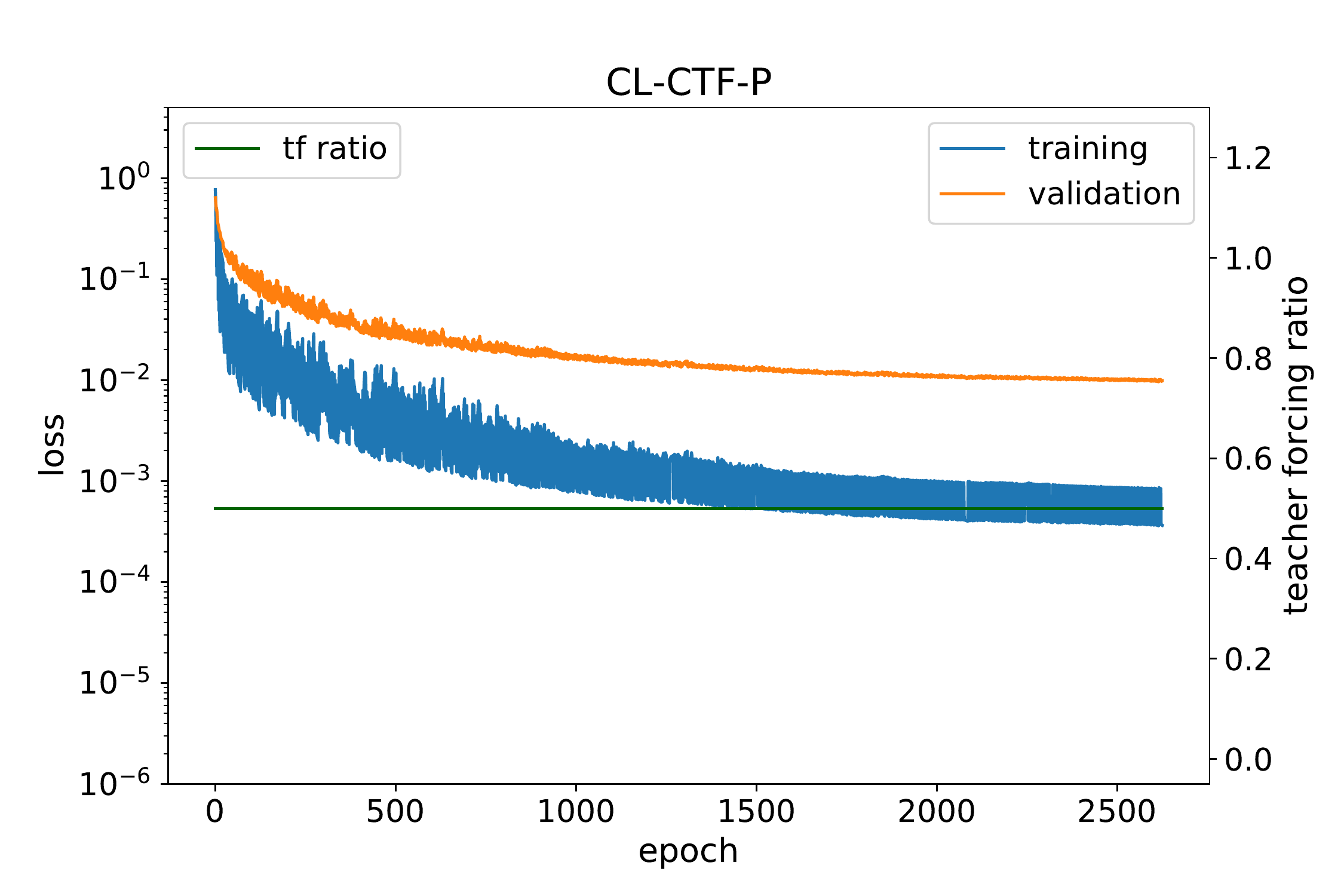}
		\caption{}
	\end{subfigure}

	\begin{subfigure}{0.40\linewidth}
		\includegraphics[width=\linewidth]{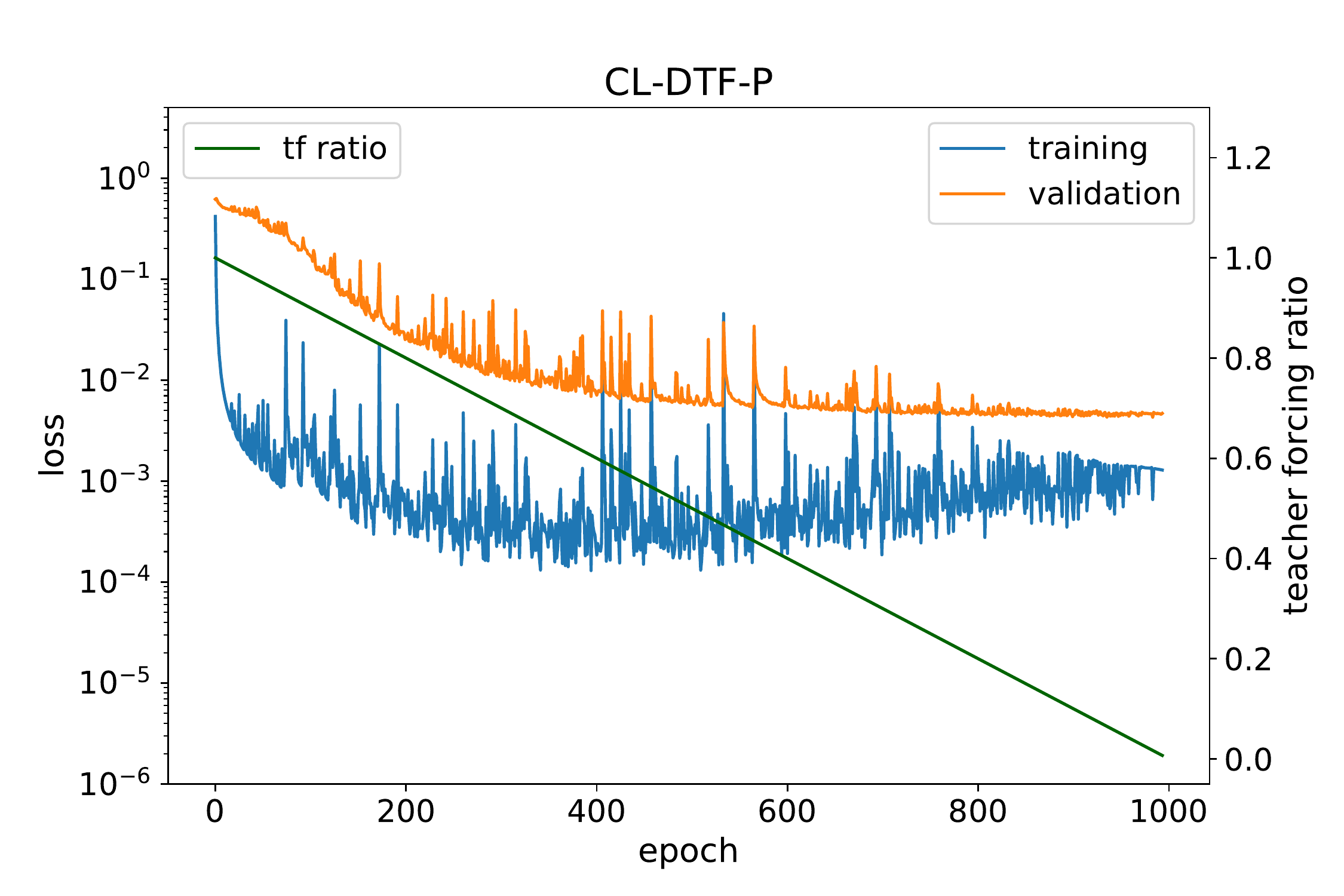}
		\caption{}
	\end{subfigure}
	\begin{subfigure}{0.40\linewidth}
		\includegraphics[width=\linewidth]{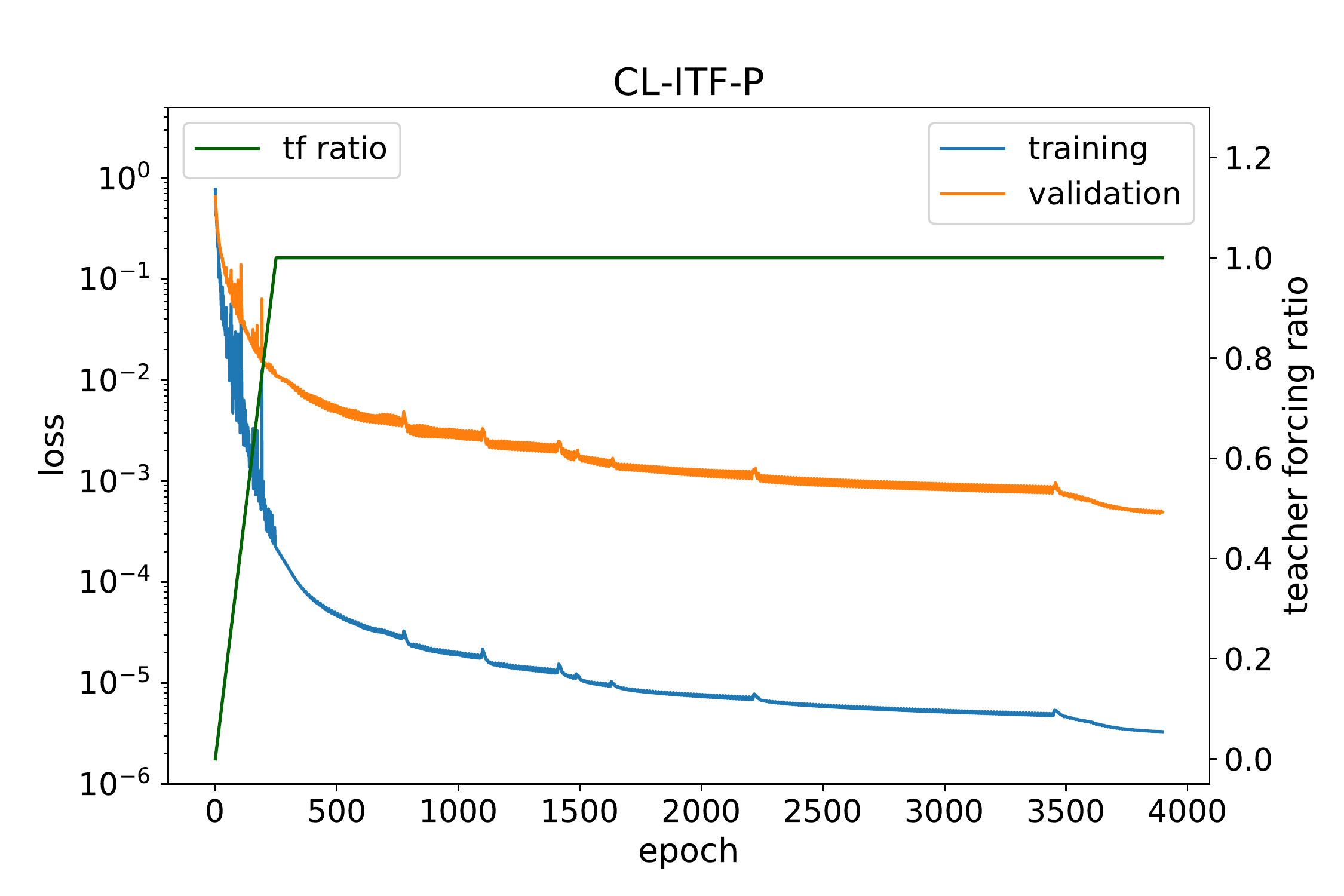}
		\caption{}
	\end{subfigure}

	\begin{subfigure}{0.409\linewidth}
		\includegraphics[width=\linewidth]{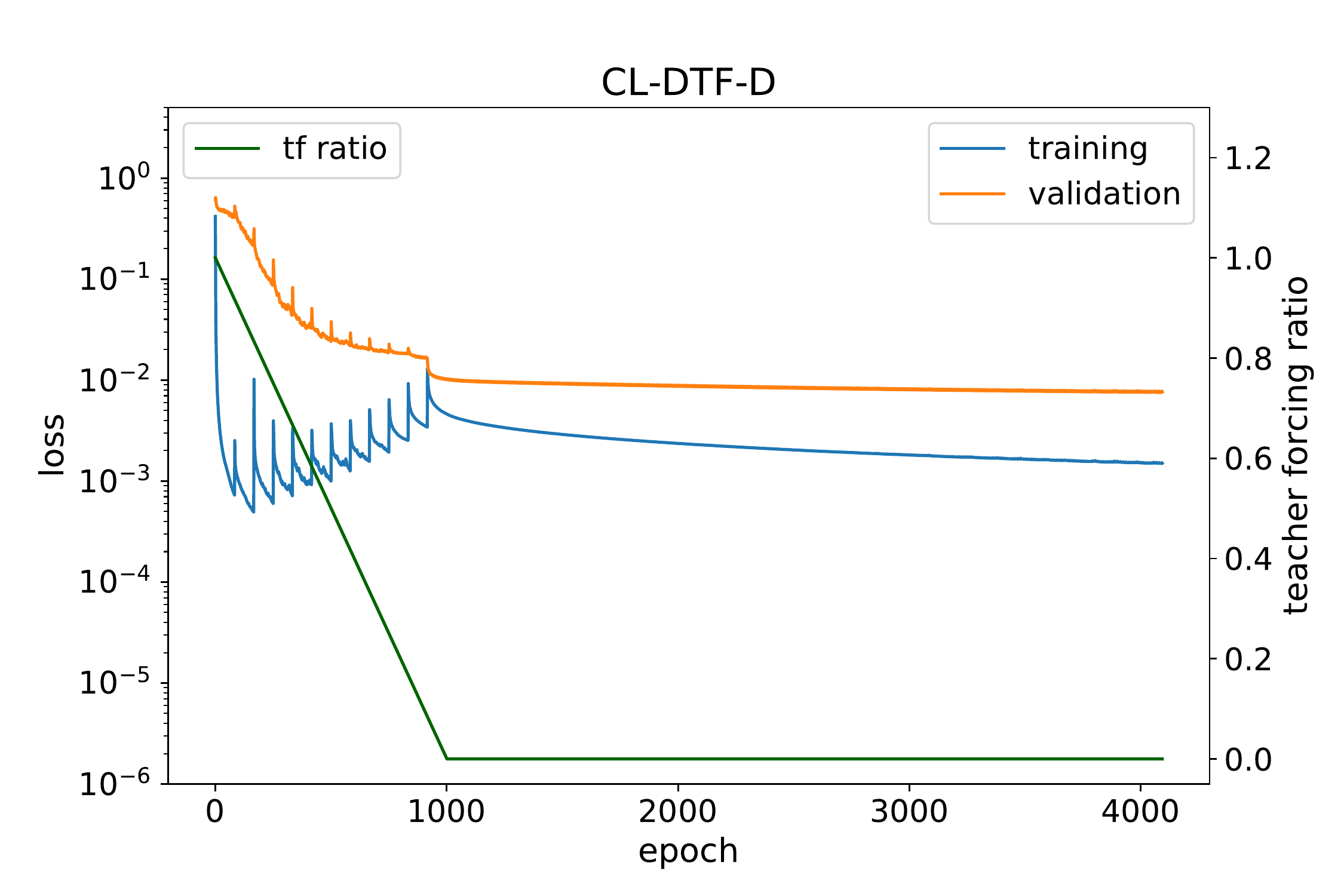}
		\caption{}
	\end{subfigure}
	\begin{subfigure}{0.40\linewidth}
		\includegraphics[width=\linewidth]{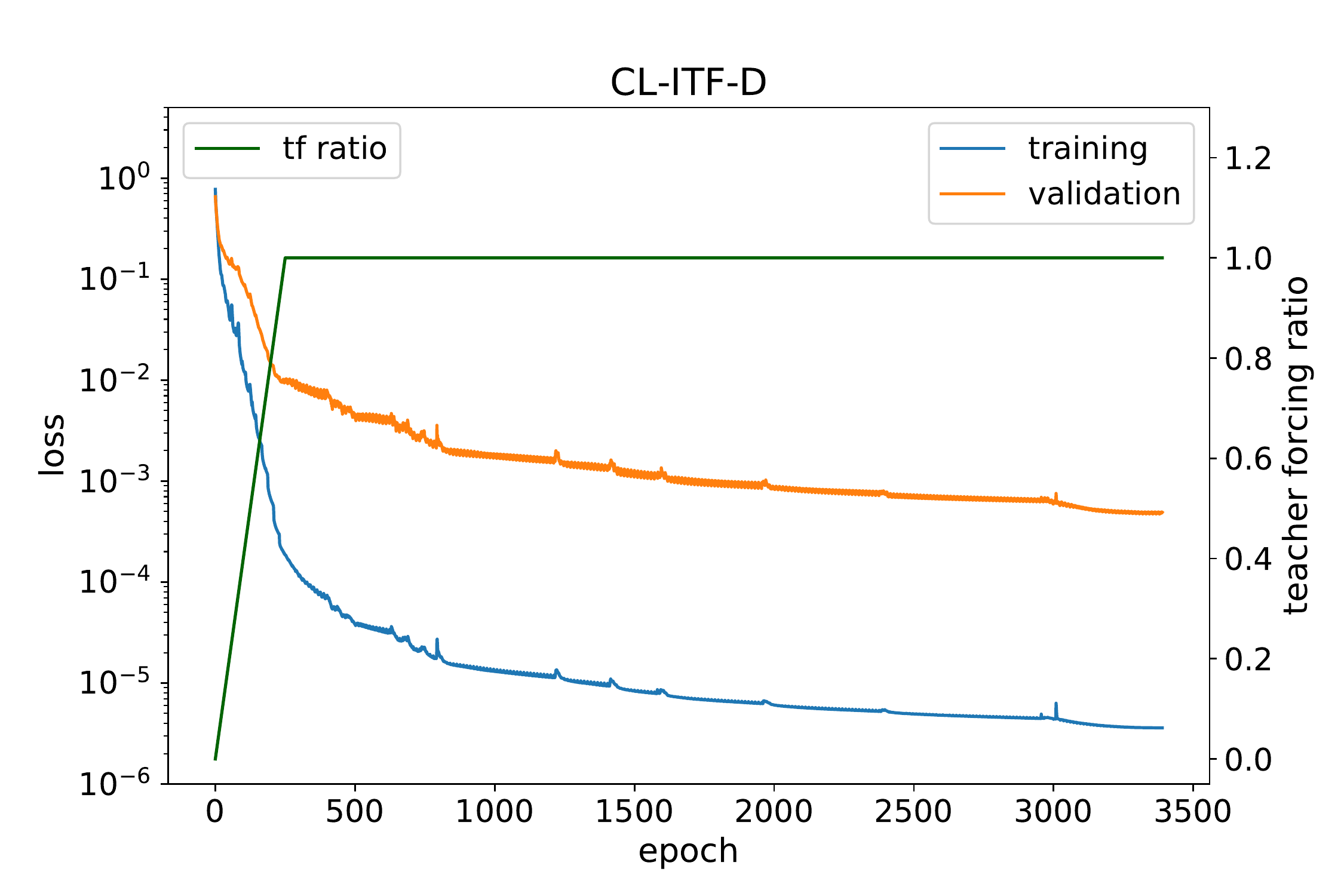}
		\caption{}
	\end{subfigure}
	
	\caption{Training and validation loss for Lorenz'96}
	\label{fig:apx_lorenz96_loss_curves}
\end{figure*}

\newpage

\section{CL-ITF-P With and Without LR Scheduler}
\label{apx:lr_scheduler}
\begin{figure*}[ph]
	\centering
	\begin{subfigure}{0.60\linewidth}
		\includegraphics[width=\linewidth]{images/rnn_l96_outlier_case/RNN_1000_loss_val_loss}
		\caption{}
	\end{subfigure}

	\begin{subfigure}{0.60\linewidth}
	\includegraphics[width=\linewidth]{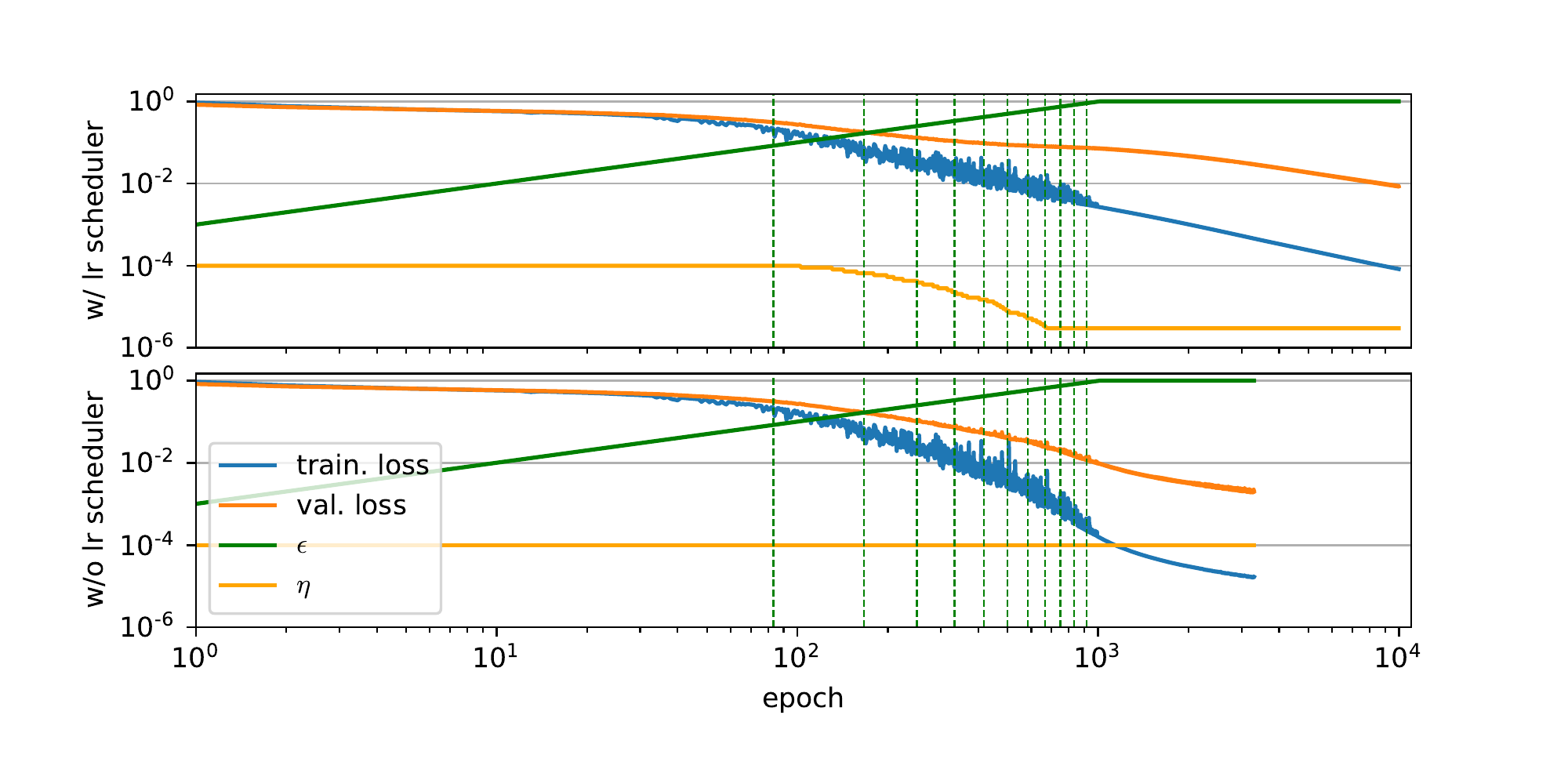}
	\caption{}
	\end{subfigure}

	\begin{subfigure}{0.60\linewidth}
		\includegraphics[width=\linewidth]{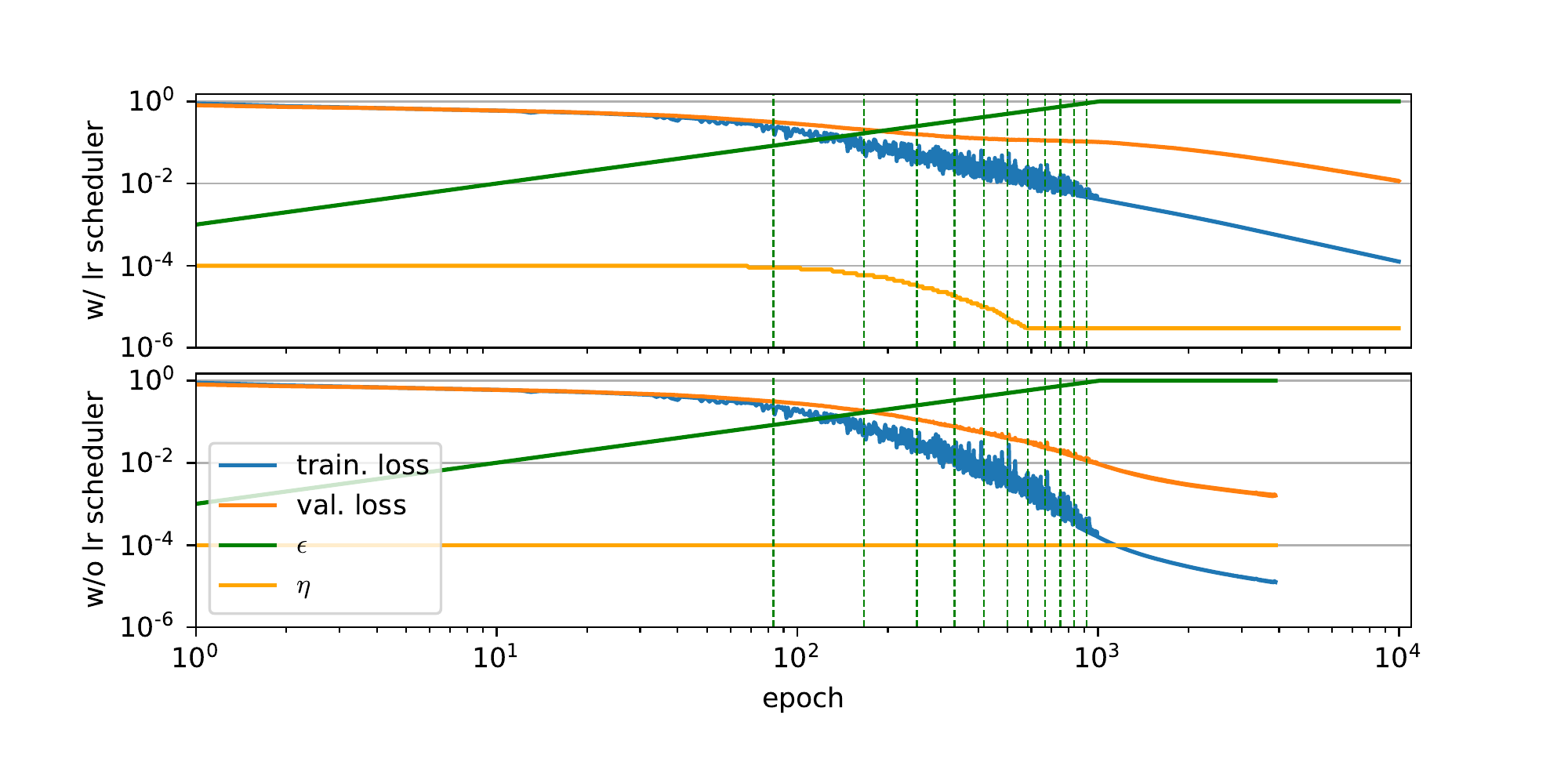}
		\caption{}
	\end{subfigure}
	\caption{Course of the training and validation loss for the basic RNN (a),  LSTM (b) and GRU (c) on Lorenz'96 data with and without using a learning rate scheduler.}
	\label{fig:apx_rnn_l96_loss_val_loss}
\end{figure*}

\newpage

\section{Different RNN Models}
\label{apx:different_rnns}
\begin{table*}[h]
	\caption{Forecasting performance of the vanilla RNN on four different chaotic datasets}
	\label{tab:rnn_results}
	\centering
	\resizebox{0.6\linewidth}{!}{
		\begin{tabular}{ll|rrrr}
			\toprule
			\textbf{System} & \textbf{Strategy} & \textbf{\L} &  \textbf{Epochs}    & \textbf{NRMSE} $\bm{\downarrow}$ & \textbf{Rel. impr.} $ \bm{\uparrow}$ \\
			\midrule
			\multirow{4}{*}{\rotatebox[origin=c]{0}{Thomas}}
			& FR                             & --                  & $ 51 $            & $ 0.48117 $                       & --            \\
			& TF                             & --                  & $ 249 $         & $ \underline{0.41274} $   & --           \\
			\cmidrule{2-6}
			& CL-ITF-P                  & $1\,000$      & $ 266 $         & $ \bm{0.17955} $                      & $\good{56.50\%}$    \\
			& CL-ITF-D                  &  $ 250 $       & $ 183 $         & $ 0.25659 $                       & $ \good{37.83\%} $ \\
			\cmidrule{1-6}
			\multirow{4}{*}{\rotatebox[origin=c]{0}{Rössler}}
			& FR                             & --           & $ 2\,292 $             & $ \underline{0.00747} $                     & --                                    \\
			& TF                             & --           & $ 1 $            & $ 0.50174 $  & --                                    \\
			\cmidrule{2-6}
			& CL-ITF-P                  & $ 500 $ & $ 2\,063 $           &  $ \bm{0.00283} $                      & $ \good{62.12\%} $      \\
			& CL-ITF-D                 & $ 1\,000 $    & $ 3\,075 $          & $ 0.00319 $                      & $ \good{57.30\%} $        \\
			\cmidrule{1-6}
			\multirow{4}{*}{\rotatebox[origin=c]{0}{Lorenz}}
			& FR                             & --           & $ 506 $             & $ 0.11389 $                     & --                                    \\
			& TF                             & --           & $ 572 $            & $ \underline{0.00913} $  & --                                    \\
			\cmidrule{2-6}
			& CL-ITF-P                  & $ 1\,000 $ & $ 746 $           &  $ 0.00603 $                      & $ \good{33.95\%} $      \\
			& CL-ITF-D                 & $ 125 $    & $ 782 $          & $ \bm{0.00378} $                      & $ \good{58.60\%} $        \\
			\cmidrule{1-6}
			\multirow{4}{*}{\rotatebox[origin=c]{0}{Lorenz'96}}
			& FR                             & --           & $ 1\,710 $             & $ \underline{0.31002}  $    & --                                    \\
			& TF                             & --           & $ 637 $            & $ 0.57870 $                           & --                                    \\
			\cmidrule{2-6}
			& CL-ITF-P                  & $ 1\,000 $ & $ 505 $           &  $ 0.56872 $                         & $ \bad{-83.45\%} $      \\
			& CL-ITF-D                 & $ 1\,000 $    & $ 6\,573 $          & $ \bm{0.07663} $                       & $ \good{75.28\%} $        \\
			\bottomrule
		\end{tabular}
	}
\end{table*}
\begin{table*}[h]
	\caption{Forecasting performance of the LSTM on four different chaotic datasets.}
	\label{tab:lstm_results}
	\centering
	\resizebox{0.6\linewidth}{!}{
		\begin{tabular}{ll|rrrr}
			\toprule
			\textbf{System} & \textbf{Strategy} & \textbf{\L} &  \textbf{Epochs}    & \textbf{NRMSE} $\bm{\downarrow}$ & \textbf{Rel. impr.} $ \bm{\uparrow}$ \\
			\midrule
			\multirow{4}{*}{\rotatebox[origin=c]{0}{Thomas}}
			& FR                             & --           & $ 45 $             & $  0.43698 $                     & --                                    \\
			& TF                             & --           & $ 818 $            & $ \underline{0.05265} $  & --                                    \\
			\cmidrule{2-6}
			& CL-ITF-P                  & $ 250 $ & $ 758 $           &  $ 0.01892 $                      & $ \good{64.06\%} $      \\
			& CL-ITF-D                 & $ 125 $    & $ 859 $          & $ \bm{0.01181} $                      & $ \good{77.57\%} $        \\
			\cmidrule{1-6}
			\multirow{4}{*}{\rotatebox[origin=c]{0}{Rössler}}
			& FR                             & --           & $ 2\,417 $             & $ 0.00210 $                      & --                                    \\
			& TF                             & --           & $ 1\,650 $             & $ \underline{0.00139} $  & --                                    \\
			\cmidrule{2-6}
			& CL-ITF-P                  & $ 1\,000 $ & $ 3\,426 $          &  $ \bm{0.00063}$                   & $ \good{54.68\%} $      \\
			& CL-ITF-D                 & $ 250 $    & $ 2\,367 $          & $ 0.00085 $                     & $ \good{38.85\%} $        \\
			\cmidrule{1-6}
			\multirow{4}{*}{\rotatebox[origin=c]{0}{Lorenz}}
			& FR                             & --           & $ 1\,154 $             & $ 0.06526  $                     & --                          \\
			& TF                             & --           & $ 398 $            & $ \underline{0.00169} $  & --                      \\
			\cmidrule{2-6}
			& CL-ITF-P                  & $ 250 $ & $ 806 $           &  $ \bm{0.00075} $                      & $ \good{55.62\%} $      \\
			& CL-ITF-D                 & $ 31 $ & $ 419 $          & $ 0.00125 $                      & $ \good{26.04\%} $         \\
			\cmidrule{1-6}
			\multirow{4}{*}{\rotatebox[origin=c]{0}{Lorenz'96}}
			& FR                             & --           & $ 4\,019 $             & $ \underline{0.13010} $   & --                                    \\
			& TF                             & --           & $ 3\,721 $            & $ 0.22757 $                       & --                                     \\
			\cmidrule{2-6}
			& CL-ITF-P                  & $ 500 $ & $ 9\,164 $           &  $ \bm{0.03935} $                      & $ \good{69.75\%} $      \\
			& CL-ITF-D                 & $ 62 $    & $ 4\,509 $          &  $ 0.06855 $                      & $ \good{47.31\%} $       \\
			\bottomrule
		\end{tabular}
	}
\end{table*}
\begin{table*}[h]
	\caption{Forecasting performance of the unitary evolution RNN on four different chaotic datasets.}
	\label{tab:urnn_results}
	\centering
	\resizebox{0.6\linewidth}{!}{
		\begin{tabular}{ll|rrrr}
			\toprule
			\textbf{System} & \textbf{Strategy} & \textbf{\L} &  \textbf{Epochs}    & \textbf{NRMSE} $\bm{\downarrow}$ & \textbf{Rel. impr.} $ \bm{\uparrow}$ \\
			\midrule
			\multirow{4}{*}{\rotatebox[origin=c]{0}{Thomas}}
			& FR                             & --           & $ 354 $             & $ \underline{0.74991} $                     & --                                    \\
			& TF                             & --           & $53 $            & $ 0.85679 $  & --                                    \\
			\cmidrule{2-6}
			& CL-ITF-P                  & $ 64\,000 $ & $ 359 $           &  $ \bm{0.61889} $                      & $ \good{17.47\%} $      \\
			& CL-ITF-D                 & $ 64\,000 $    & $ 340 $          & $ 0.73331 $                      & $ \good{2.21\%} $        \\
			\cmidrule{1-6}
			\multirow{4}{*}{\rotatebox[origin=c]{0}{Rössler}}
			& FR                             & --           & $ 911 $             & $ 0.03726 $                      & --                                    \\
			& TF                             & --           & $ 1\,287 $             & $ \underline{0.03317} $  & --                                    \\
			\cmidrule{2-6}
			& CL-ITF-P                  & $ 1\,000 $ & $ 1\,126 $          &  $ 0.02473 $                   & $ \good{25.44\%} $      \\
			& CL-ITF-D                 & $ 500 $    & $ 1\,402 $          & $ \bm{0.02420} $                     & $ \good{27.04\%} $        \\
			\cmidrule{1-6}
			\multirow{4}{*}{\rotatebox[origin=c]{0}{Lorenz}}
			& FR                             & --           & $ 758 $             & $ \underline{0.13021} $                     & --                          \\
			& TF                             & --           & $ 4 $            & $ 0.96357 $  & --                      \\
			\cmidrule{2-6}
			& CL-ITF-P                  & $ 1\,000 $ & $ 758 $           &  $ \bm{0.03098} $                      & $ \good{76.21\%} $      \\
			& CL-ITF-D                 & $ 500 $ & $ 1\,096 $          & $ 0.02897 $                      & $ \good{77.75\%} $         \\
			\cmidrule{1-6}
			\multirow{4}{*}{\rotatebox[origin=c]{0}{Lorenz'96}}
			& FR                             & --           & $ 1\,043 $             & $ \underline{0.26359} $   & --                                    \\
			& TF                             & --           & $ 398 $            & $ 0.61159 $                       & --                                     \\
			\cmidrule{2-6}
			& CL-ITF-P                  & $ 2\,000 $ & $ 697 $             &  $ 0.26253 $                      & $ \good{0.4\%} $      \\
			& CL-ITF-D                 & $ 2\,000 $    & $ 1\,335 $          &  $ \bm{0.18483} $                      & $ \good{29.88\%} $       \\
			\bottomrule
		\end{tabular}
	}
\end{table*}
\begin{table*}[h]
	\caption{Forecasting performance of the Lipschitz RNN on four different chaotic datasets.}
	\label{tab:lrnn_results}
	\centering
	\resizebox{0.6\linewidth}{!}{
		\begin{tabular}{ll|rrrr}
			\toprule
			\textbf{System} & \textbf{Strategy} & \textbf{\L} &  \textbf{Epochs}    & \textbf{NRMSE} $\bm{\downarrow}$ & \textbf{Rel. impr.} $ \bm{\uparrow}$ \\
			\midrule
			\multirow{4}{*}{\rotatebox[origin=c]{0}{Thomas}}
			& FR                             & --           & $ 879 $             & $ 0.11697 $                     & --                                    \\
			& TF                             & --           & $ 928 $            & $ \underline{0.05714} $  & --                                    \\
			\cmidrule{2-6}
			& CL-ITF-P                  & $ 250 $ & $ 879 $           &  $ 0.05237 $                      & $ \good{8.35\%} $      \\
			& CL-ITF-D                 & $ 500 $ & $ 1\,191 $          & $ \bm{0.03926} $                      & $ \good{31.29\%} $        \\
			\cmidrule{1-6}
			\multirow{4}{*}{\rotatebox[origin=c]{0}{Rössler}}
			& FR                             & --           & $ 3\,704 $             & $ \underline{0.27122} $                      & --                                    \\
			& TF                             & --           & $ 79 $             & $ 0.90844 $  & --                                    \\
			\cmidrule{2-6}
			& CL-ITF-P                  & $ 250 $ & $ 2\,304 $          &  $ 0.05185 $                   & $ \good{80.88\%} $      \\
			& CL-ITF-D                 & $ 500 $    & $ 3\,149 $          & $ \bm{0.04409} $                     & $ \good{83.74\%} $        \\
			\cmidrule{1-6}
			\multirow{4}{*}{\rotatebox[origin=c]{0}{Lorenz}}
			& FR                             & --           & $ 1\,298 $             & $ 0.06675 $                     & --                          \\
			& TF                             & --           & $ 625 $            & $ \underline{0.02830} $  & --                      \\
			\cmidrule{2-6}
			& CL-ITF-P                  & $ 62 $ & $ 914 $           &  $ \bm{0.01732} $                      & $ \good{38.80\%} $      \\
			& CL-ITF-D                 & $ 250 $ & $ 686 $          & $ 0.02417 $                      & $ \good{14.59\%} $         \\
			\cmidrule{1-6}
			\multirow{4}{*}{\rotatebox[origin=c]{0}{Lorenz'96}}
			& FR                             & --           & $ 489 $             & $ \underline{0.62710} $  & --                                    \\
			& TF                             & --           & $ 469 $            & $ 0.65122 $                       & --                                     \\
			\cmidrule{2-6}
			& CL-ITF-P                  & $ 1\,000 $ & $ 477 $             &  $ \bm{0.62170} $             & $ \good{0.86\%} $      \\
			& CL-ITF-D                 & $ 4\,000 $    & $ 489 $          &  $ 0.62581 $                      & $ \good{0.21\%} $       \\
			\bottomrule
		\end{tabular}
	}
\end{table*}

\end{document}